\documentclass[letterpaper]{article}
\usepackage{uai2020}
\usepackage[margin=1in]{geometry}
\usepackage{amssymb}
\usepackage{tikz}
\usepackage{xcolor}
\usepackage{makecell}
\usepackage{subfig}
\usepackage{float}
\usetikzlibrary{calc,arrows.meta,positioning,snakes,fit}
\usepackage{amsmath}
\usepackage{mathtools}
\usepackage{comment}
\usepackage{array,longtable,caption,bigstrut}
\usepackage{etoolbox}

\usepackage{hyperref}
\hypersetup{colorlinks,linkcolor={blue},citecolor={blue},urlcolor={red}}

\DeclareMathOperator{\sigmoid}{sigmoid}
\DeclareMathOperator{\softplus}{softplus}
\newcommand{\Rho}{\mathrm{P}}

\usepackage{times}

\usepackage[round]{natbib}
\title{Neural Likelihoods via Cumulative Distribution Functions}

\author{
	Pawel Chilinski\\
	University College London\\
	pawel.chilinski.14@ucl.ac.uk
	\And
	Ricardo Silva\thanks{Partially supported by EPSRC grant EP/N510129/1.}\\
	University College London 
	and The Alan Turing Institute\\
	ricardo@stats.ucl.ac.uk
}

\patchcmd{\thebibliography}{\chapter*}{\section*}{}{}

\begin{document}

\maketitle

\begin{abstract}
We leverage neural networks as universal approximators of monotonic
functions to build a parameterization of conditional cumulative
distribution functions (CDFs). By the application of automatic
differentiation with respect to response variables and then to
parameters of this CDF representation, we are able to build black box
CDF and density estimators. A suite of families is introduced as
alternative constructions for the multivariate case. At one extreme,
the simplest construction is a competitive density estimator against
state-of-the-art deep learning methods, although it does not provide
an easily computable representation of multivariate CDFs. At the other
extreme, we have a flexible construction from which multivariate CDF
evaluations and marginalizations can be obtained by a simple forward
pass in a deep neural net, but where the computation of the likelihood
scales exponentially with dimensionality. Alternatives in between the
extremes are discussed. We evaluate the different representations
empirically on a variety of tasks involving tail area probabilities,
tail dependence and (partial) density estimation.
\end{abstract}

\section{CONTRIBUTION}

We introduce a novel parameterization of multivariate cumulative
distribution functions (CDFs) using deep neural networks. We explain
how training can be done by a straightforward adaptation of standard
methods for neural networks. The main motivations behind our work
include: a direct evaluation of tail area probabilities; coherent
estimation of low dimensional marginals of a joint distribution
without the requirement of fitting a full joint; and
supervised/unsupervised density estimation.

The first two tasks benefit directly from a CDF computed by a forward
pass in a neural network, as tail probabilities and marginal CDFs can
be read-off essentially directly in this representation. The latter
has been tackled by an increasingly large literature on neural density
estimators. This dates back at least to \cite{Bishop94mixturedensity},
who used multilayer perceptrons to encode conditional means,
variances, and mixture probabilities for a (conditional) mixture of
Gaussians. Recently, models using transformations of a
simple distribution into more complex ones were
proposed. \cite{DinhLaurent2014NNIC},\cite{DinhLaurent2016DeuR},\cite{PapamakariosGeorge2017MAFf},
\cite{HuangChin-Wei2018NAF} and \cite{DeCaoNicola2019BNAF} are
examples of the state of the art for density estimation. They use
invertible transformations from simple base distributions where the
determinant of the Jacobian is easy to compute and Monte Carlo
approaches for computing gradients become feasible. Depending on the
architecture, they are optimized for density estimation or
sampling. We show we remain competitive against these methods while
keeping a comparatively simple uniform structure with few
hyperparameters. For instance, compared to the method of
\cite{Bishop94mixturedensity}, there is no need to choose the base
distribution of the mixture nor the number of mixtures.

All of the above is predicated on how we construct multivariate CDFs.
The most direct extension from univariate to multivariate CDFs is
conceptually simple, but the calculation of the likelihood grows
exponentially in the dimensionality. This is essentially the
counterpart to computing a partition function in an undirected
graphical model, where here the problem is differentiation as opposed
to integration.  Compromises are discussed, including the relationship
to Gaussian copula models and other CDF constructions based on small
dimensional marginals.  One extreme sacrifices the ability of
representing a CDF by a single forward pass in exchange for
scalability to high dimensions, where we can compare it against
state-of-the-art neural density estimators.

This paper is organized as follows. Section \ref{sec:method} describes
our main approach. Related work is described in Sections
\ref{sec:related} and \ref{sec:related_detailed}. Experiments are
discussed in Section \ref{sec:experiments}. We show that our models
are competitive for density estimation while able to directly tackle
some modelling problems where recent neural-based models could not be
applied in an obvious manner.

\section{THE MONOTONIC NEURAL DENSITY ESTIMATOR}
\label{sec:method}

We now introduce the Monotonic Neural Density Estimator (MONDE),
inspired by neural network methods for parameterizing monotonic
functions. The primary usage of MONDE is to compute conditional CDFs
with a single forward pass in a deep neural network, while allowing
for the calculation of the corresponding conditional densities by
tapping into existing methods for computing derivatives in deep
learning. The latter is particularly relevant for likelihood-based
fitting methods such as maximum (composite) likelihood. We will focus
on the continuous case only, where probability density functions
(pdfs) are defined, although extensions to include mixed combinations
of discrete and continuous variables are straightforward by
considering difference operations as opposed to differentiation
operations. We start with the simplest but important univariate case,
where dependency between variables does not have to be modelled. We
progress through more complex constructions to conclude with the most
flexible but computationally demanding case, where we deal with
multivariate data without assuming any specific families of
distributions for the data generating process.

Here we use the following notation.\\
$F(\mathbf{y}|\mathbf{x})$: multivariate conditional CDF, where
$\mathbf{y} \in \mathbb{R}^K$ is the response vector and $\mathbf{x}
\in \mathbb{R}^D$ is a covariate vector;\\
$F_k(y_k|\mathbf{x})$: $k$-th marginal conditional CDF;\\
$f(\mathbf{y}|\mathbf{x})$: multivariate conditional pdf;\\
$f_k(y_k|\mathbf{x})$: $k$-th marginal conditional pdf;\\
$w$: the set of parameters of a neural network, where $w_{ij}^l$ represents
a particular weight connecting two nodes $i$ and $j$, with $i$ located at layer $l$
of the network.

\subsection{UNIVARIATE CASE}
\label{sec:univariate}
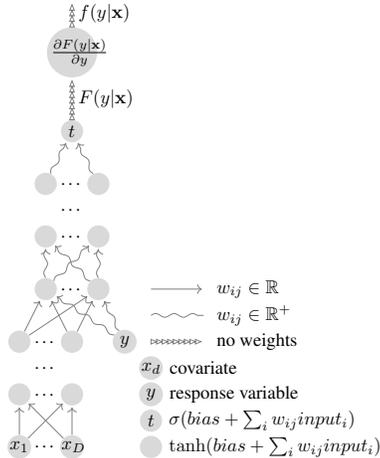
\begin{figure}
\centering
\scalebox{.7}{\def\layersep{1}

\begin{tikzpicture}[draw=black!50, node distance=\layersep]
            
    \tikzstyle{every pin edge}=[<-,shorten <=1pt]
    \tikzstyle{positive}=[snake=coil,segment aspect=0,segment amplitude=1pt];
    \tikzstyle{neuron}=[circle,fill=gray!30,minimum size=12pt,inner sep=0pt]
    \tikzstyle{input x neuron}=[neuron, fill=gray!30,text width=4mm,align=center];
    \tikzstyle{input y neuron}=[neuron, fill=gray!30,text width=4mm,align=center];
    \tikzstyle{cdf neuron}=[neuron, fill=gray!30];
    \tikzstyle{pdf neuron}=[neuron, fill=gray!30,text width=8mm,align=center];
    \tikzstyle{hidden neuron}=[neuron, fill=gray!30,text width=4mm,align=center];
    \tikzstyle{annot} = [text width=4em, text centered]
    \tikzstyle{arrow}=[shorten >=1pt,->]

    \node[input x neuron] (x1) at (0,0) {$x_1$};
    \node[input x neuron] (xd) at (\layersep,0) {$x_D$};    
    \node at ($(x1)!.5!(xd)$) {\ldots};
    
    \node[hidden neuron] (xh1) at (0,\layersep) {};
    \node[hidden neuron] (xhd) at (\layersep,\layersep) {};
    \node (xh_dots) at ($(xh1)!.5!(xhd)$) {\ldots};
    
    \path[arrow] (x1) edge (xh1);
    \path[arrow] (x1) edge (xhd);
    \path[arrow] (xd) edge (xh1);
    \path[arrow] (xd) edge (xhd);
    
    \node[hidden neuron] (xh_out1) at (0,2*\layersep) {};
    \node[hidden neuron] (xh_outd) at (\layersep,2*\layersep) {};
    \node (xh_out_dots) at ($(xh_out1)!.5!(xh_outd)$) {\ldots};
    
    \node at ($(xh_dots)!.5!(xh_out_dots)$) {\ldots};
    
    \node[input y neuron] (y) at (2*\layersep,2*\layersep) {$y$};
    
    \node[hidden neuron] (xy_1) at (0.5,3*\layersep) {};
    	\node[hidden neuron] (xy_d) at (1.5,3*\layersep) {};
    	\node (xy_dots) at ($(xy_1)!.5!(xy_d)$) {\ldots};
    	
    \path[arrow] (xh_out1) edge (xy_1);
    \path[arrow] (xh_out1) edge (xy_d);
    \path[arrow] (xh_outd) edge (xy_1);
    \path[arrow] (xh_outd) edge (xy_d);
    \draw[arrow,positive]  (y) -- (xy_1);
    \draw[arrow,positive] (y) -- (xy_d);
    
    \node[hidden neuron] (xy_h1_1) at (0.5,4*\layersep) {};
    	\node[hidden neuron] (xy_h1_d) at (1.5,4*\layersep) {};
    	\node (xy_h1_dots) at ($(xy_h1_1)!.5!(xy_h1_d)$) {\ldots};
    	
    \draw[positive,arrow] (xy_1) -- (xy_h1_1);
    \draw[positive,arrow] (xy_1) -- (xy_h1_d);
    \draw[positive,arrow] (xy_d) -- (xy_h1_1);
    \draw[positive,arrow] (xy_d) -- (xy_h1_d);
    	
    \node[hidden neuron] (xy_h2_1) at (0.5,5*\layersep) {};
    	\node[hidden neuron] (xy_h2_d) at (1.5,5*\layersep) {};
    	\node (xy_h2_dots) at ($(xy_h2_1)!.5!(xy_h2_d)$) {\ldots};
     
    \node at ($(xy_h1_dots)!.5!(xy_h2_dots)$) {\ldots};
    
    \node[cdf neuron] (cdf) at (1,6*\layersep) {$t$};
    
    \draw[positive,arrow] (xy_h2_1) -- (cdf);
    \draw[positive,arrow] (xy_h2_d) -- (cdf);
    
    \node[pdf neuron] (pdf) at (1,7.5*\layersep) {$\frac{\partial{F(y|\mathbf{x})}}{\partial{y}}$};
    
    \draw[snake=triangles,segment object length=3pt, segment length=3pt] (cdf) -- node[right]{$F(y|\mathbf{x})$} (pdf);
    
    	 \draw[snake=triangles,segment object length=3pt, segment length=3pt] (pdf) -- node[right]{$f(y|\mathbf{x})$} (1,8.5*\layersep);

    	\path[arrow] (2.5, 3) edge node[at end, label=right:{$w_{ij} \in \mathbb{R}$}]{} (3.5,3);
    	\draw[positive] (2.5, 2.5) -- node[at end, label=right:{$w_{ij} \in \mathbb{R}^+$}]{} (3.5,2.5);
    	\draw[snake=triangles,segment object length=3pt, segment length=3pt] (2.5,2) -- node[at end, label=right:{no weights}]{} (3.5,2);
    	\node[input x neuron] (input_x_legend) at (2.5,1.5) {$x_d$} node[right=0pt of input_x_legend] {covariate};
    	\node[input y neuron] (input_y_legend) at (2.5,1) {$y$} node[right=0pt of input_y_legend] {response variable}; 
    	\node[cdf neuron] (cdf_legend) at (2.5,0.5) {$t$} node[right=0pt of cdf_legend] {$\sigma(bias+\sum_i w_{ij}input_i)$};
    	\node[hidden neuron] (hidden_legend) at (2.5,0) {} node[right=0pt of hidden_legend] {$\tanh(bias+\sum_i w_{ij}input_i)$}; 
    	
    	
    	

\end{tikzpicture}}
\caption{The graph representing Univariate Monotonic Neural Density
  Estimator computational structure. The last node symbolizes the
  operation of differentiating parameterized conditional distribution
  function $F(y|\mathbf{x})$ with respect to the input $y$. Its output
  $f(y|\mathbf{x})$ encodes the conditional density function. The
  legend explains the symbols used.}
\label{figure:monde_univariate}
\end{figure}
The structure of MONDE for univariate responses is sketched in
Figure\ref{figure:monde_univariate} as a directed graph\footnote{This
  graph is to be interpreted as a high-level simplified computation
  graph as opposed to a graphical model. It does not illustrate a
  generative process, hence the placement of output random variable
  $y$ as an input to other variables.} with two types of edges and a
layered structure so that two consecutive layers are fully connected,
with no further edges. The definition of layer in this case follows
immediately from the topological ordering of the graph. Covariates
$x_i, \dots, x_D$ and response variable $y$ are nodes without parents
in the graph, with the layer of the covariates defined to be layer
1. The response variable $y$ is positioned in some layer $1 < l_y <L$,
where $L$ is the final layer. Each intermediate node $i$ at layer$l$,
$h_i^l$, returns a non-linear transformation of a weighted sum of all
nodes in layer $l - 1$. Here, as commonly used in the neural network
literature and based on preliminary results from our experiments, we
use the hyperbolic tangent function such that $h_i^l=
\tanh\left(\sum_{v_j \in \mathcal V_{l - 1}} w_{ij}^l v_j
+w_{i0}^l\right)$, where $1 < l < L$ and $\mathcal V_l$ is the set of
nodes in layer $l$. The final layer $L$ consists of a single node
$t(y, \mathbf x) \equiv \sigmoid\left(\sum_{v_j \in \mathcal V_{L -
    1}} w_{ij}^L v_j + w_{i0}^L\right)$, representing the probability
$P(Y \leq y~|~\mathbf X = \mathbf x)$ as encoded by the weights of the
neural net. In other words, $t(y, \mathbf x)$ is interpreted as a CDF
$F_w(y~|~\mathbf x)$ encoded by some $w$.

Assuming $w$ parameterizes a valid CDF, we can use an automatic
differentiation method to generate the density function
$f_w(y~|~\mathbf x)$ corresponding to $F_w(y~|~\mathbf x)$ by
differentiating $t(y, \mathbf x)$ with respect to $y$. The same
principle behind backpropagation applies here, and in our
implementation we use Tensorflow\footnote{https://www.tensorflow.org}
to construct the computation graph that generates $t(y, \mathbf
x)$. Once the pdf is constructed, automatic differentiation can once
again be used, now with respect to $w$, to generate gradients to be
plugged into any gradient-based learning algorithm.

To guarantee that $t(y, \mathbf x)$ is a valid CDF, we must enforce
three constraints: (i) $\lim_{y \rightarrow -\infty} t(y, \mathbf x) =
0$; (ii) $\lim_{y \rightarrow +\infty} t(y, \mathbf x) = 1$; (iii)
$\partial{t(y, \mathbf x)} / \partial{y} \geq 0$. We chose the
following design to meet these conditions: for every $w_{ij}^l$ where
$v_j \in \mathcal V_{l - 1}$ is a descendant of $y$ in the
corresponding directed graph, enforce $w_{ij}^l \geq 0$. This means
that for all layers $l > l_y$, all weights $\{w_{ij}^l\}$ are
constrained to be non-negative, while $\{w_{i0}^l\}$, representing
bias parameters, are unconstrained. Meanwhile, $w_{ij}^l \in
\mathbb{R}$ for $l \leq l_y$. In Figure \ref{figure:monde_univariate},
the constrained weights are represented as squiggled edges. This
guarantees monotonicity condition (iii).  Due to the range of the
logistic function being $[0, 1]$ and $t(y, \mathbf x)$ being monotonic
with respect to $y$, (i) and (ii) are also guaranteed to some extent:
an arbitrary choice of parameter vector $w$ will not imply e.g. that
$\lim_{y \rightarrow +\infty} t(y, \mathbf x)=1$ since the
contribution of $y$ to the final layer is bounded by the tanh
non-linearity. However, by learning $w$, this limit is satisfied
approximately since a likelihood-based fitting method will favour
$\sum_{i = 1}^nF(y_{max}~|~\mathbf x^{(i)}) / n \approx
1$\footnote{The expression is an empirical estimate of the marginal
  $F(y_{max})$ which is 1 in the empirical distribution. The claim follows as our
  estimator is chosen to minimize the KL divergence with respect to
  the empirical distribution.}, where $n$ is the sample size, $i$
indexes the training sample, and $y_{max}$ is the maximum observed
training value of $Y$ in the sample. There are ways of ``normalizing''
$t(y, \mathbf x)$ so that the limit is achieved exactly for all
parameter configurations (see Section \ref{sec:pumonde}). We have not
found it mandatory in order to obtain satisfactory empirical
results. This happens even though our ``unnormalized'' implementation
is at a theoretical disadvantage, as it can potentially represent
densities with the total mass less than 1. This is verified by the
experiments described in Section \ref{sec:experiments}.

\subsection{AUTOREGRESSIVE MONDE}
\label{sec:autoregressive}
\begin{figure}
\centering
\scalebox{.7}{\def\layersep{1}

\begin{tikzpicture}[draw=black!50, node distance=\layersep]
	\tikzstyle{vector}=[rectangle,fill=gray!30,minimum size=12pt,inner sep=0pt]
	\tikzstyle{y vector}=[vector, fill=gray!30,text width=4mm,align=center];
	\tikzstyle{arrow}=[shorten >=1pt,->];
	\tikzstyle{neuron}=[circle, fill=gray!30,minimum size=12pt,inner sep=0pt,text width=8mm,align=center];
	
    \node[y vector] (x) at (-0.5*\layersep,0*\layersep) {$\mathbf{x}$};
    \node[y vector] (y) at (0.5*\layersep,0*\layersep) {$\mathbf{y_1^0}$};
    
    \node[y vector] (y_1_1) at (-1*\layersep,1*\layersep) {$\mathbf{y}_1^1$};
    \node[y vector] (y_M_1) at (1*\layersep,1*\layersep) {$\mathbf{y}_M^1$};
    \node at ($(y_1_1)!.5!(y_M_1)$) {\ldots};
    
    \path[arrow] (y) edge (y_1_1);
    \path[arrow] (y) edge (y_M_1);
    
    \path[arrow] (x) edge (y_1_1);
    \path[arrow] (x) edge (y_M_1);
    
    \node[y vector] (y_1_2) at (-1*\layersep,2*\layersep) {$\mathbf{y}_1^2$};
    \node[y vector] (y_M_2) at (1*\layersep,2*\layersep) {$\mathbf{y}_M^2$};
    \node (y_2_dots) at ($(y_1_2)!.5!(y_M_2)$) {\ldots};
    
    \path[arrow] (y_1_1) edge (y_1_2);
    \path[arrow] (y_1_1) edge (y_M_2);
    \path[arrow] (y_M_1) edge (y_1_2);
    \path[arrow] (y_M_1) edge (y_M_2);
    
    \node[y vector] (y_1_Lmin1) at (-1*\layersep,3*\layersep) {$\mathbf{y}_1^{L-1}$};
    \node[y vector] (y_M_Lmin1) at (1*\layersep,3*\layersep) {$\mathbf{y}_M^{L-1}$};
    \node (y_Lmin1_dots) at ($(y_1_Lmin1)!.5!(y_M_Lmin1)$) {\ldots};
    
    \node at ($(y_2_dots)!.5!(y_Lmin1_dots)$) {\ldots};
    
    \node[y vector] (y_1_L) at (0*\layersep,4*\layersep) {$\mathbf{y}_1^L$};
    
    \path[arrow] (y_1_Lmin1) edge (y_1_L);
    \path[arrow] (y_M_Lmin1) edge (y_1_L);
    
    \node[neuron] (dF1dy1) at (-1*\layersep,5.5*\layersep) {$\frac{\partial{F_1(y_1|\mathbf{x})}}{\partial{y_1}}$};
    \node[neuron] (dFKdyK) at (1*\layersep,5.5*\layersep) {$\frac{\partial{F_K(y_K|\mathbf{x},\mathbf{y}_{<K})}}{\partial{y_K}}$};
    \node (dFdy_dots) at ($(dF1dy1)!.5!(dFKdyK)$) {\ldots};
    
    \draw[snake=triangles,segment object length=3pt, segment length=3pt] (y_1_L) -- node[right]{$=F_1(y_1|\mathbf{x}),F_2(y_2|\mathbf{x},y_1) \ldots F_k(y_K|\mathbf{x},\mathbf{y}_{<K})$} (dFdy_dots);
    
    	\node[neuron] (prod) at (0*\layersep,6.5*\layersep) {$\prod$};
    	
    	 \draw[snake=triangles,segment object length=3pt, segment length=3pt] (dFKdyK) -- node[above right]{$f_K(y_K|\mathbf{x},\mathbf{y}_{<K})$} (prod); 
    	 \draw[snake=triangles,segment object length=3pt, segment length=3pt] (dF1dy1) -- node[above left]{$f_1(y_1|\mathbf{x})$} (prod);
    	 \draw[snake=triangles,segment object length=3pt, segment length=3pt] (prod) -- node[right]{$f(\mathbf{y}|\mathbf{x})$} (0*\layersep, 7.5*\layersep);
    
    \node[y vector] (x_legend) at (2*\layersep,1.5*\layersep) {$\mathbf{x}$} node[right=0pt of x_legend] {$=x_0 \ldots x_D$};
    \node[y vector] (y_1_0_legend) at (2*\layersep,1*\layersep) {$\mathbf{y}_1^0$} node[right=0pt of y_1_0_legend] {$=y_1 \ldots y_K$};
    \node[y vector] (y_m_l_legend) at (2*\layersep,0.5*\layersep) {$\mathbf{y}_m^l$} node[right=0pt of y_m_l_legend] {$=y_{m,1}^l \ldots y_{m,K}^l$};
    \node[y vector] (y_1_L_legend) at (2*\layersep,0*\layersep) {$\mathbf{y}_1^L$} node[right=0pt of y_1_L_legend] {$=F_1(y_1),F_2(y_2|y_1) \ldots F_k(y_K|\mathbf{y}_{<K})$};
    
\end{tikzpicture}}
\caption{Autoregressive Model, MONDE MADE, The Multivariate Monotonic
  Neural Density Estimator architecture with shared parametrization
  inspired by MADE. The square nodes of the computational graph
  contain vectors, to differentiate from the oval nodes representing
  scalar values.}
\label{figure:monde_ar_made}
\end{figure}
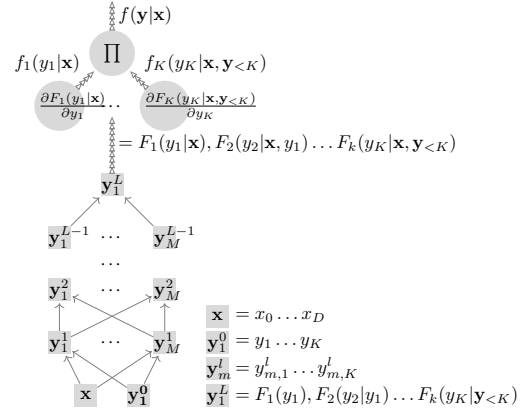
The first variation of the MONDE model, capable of efficiently encoding
multivariate distributions, is presented in the Figure
\ref{figure:monde_ar_made}. It uses a similar approach to univariate
output distributions, as described in Section \ref{sec:univariate}, to
parameterize each factor according to a fully connected probabilistic
directed acyclic graph (DAG) model. That is, we assume a given
ordering $y_1, \dots, y_K$ defining the fully connected DAG model:
\begin{equation}
f(\mathbf{y}~|~\mathbf{x}) = \prod_{k=1}^K f_k(y_k~|~\mathbf{x},\mathbf{y}_{<k}),
\end{equation}
\noindent where $\mathbf{y}_{<k}$ is set of response variables with
index smaller than $k$. In theory, the indexing of variables can be
chosen arbitrarily. In this work, we do not try to optimize it. This
type of DAG parameterization was called ``autoregressive'' in the
neural density estimator of \cite{DBLP:conf/nips/UriaML13}, a
nomenclature we use here to emphasize that this is a related
method. Our implementation of the autoregressive model uses parameter
sharing inspired by MADE \citep{GermainMathieu2015MMAf}.

The input to the computational graph is a $K$-dimensional vector
$\mathbf{y}$ of response variables and a $D$-dimensional vector
$\mathbf{x}$ of covariate variables. These vectors comprise the first
layer of the network. Each consecutive hidden layer is an affine
transformation of the previous layer proceeded by a nonlinear
elementwise map transforming its inputs via sigmoid function. Each
hidden layer is composed of $M$ $K$-dimensional vectors
$\mathbf{y}_m^l$, where $l$ indexes the layer and $m$ is vector index
within layer $l$. The affine transformation matrix is constrained so
that the $k$-th element of $m$-th vector, i.e. $\mathbf{y}_{m,k}^l$,
depends on a subset of the elements of the previous layer,
i.e. $\mathbf{y}_{.,<k}^{l-1}$, and is monotonically non-decreasing
with respect to $\mathbf{y}_{.,k}^{l-1}$. Here, dot $.$ represents all
possible indices $m \in \{1, 2, \dots, M\}$. Monotonicity is preserved
using non-negative weights in the respective elements of the
transformation matrix.

Finally, the $L$-th layer consists of a single $K$-dimensional vector
$\mathbf{y}_1^L$, with each element representing a CDF factor,
$F_1(y_1), \dots, F_k(y_k|\mathbf{x},\mathbf{y}_{<k})$. Each CDF
factor is differentiated with respect to its respective response
variable to obtain its pdf. The product of all pdf factors provide the
density function $f(\mathbf{y}|\mathbf{x})$. We provide the
implementation details in the supplement, Section
\ref{sec:autoregressive_sup}.
\subsection{GAUSSIAN COPULA MODELS}
\label{sec:copula}
\begin{figure}
\centering
\scalebox{.7}{\input{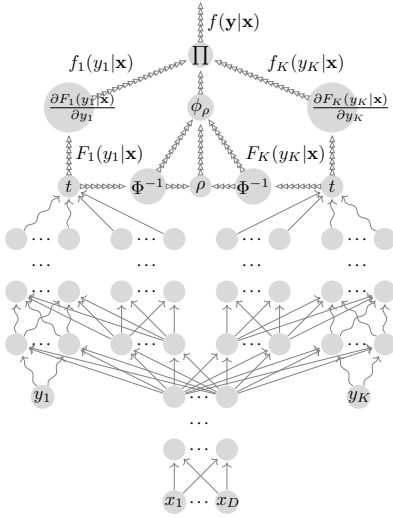}}
\caption{Multivariate Monotonic Neural Density Estimator with Gaussian Copula Dependency and Constant Covariance.}
\label{figure:monde_mv_constant_copula}
\end{figure}

A standard way of extending univariate models to multivariate models
is to use a copula model \citep{Skla59,Schmidt2006}. In a nutshell, we
can write a multivariate CDF $F(\mathbf y)$ as $F(\mathbf y) =
P(\mathbf Y \le \mathbf y) = P(F^{-1}(F(\mathbf Y)) \le \mathbf y) =
P(\mathbf U \leq F(\mathbf y))$. Here, $\mathbf U$ is a random vector
with uniformly distributed marginals in the unit hypercube and
$F^{-1}(\cdot)$ is the inverse CDF, applied elementwise to $\mathbf Y$, which will be unique for
continuous data as targeted in this paper. The induced multivariate
distribution with uniform marginals, $P(\mathbf U \leq \mathbf u)$, is
called the \emph{copula} of $F(\cdot)$. \cite{Elidan:2013} presents
an overview of copulas from a machine learning perspective.

This leads to a way of creating new distributions. Starting from a
multivariate distribution, we extract its copula. We then replace its
uniform marginals with any marginals of interest, forming a
\emph{copula model}. In the case of the multivariate Gaussian
distribution, the density function
\begin{align}
\label{eq:copula_pdf}
&f(\mathbf y)\!=\!\phi_{\mathbf{\rho}}(\Phi^{-1}(F_1(y_1)),\ldots,\Phi^{-1}(F_K(y_K)))\!\prod_{k=1}^{K}\! f_k(y_k)
\end{align}
is a Gaussian copula model where $\phi_{\mathbf{\rho}}$ is a Gaussian
density function with zero mean and correlation matrix $\rho$,
$\Phi^{-1}$ is the inverse CDF of the standard Gaussian, $f_k(\cdot)$
is any arbitrary univariate density function and $F_k(\cdot)$ its
respective distribution function. We can show that the $k$-th marginal
of this density is indeed $f_k(\cdot)$.

We extend the density estimator from the previous section to handle a
$K$-dimensional multivariate output $\mathbf y$ by exploiting two
(conditional) copula variations. The first variation is shown in
Figure \ref{figure:monde_mv_constant_copula}. Weight sharing is done
so that all output variables $y_k$ are placed in layer $l_y$, with all
weights $w_{ij}^{l'}$, $l' \leq l_y$, producing transformations of the
input $\mathbf x$ that is shared by all conditional marginals
$F_k(y_k~|~\mathbf x)$. From layers $l_{y} + 1, \dots, L$, the neural
network is divided into $K$ disjoint blocks, each composed of
two partitions: the first depending monotonically on its respective
$y_k$, and the second depending on shared transformation of $\mathbf
x$. The first partition depends on the second but not vice versa so
monotonicity with respect to $y_k$ is preserved (all the paths from
$y_k$ to $t_k$ in the computational graph use non-negative parameters,
as shown in the diagram). Each of the $K$ blocks generates output
$t_k(y_k, \mathbf x)$ representing an estimate of the corresponding
marginal $F_k(y_k~|~\mathbf x)$. The $k$-th marginal pdf can be obtained
by applying backpropagation with respect to $y_k$:
$f_k(y_k)=\partial{t_k(y_k, \mathbf x)}/\partial{y_k}$. Next, the
individual marginal distributions evaluated at each training point are
transformed via standard normal quantile functions. Such quantiles
$\Phi^{-1}(F_k(y_k~|~\mathbf x))$ are standard normal variables, which
we use to estimate the correlation matrix for the entire training
set. The estimated marginals and correlation matrix fully define our
model. Taking the product of the estimated copula and estimated
marginal densities (as shown in Equation \ref{eq:copula_pdf}) gives us
an estimate of the joint density with a correlation matrix that does
not change with $\mathbf x$ but which is simple to estimate by
re-using the univariate MONDE. We call this the Constant Covariance
Copula Model.

The next improvement, achieved at a higher computational cost, consists of
parameterizing the correlation matrix using a covariate
transformation. The diagram of this model is presented in Figure
\ref{figure:monde_mv_param_copula} in the supplement. This time the
correlation matrix is parameterized via a low rank factorization of
the covariance matrix which is a function of the covariates, 
allowing for a model with heteroscedasticity in the copula of the
output variables. The correlation matrix parameterization is as
follows:
\begin{align}
\mathbf{\Sigma}(\mathbf{x}) &= \mathbf{u}(\mathbf{x}) \cdot \mathbf{u}(\mathbf{x})^T + diag(\mathbf{d}(\mathbf{x})) \\
\mathbf{D}(\mathbf{x}) &\equiv \sqrt{diag(\mathbf{\Sigma}(\mathbf{x}))}, \\
\label{eq:monde_copul:cor_cov_parametrization}
\mathbf{\rho}(\mathbf{x}) &\equiv \mathbf{D}^{-1}(\mathbf{x}) \cdot \mathbf{\Sigma}(\mathbf{x}) \cdot \mathbf{D}^{-1}(\mathbf{x}),
\end{align}
 where $\mathbf{\Sigma}(\mathbf{x})$ is the covariate-parameterized
 low rank covariance matrix; $\mathbf{u}(\mathbf{x}) \in \mathbb{R}^K$
 and $\mathbf{d}(\mathbf{x}) \in \mathbb{R}_+^K$ are
 covariate-parameterized vectors; $diag$ is an operator which extracts
 a diagonal vector from the square matrix or creates a diagonal matrix
 from a vector (according to context); $\mathbf{\rho}(\mathbf{x})$ is
 the resulting covariate-parameterized correlation matrix.

\subsection{PUMONDE: PURE MONOTONIC NEURAL DENSITY ESTIMATOR}
\label{sec:pumonde}

\begin{figure}
\centering
\scalebox{.7}{\def\layersep{1}

\begin{tikzpicture}[draw=black!50, node distance=\layersep]
            
    \tikzstyle{every pin edge}=[<-,shorten <=1pt]
    \tikzstyle{positive}=[snake=coil,segment aspect=0,segment amplitude=1pt];
    \tikzstyle{neuron}=[circle,fill=gray!30,minimum size=12pt,inner sep=0pt]
    \tikzstyle{input x neuron}=[neuron, fill=gray!30,text width=4mm,align=center];
    \tikzstyle{input y neuron}=[neuron, fill=gray!30,text width=4mm,align=center];
    \tikzstyle{cdf neuron}=[neuron, fill=gray!30];
    \tikzstyle{pdf neuron}=[neuron, fill=gray!30,text width=8mm,align=center];
    \tikzstyle{hidden neuron}=[neuron, fill=gray!30,text width=4mm,align=center];
    \tikzstyle{annot} = [text width=4em, text centered]
    \tikzstyle{arrow}=[shorten >=1pt,->]
    
    \tikzset{
  		pics/relu/.style args={#1}{
     		code={
     			\node[hidden neuron] (#1) at (0,0) {};
       			\draw[color=black] (-1.5mm,-1mm) -- (0,-1mm) -- (1mm,1mm);
     		}
  		}
	}
	
	\tikzset{
  		pics/cdot/.style args={#1}{
     		code={
     			\node[hidden neuron] (#1) at (0,0) {};
       			\draw[color=black, fill=black] (0,0) circle (0.2mm);
     		}
  		}
	}

    \node[input x neuron] (x1) at (0,0) {$x_1$};
    \node[input x neuron] (xd) at (\layersep,0) {$x_D$};    
    \node at ($(x1)!.5!(xd)$) {\ldots};
    
    \node[hidden neuron] (xh1) at (0,\layersep) {$\sigma$};
    \node[hidden neuron] (xhd) at (\layersep,\layersep) {$\sigma$};
    \node (xh_dots) at ($(xh1)!.5!(xhd)$) {\ldots};
    
    \path[arrow] (x1) edge (xh1);
    \path[arrow] (x1) edge (xhd);
    \path[arrow] (xd) edge (xh1);
    \path[arrow] (xd) edge (xhd);
    
    \node[hidden neuron] (xh_out1) at (0,2*\layersep) {$\sigma$};
    \node[hidden neuron] (xh_outd) at (\layersep,2*\layersep) {$\sigma$};
    \node (xh_out_dots) at ($(xh_out1)!.5!(xh_outd)$) {\ldots};
    
    \node at ($(xh_dots)!.5!(xh_out_dots)$) {\ldots};
    
    \node[draw,dotted,thick,fit=(xh1) (xhd) (xh_out1) (xh_outd), inner sep=2pt] (x_block) {};    
    \node[below right, inner sep=0pt] at (x_block.west) {$h_{x}$};
    
	
    \node[input y neuron] (y_1) at (-\layersep,2*\layersep) {$y_1$};    
    
    \node[hidden neuron] (xy_1_1) at (-0.5,3*\layersep) {$\sigma$};
        \node[hidden neuron] (xy_1_d) at (-1.5,3*\layersep) {$\sigma$};
    	\node (xy_1_dots) at ($(xy_1_1)!.5!(xy_1_d)$) {\ldots};
    	
    \path[arrow] (xh_out1) edge (xy_1_1);
    \path[arrow] (xh_out1) edge (xy_1_d);
    \path[arrow] (xh_outd) edge (xy_1_1);
    \path[arrow] (xh_outd) edge (xy_1_d);
    \draw[arrow,positive]  (y_1) -- (xy_1_1);
    \draw[arrow,positive] (y_1) -- (xy_1_d);
    
    \node[hidden neuron] (xy_1_h1_1) at (-0.5,4*\layersep) {$\sigma$};
        \node[hidden neuron] (xy_1_h1_d) at (-1.5,4*\layersep) {$\sigma$};
    	\node (xy_1_h1_dots) at ($(xy_1_h1_1)!.5!(xy_1_h1_d)$) {\ldots};
    	
    \draw[positive,arrow] (xy_1_1) -- (xy_1_h1_1);
    \draw[positive,arrow] (xy_1_1) -- (xy_1_h1_d);
    \draw[positive,arrow] (xy_1_d) -- (xy_1_h1_1);
    \draw[positive,arrow] (xy_1_d) -- (xy_1_h1_d);
    	
    \node[hidden neuron] (xy_1_h2_d) at (-0.5,5*\layersep) {$\sigma$};
        \node[hidden neuron] (xy_1_h2_1) at (-1.5,5*\layersep) {$\sigma$};
    	\node (xy_1_h2_dots) at ($(xy_1_h2_1)!.5!(xy_1_h2_d)$) {\ldots};
     
    \node at ($(xy_1_h1_dots)!.5!(xy_1_h2_dots)$) {\ldots};
    
    \node[draw,dotted,thick,fit=(xy_1_1) (xy_1_d) (xy_1_h1_1) (xy_1_h1_d) (xy_1_h2_1) (xy_1_h2_d), inner sep=8pt] (xy_block_1) {};    
    \node[below right, inner sep=0pt] at (xy_block_1.north west) {$h_{xy1}$};

    \node[input y neuron] (y_k) at (2*\layersep,2*\layersep) {$y_2$};
    
    \node[hidden neuron] (xy_k_1) at (1.5,3*\layersep) {$\sigma$};
        \node[hidden neuron] (xy_k_d) at (2.5,3*\layersep) {$\sigma$};
    	\node (xy_k_dots) at ($(xy_k_1)!.5!(xy_k_d)$) {\ldots};
    	
    \path[arrow] (xh_out1) edge (xy_k_1);
    \path[arrow] (xh_out1) edge (xy_k_d);
    \path[arrow] (xh_outd) edge (xy_k_1);
    \path[arrow] (xh_outd) edge (xy_k_d);
    \draw[arrow,positive]  (y_k) -- (xy_k_1);
    \draw[arrow,positive] (y_k) -- (xy_k_d);
    
    \node[hidden neuron] (xy_k_h1_1) at (1.5,4*\layersep) {$\sigma$};
        \node[hidden neuron] (xy_k_h1_d) at (2.5,4*\layersep) {$\sigma$};
    	\node (xy_k_h1_dots) at ($(xy_k_h1_1)!.5!(xy_k_h1_d)$) {\ldots};
    	
    \draw[positive,arrow] (xy_k_1) -- (xy_k_h1_1);
    \draw[positive,arrow] (xy_k_1) -- (xy_k_h1_d);
    \draw[positive,arrow] (xy_k_d) -- (xy_k_h1_1);
    \draw[positive,arrow] (xy_k_d) -- (xy_k_h1_d);
    	
    \node[hidden neuron] (xy_k_h2_1) at (1.5,5*\layersep) {$\sigma$};
        \node[hidden neuron] (xy_k_h2_d) at (2.5,5*\layersep) {$\sigma$};
    	\node (xy_k_h2_dots) at ($(xy_k_h2_1)!.5!(xy_k_h2_d)$) {\ldots};
     
    \node at ($(xy_k_h1_dots)!.5!(xy_k_h2_dots)$) {\ldots};
    
    \node[draw,dotted,thick,fit=(xy_k_1) (xy_k_d) (xy_k_h1_1) (xy_k_h1_d) (xy_k_h2_1) (xy_k_h2_d), inner sep=8pt] (xy_block_k) {};    
    \node[below left, inner sep=0pt] at (xy_block_k.north east) {$h_{xy2}$};
    
    
    	\draw (0,6*\layersep) pic{cdot={xy_cdot_1}};
    	\draw (1,6*\layersep) pic{cdot={xy_cdot_n}};
    	\node (xy_cdot_dots) at ($(xy_cdot_1)!.5!(xy_cdot_n)$) {\ldots};
    	
    \node[draw,dotted,thick,fit=(xy_cdot_1) (xy_cdot_n), inner sep=5pt] (mult_block) {};    
    \node[below right, inner sep=0pt] at (mult_block.north west) {$m$};
    	
    	\draw[snake=triangles,segment object length=3pt, segment length=3pt] (xy_1_h2_1) --  (xy_cdot_1);
    	\draw[snake=triangles,segment object length=3pt, segment length=3pt] (xy_1_h2_d) -- (xy_cdot_n);
    	
    	\draw[snake=triangles,segment object length=3pt, segment length=3pt] (xy_k_h2_1) -- (xy_cdot_1);
    	\draw[snake=triangles,segment object length=3pt, segment length=3pt] (xy_k_h2_d) -- (xy_cdot_n);
    	
    \draw (0,7*\layersep) pic{relu={xy_cdot_h_1_1}};
    	\draw (1,7*\layersep) pic{relu={xy_cdot_h_1_n}};
    	\node (xy_cdot_h_1_dots) at ($(xy_cdot_h_1_1)!.5!(xy_cdot_h_1_n)$) {\ldots};
    
    \draw[arrow,positive]  (xy_cdot_1) -- node[below left]{} (xy_cdot_h_1_1);
    \draw[arrow,positive] (xy_cdot_1) -- (xy_cdot_h_1_n);
    \draw[arrow,positive]  (xy_cdot_n) -- (xy_cdot_h_1_1);
    \draw[arrow,positive] (xy_cdot_n) -- node[below right]{} (xy_cdot_h_1_n);
    
    \draw (0,8*\layersep) pic{relu={xy_cdot_h_2_1}};
    	\draw (1,8*\layersep) pic{relu={xy_cdot_h_2_n}};
    	\node (xy_cdot_h_2_dots) at ($(xy_cdot_h_2_1)!.5!(xy_cdot_h_2_n)$) {\ldots};
     
    \node at ($(xy_cdot_h_1_dots)!.5!(xy_cdot_h_2_dots)$) {\ldots};
    
	\draw (0.5,9*\layersep) pic{relu={xy_cdot_cdf_unnormalized}};
    
    \node[draw,dotted,thick,fit=(xy_cdot_cdf_unnormalized) (xy_cdot_h_2_1) (xy_cdot_h_2_n) (xy_cdot_h_1_1) (xy_cdot_h_1_n)] (t_transform) {};    
    \node[below right, inner sep=1pt] at (t_transform.north west) {$t$};
    
    \draw[arrow,positive]  (xy_cdot_h_2_1) -- (xy_cdot_cdf_unnormalized);
    \draw[arrow,positive] (xy_cdot_h_2_n) -- (xy_cdot_cdf_unnormalized);

    \draw[snake=triangles,segment object length=3pt, segment length=3pt] (xy_cdot_cdf_unnormalized) -- node[right]{$\propto F(\mathbf{y}|\mathbf{x})$} (0.5,10*\layersep);
    
        \node[hidden neuron] (sigma_legend) at (2,\layersep) {$\sigma$} node[right=0pt of sigma_legend] {$sigmoid(bias+\sum_j w_{j}input_j)$}; ;
    	\draw (2,0.5*\layersep) pic{relu={softplus_legend}} node[right=0pt of softplus_legend] {$softplus(bias+\sum_j w_{j}input_j)$}; 
    	\draw (2,0*\layersep) pic{cdot={cdot_legend}} node[right=0pt of cdot_legend] {Scalar multiplication};

\end{tikzpicture}}
\caption{Graph of an ``Unnormalized'' Distribution Function of
  PUMONDE, Pure Monotonic Neural Density Estimator. It shows two
  response variables $y_1$, $y_2$ and covariates $\mathbf{x}$
  transformed via computational graphs: $h_x$, $h_{xy1}$, $h_{xy2}$,
  $m$ and $t$.}
\label{figure:pumonde}
\end{figure}
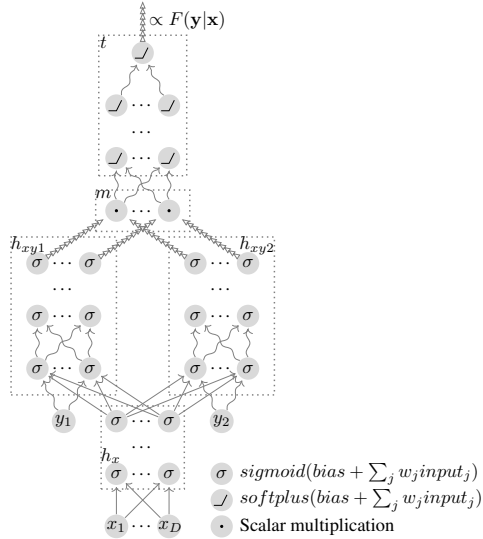

Our final model family is a flexible multivariate CDF parameterization. It can be combined with multivariate differentiation, with respect to multiple response variables, to
provide a likelihood function. The higher order derivative $\partial^K{t(\mathbf y, \mathbf x)}/\partial y_1 \dots \partial y_K$ has to be non-negative so that the model can represent a valid density function\footnote{This condition rules out $\sigmoid$ as the final   transformation of the computational graph for the distribution function because $\partial^2{\sigma(z)}/\partial^2{z} \in \mathbb{R}$, therefore this version of the CDF estimator uses different approach to map its output to be in the $(0,1)$ range.}. The graph representing a monotone function with respect to each response variable with no finite upper bound (to be later ``renormalized'') is presented in Figure \ref{figure:pumonde}. It is composed of several transformations, each of them represented in the computational graph as dashed rectangle containing the nodes and edges symbolizing its computations: $h_x$, a transformation the covariates using a standard multilayer network of sigmoid transformations; $h_{xyi}$, a sequential composition of monotonic nonlinear mappings starting with $h_x$ and response variable $y_i$; $m$, the element-wise
multiplication $\odot_{i = 1}^K h_{xyi}$ assuming all $h_{xyi}$ have the same dimensionality; and $t$, a monotonic transformation with respect to all its inputs that returns a positive real valued scalar. The last transformation $t$ uses only softplus as it non-linear transformations ($\softplus(x) \equiv \log(1+\exp{x})$). The function $t$ is non-decreasing with respect to any response variable on the same premises as previous models.

In this model, we replaced $\tanh$ with $\sigmoid$ and $\softplus$, where a hidden unit uses $\softplus$ if it has more than one ancestor in $y_1, \dots, y_K$ and $\sigmoid$ otherwise. This is because non-convex activation functions such as the $\sigmoid$ will not
guarantee e.g. $\partial^2 t(\mathbf y, \mathbf x) / \partial y_1 \partial y_2 \geq 0$ for units which have more than one target variable as an ancestor. Higher order derivatives with respect to the same response variable can take any real number because of the
properties of the computational graph i.e., using products of non-decreasing functions which are always positive and noting the fact that second order derivative with respect to the same response variable transformed by $\sigmoid$ can take any real value.

The density is then computed from the following transformations, here exemplified for a bivariate model:
\begin{align}
\label{eq:pumonde:distribution}
F_w(y_1,y_2~|~\mathbf{x})&=\frac{t(m(h_{xy1}(y_1,h_x(x)), h_{xy2}(y_2,h_x(x))))}{t(\mathbf{1})},\\
f_w(y_1,y_2~|~\mathbf{x})&=\frac{\partial^2{F_w(y_1,y_2~|~\mathbf{x})}}{\partial{y_1}\partial{y_2}}.
\end{align}
All output elements of $m(\cdot)$ have values in the $[0,1]$ range because it is element-wise multiplication of vectors with component values in $[0,1]$. By plugging-in the maximum value $\mathbf{1}$ as input of the $t$ transformation (as shown in denominator of Equation \ref{eq:pumonde:distribution}) we normalize the output of the distribution estimator $F_w$ to lie within $[0,1]$ so to output a valid CDF. The guarantee of non-decreasing monotonicity and positiveness of the $F_w$ with respect to each element of $\mathbf{y}$ assures that the range of the proposed estimator of a distribution function is in $[0,1]$\footnote{The discussion at the end of Section   \ref{sec:univariate}, about the univariate MONDE not being able strictly attain $0$ or $1$ is applicable here as well, because of the $h_{xyi}(y_i,\mathbf{x})$ transformation using bounded non-linearities. However, we can modify the initial layer at $l_y$ to simply monotonically map the real line to $[0, 1]$ (or whatever the support of each $Y_k$ is), and do the normalization with respect to the output of $l_y$ having value $\mathbf 1$, as opposed to the output of $m$.  We decided to omit this in order to make the description of the model simpler, and due to the lack of early evidence that this pre-processing was useful in practice.}.
  
\subsubsection{Composite Log-likelihood}
\label{sec:pumonde_cl}

It must be stressed that an unstructured PUMONDE with full connections will in general require an exponential number of steps (as a function of $K$) for the gradient to be computed, mirroring the problem of computing partition functions in undirected graphical models. Here we explore the alternative with the use of composite likelihood
\citep{Varin:2011}.

We train the PUMONDE model by minimizing the objective composed of the sum of the bivariate negative log-likelihoods ($LL$) for each pair of response variables (composite likelihood):
\begin{align}
LL &= \sum_{i=1..K,j=1..K, i<j} \log \frac{\partial^2{F_w(y_i,y_j|\mathbf{x})}}{\partial{y_i}\partial{y_j}}.
\end{align}
We compute estimates of such sums over mini-batches of data sampled from the training set. We update parameters using stochastic gradient descent as in other methods presented in this work. In the future, we want to check its role in graphical models for CDFs \citep{Huang:2008,Silva:2011}. For now we will restrict PUMONDE to small dimensional problems.
  
\section{RELATED WORK}
\label{sec:related}

Our work is inspired by the literature on neural networks applied to monotonic function approximation and to density estimation which is reviewed in the supplement Section \ref{sec:related_detailed}. 

\section{EXPERIMENTS}
\label{sec:experiments}

In this section, we describe experiments in which we compare our and
baseline models on various datasets and five success criteria. In what
follows, Tasks I, III and IV show how MONDE variations are competitive
against the state-of-the-art on modelling dependencies. Given that,
Tasks II and V advertise the convenience of a CDF parameterization
against other approaches.  As baselines, depending on the task, we use
the following models: RNADE
\citep{DBLP:conf/nips/UriaML13,Uria:2014:DTD:3044805.3044859}, MDN
\citep{Bishop94mixturedensity}, MADE \citep{GermainMathieu2015MMAf},
MAF \citep{PapamakariosGeorge2017MAFf}, TAN
\citep{OlivaJunierB.2018TAN} and NAF
\citep{HuangChin-Wei2018NAF,DeCaoNicola2019BNAF}.  More experiments
are included in the supplement, Section
\ref{sup:sec:experiments}.

\subsection{TASK I: DENSITY ESTIMATION}

\begin{table*}[th]
\tiny
\caption{Mean Loglikelihoods - Large UCI Datasets.}
\begin{center}
\begin{tabular}{lrrrrrrrrrr}
\makecell{} & \makecell{Power} & \makecell{Gas} & \makecell{Hepmass} & \makecell{Miniboone} &
\makecell{Bsds300} \\
\hline \\
\makecell[l]{MADE MoG}& $0.40 \pm 0.01$  & $8.47 \pm 0.02$ & $-15.15 \pm 0.02$ & $-12.27 \pm 0.47$ & $153.71 \pm 0.28$\\
\makecell[l]{MAF-affine (5)}& $0.14 \pm 0.01$ & $9.07 \pm 0.02$ & $-17.70 \pm 0.02$ & $-11.75 \pm 0.44$ & $155.69 \pm 0.28$\\
\makecell[l]{MAF-affine (10)}& $0.24 \pm 0.01$ & $10.08 \pm 0.02$ & $-17.73 \pm 0.02$ & $-12.24 \pm 0.45$ & $154.93 \pm 0.28$\\
\makecell[l]{MAF-affine MoG (5)}& $0.30 \pm 0.01$ & $9.59 \pm 0.02$ & $-17.39 \pm 0.02$ & $-11.68 \pm 0.44$ & $156.36 \pm 0.28$ \\
\makecell[l]{TAN (various architectures)}& $0.48 \pm 0.01$ & $11.19 \pm 0.02$ & $-15.12 \pm 0.02$ & $-11.01 \pm 0.48$ & $157.03 \pm 0.07$ \\
\makecell[l]{NAF}& $\mathbf{0.62 \pm 0.01}$ & $11.96 \pm 0.33$ & $-15.09 \pm 0.40$ & $\mathbf{-8.86 \pm 0.15}$ & $\mathbf{157.73 \pm 0.04}$ \\
\makecell[l]{B-NAF}& $0.61 \pm 0.01$ & $12.06 \pm 0.09$ & $\mathbf{-14.71 \pm 0.38}$ & $-8.95 \pm 0.07$ & $157.36 \pm 0.03$ \\
\hline \\
\makecell[l]{MONDE MADE}& \makecell{$ \mathbf{0.62 \pm 0.01}$} & \makecell{$\mathbf{12.12 \pm 0.02}$} & \makecell{$-15.83 \pm 0.06$} & \makecell{$-10.7 \pm 0.46$} & \makecell{$153.17 \pm 0.29$} \\
\hline
\end{tabular}
\end{center}

\label{table:uci_large_ll}
\end{table*}

In this section, we show results on density estimation using UCI
datasets. We use the same experimental setup as in
\citep{PapamakariosGeorge2017MAFf,HuangChin-Wei2018NAF,DeCaoNicola2019BNAF}
to compare recently proposed learning algorithms to one introduced in
this work. In particular, we evaluate a MONDE MADE variant which is
described in Section \ref{sec:autoregressive}. It is a simple
extension of our MONDE model to multivariate response variables using
autoregressive factorization. Among our methods, it is the only viable
option to be applied to high dimensional and large datasets that does
not make use of a parametric component, as in the Gaussian copula
variants. Results are presented in Table \ref{table:uci_large_ll},
which contains test log-likelihoods and error bars of 2 standard
deviations on five datasets. MONDE MADE matched the performance of the
state of the art NAF model from \citep{HuangChin-Wei2018NAF} for the
POWER dataset, and exceeded the performance of the NAF for the GAS
dataset. We achieved slightly worse results on the other UCI datasets
but we noticed that our model had a tendency to overfit the training
data in these cases. We have not applied techniques that could improve
generalization like batch normalization which were used in the
baseline models. We conclude that our models, by achieving comparable
results and having a complementary inductive bias to the baselines,
can be used as yet another tool for the benefit of practitioners.

\subsection{TASK II: TAIL EVENT CLASSIFICATION}
\label{sec:classification}

\begin{figure}[ht]
\begin{center}
\subfloat[ROC Curves]{\includegraphics[trim=20 0 20 0,clip,scale=0.28]{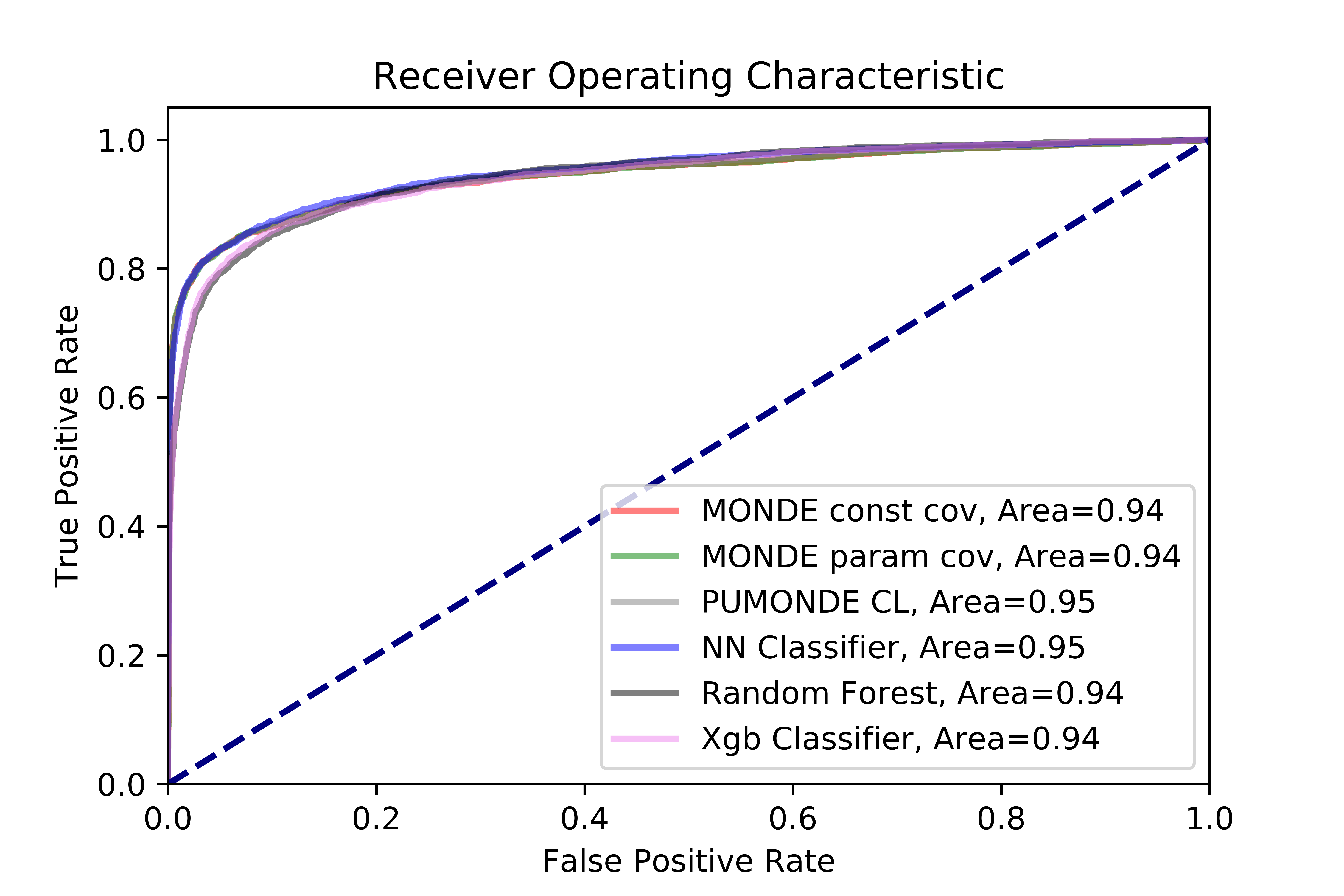}}
\hspace{\fill}
\subfloat[Precision/Recall Curves]{\includegraphics[trim=20 0 20 0,clip,scale=0.28]{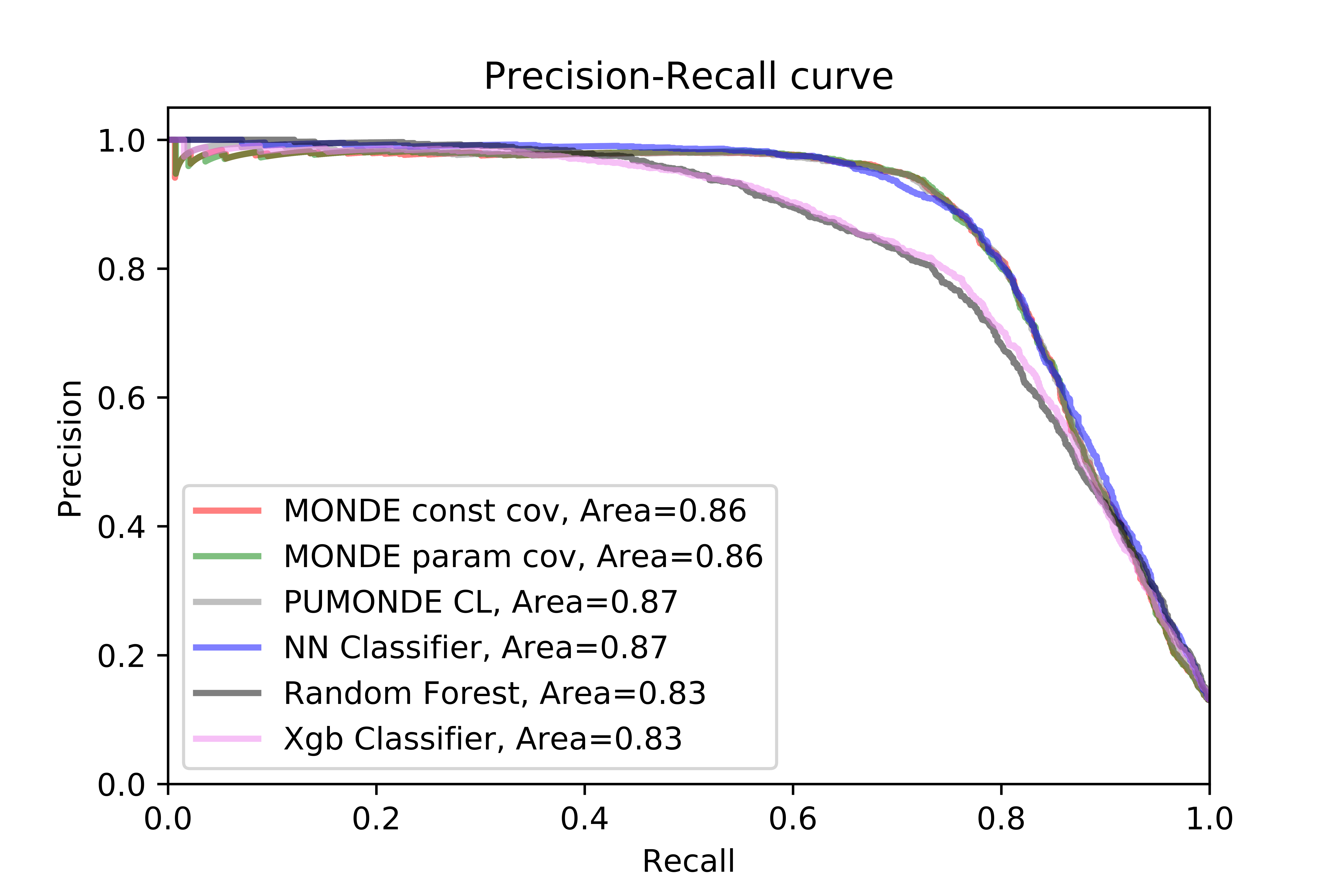}}
\end{center}
\caption{ROC Curves/AUC Scores (Area) and Precision-Recall Curves/Average Precision Scores
  (Area). RF and Xgb clustered together at a lower TPR and precision - Classification Task (better seen in color).}
\label{figure:roc_pr_all_models}
\end{figure}

We tested the Copula MONDE (Section \ref{sec:copula}) and PUMONDE
(Section \ref{sec:pumonde} and \ref{sec:pumonde_cl}) models on a
problem of detecting events falling at the tail of a distribution
which, for a fixed threshold defining the tail, can be compared
against standard classifiers. We use foreign exchange financial data
described in section \ref{sec:financial_data}. Data for the experiment
was prepared as follows: 1) Sample one minute negative log returns of
12 financial instruments. At each time $t_i$, we obtain a 12 element
vector $\mathbf{r}(t_i)=\log{\mathbf{p}(t_{i-1})} -
\log{\mathbf{p}(t_{i})}$, where $\mathbf{p}(t_{i})$ is the vector of
12 instruments mid prices at time $t_i$. Each $\mathbf{r}(t_i)$
represents a vector of 1 minute losses.  2) For each $t_i$, we collect
$\mathbf{y}=\mathbf{r}(t_i)_{10,11,12}$ and
$\mathbf{x}=\mathbf{r}(t_i)_{1..9},\mathbf{r}(t_{i-1})$. This
composition of data encodes a 3 dimensional response variable
representing 1 minute loss from the last 3 instruments at time $t_i$
and covariates are 1 minute losses from the rest of the instruments at
time $t_i$, combined together with the previous period $t_{i-1}$ 1
minute losses from all the instruments ($x$ is 21 dimensional
vector). We train our estimators on such constructed data by maximizing
the log-likelihood function.

We want to assess the models' ability to correctly rank tail events of
any of the 3 assets experiencing loss at least in the 95 percentile of
the historical loss in the next minute. To do this, we obtain the
95-th percentile threshold for each dimension of the $\mathbf{y}$
measured on the training set: $\mathbf{y}^{95}$. We compute the labels
on the test partition as: $l=1(\mathbf{y}_1 > \mathbf{y}^{95}_1\lor
\mathbf{y}_2 > \mathbf{y}^{95}_2 \lor \mathbf{y}_3
>\mathbf{y}^{95}_1)$ i.e. the label is $1$ whenever value at any of
the dimensions is larger then its 95-th percentile, otherwise is 0. We
compute the test score for the trained estimator by feeding it with
test set covariates $\mathbf{x}$ and plugging in $\mathbf{y}^{95}$ as
the response vector (the same response vector for each test covariate
vector). The CDF output from the model is the estimate of the
probability $P(\mathbf{Y}\leq
\mathbf{y}^{95}|\mathbf{x})=F(\mathbf{y}^{95}|\mathbf{x})$. We
estimate the tail probability of the label being equal 1
i.e. $P(L=1~|~\mathbf x)=P(\mathbf{Y}_1 > \mathbf{y}^{95}_1 \lor
\mathbf{Y}_2 >\mathbf{y}^{95}_2 \lor \mathbf{Y}_3 >
\mathbf{y}^{95}_1~|~\mathbf x) =1-F(\mathbf{y}^{95}|\mathbf{x})$. Such
computed ranks and the true labels are used to compute the Receiver
Operating Characteristic curve, Area Under Curve score,
Precision-Recall curve and Average Precision score on the test
partition of the dataset. These performance measures are used to
compare our estimators to a multilayer perceptron with sigmoid
outputs, Random Forests and Gradient Boosting Trees
\citep{Chen:2016}. These discriminative methods are trained directly
on labels pre-defined before training. Our estimators do not have to
use a particular threshold at a test time. It can be changed after
training is completed which is not possible for discriminative
models\footnote{This is analogous to a Bayesian Network providing the answer to
  any query, as opposed to a specialized predictor fit to answer a
  single predefined query.}.

ROC plots are shown in Figure \ref{figure:roc_pr_all_models}. ROC curves for the XGBoost and Random Forest classifiers cluster at the lower level of TPR for small values of FPR. The other models have ROC curves placed slightly higher. We see that results for all models are similar, where the multilayer perceptron and PUMONDE models achieved
slightly higher AUC score than the rest of the classifiers. PR plots are shown in Figure \ref{figure:roc_pr_all_models}. PR curves and average precision score (labelled ``Area'' in the legend of the Figure) tell a similar story.  In conclusion, we showed evidence that
our method is competitive in this task against black-boxed models finely tuned to a particular choice of threshold, but where we can instantaneoulsy re-evaluate classifications by changing the decision threshold without retraining the model. This is not possible with the baseline models, which are also less interpretable as they do not show how the distribution of the original continuous measurements changes
around the tails.

\subsection{TASK III: TAIL DEPENDENCE}
\label{sec:tail_dependence}
In this experiment, we assess whether our models can be used to estimate a measure of extreme dependence between two random variables $Y_i$ and $Y_j$, \emph{tail dependence} \citep{JoeHarry1997Mmad}:
\begin{align*}
\lambda_{L}(u) &= \lim_{u \to 0^+} P(Y_i \leq F_{i}^{-1}(u) | Y_j \leq F_{j}^{-1}(u) ) \\
&= \lim_{u \to 0^+} \frac{P(Y_i \leq F_{i}^{-1}(u) , Y_j \leq F_{j}^{-1}(u) )}{P(Y_i \leq F_{i}^{-1}(u))}
\end{align*} 
\begin{align*}
\lambda_{R}(u) &= \lim_{u \to 1^-} P(Y_i > F_{i}^{-1}(u) | Y_j > F_{j}^{-1}(u) ) \\
&= \lim_{u \to 1^-} \frac{P(Y_i > F_{i}^{-1}(u) , Y_j > F_{j}^{-1}(u) )}{P(Y_i > F_{i}^{-1}(u))} \\
&= \lim_{u \to 1^-} \frac{1 - 2u + F_{ij}(F_{i}^{-1}(u) , F_{j}^{-1}(u) )}{1-u},
\end{align*} 
where $\lambda_{L}(u)$ and $\lambda_{R}(u)$ are lower and upper tail dependence indices respectively, $F_{i}$ is the marginal distribution function for random variable $Y_i$, $F_{ij}$ is the bivariate marginal for random variables $Y_i$ and $Y_j$. In our experiment we use conditional distributions so distributions depend on covariates: $F_i(y_i~|~\mathbf{x})$ and $F_{ij}(y_i, y_j~|~\mathbf{x})$.

In order to have ground truth and provide some interpretability, we generate synthetic data as follows. We sample a Bernoulli random variable $C \in \{0, 1\}$ that indicates which of two components generates the covariates $\mathbf{X}$. The components are two Gaussian multivariate distributions with different means and identity covariance matrices. The choice of component also generates response variables $\mathbf{Y}$. In this case, the two distributions are such that the first is normally distributed (no tail dependence) and the second is t-distributed with 2 degrees of freedom. We repeat this process
independently for each point in the dataset: 
\begin{equation*}
\begin{aligned}[c]
\mathbf{C} \sim& Bernouli(0.5) \\
\mathbf{X} \sim& \mathbf{X}_c \\
\mathbf{X}_0 \sim& N((-2,-3), \mathbf{I}) \\ 
\mathbf{X}_1 \sim& N((2,5),\mathbf{I}) \\
\mathbf{Y} \sim& \mathbf{Y}_c \\ 
\mathbf{Y}_0 \sim& N((0,0,0), \Sigma) \\
\mathbf{Y}_1 \sim& t((0,0,0), 2, \Sigma) 
\end{aligned}
\qquad\qquad
\begin{aligned}[c]
\Sigma =& \sigma \Rho \sigma \\
\sigma =&
\begin{bmatrix}
0.4 & 0 & 0 \\
0 & 0.5 & 0 \\
0 & 0 & 0.8
\end{bmatrix}
\\
\Rho =&
\begin{bmatrix}
1.0 & 0.8 & 0.1 \\
0.8 & 1.0 & -0.5 \\
0.1 & -0.5 & 1.0
\end{bmatrix}.
\end{aligned}
\end{equation*}
\begin{figure}[ht!]
\begin{center}
\subfloat[Gaussian Component]{\includegraphics[scale=0.3]{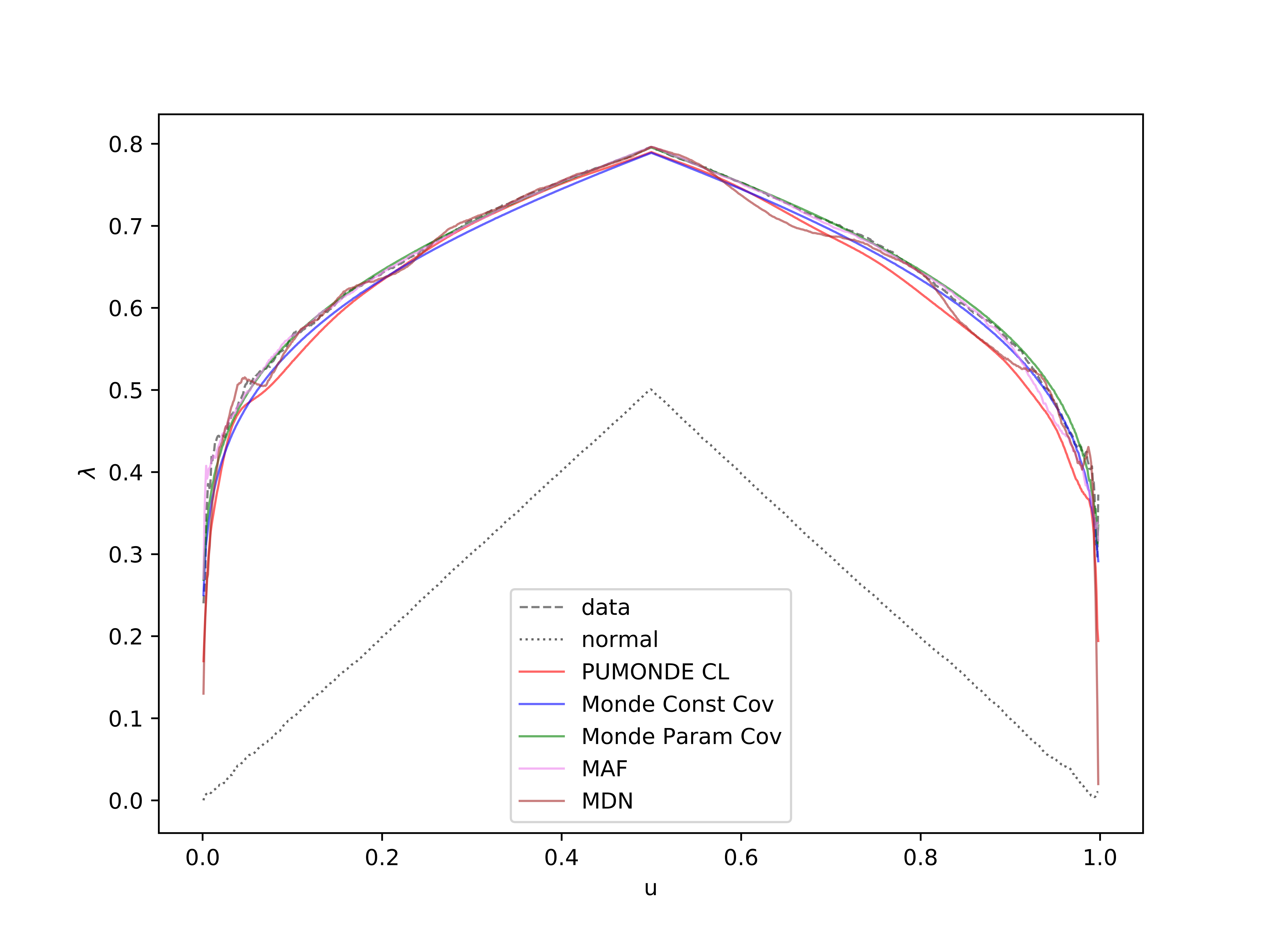}}\\
\subfloat[T Component]{\includegraphics[scale=0.3]{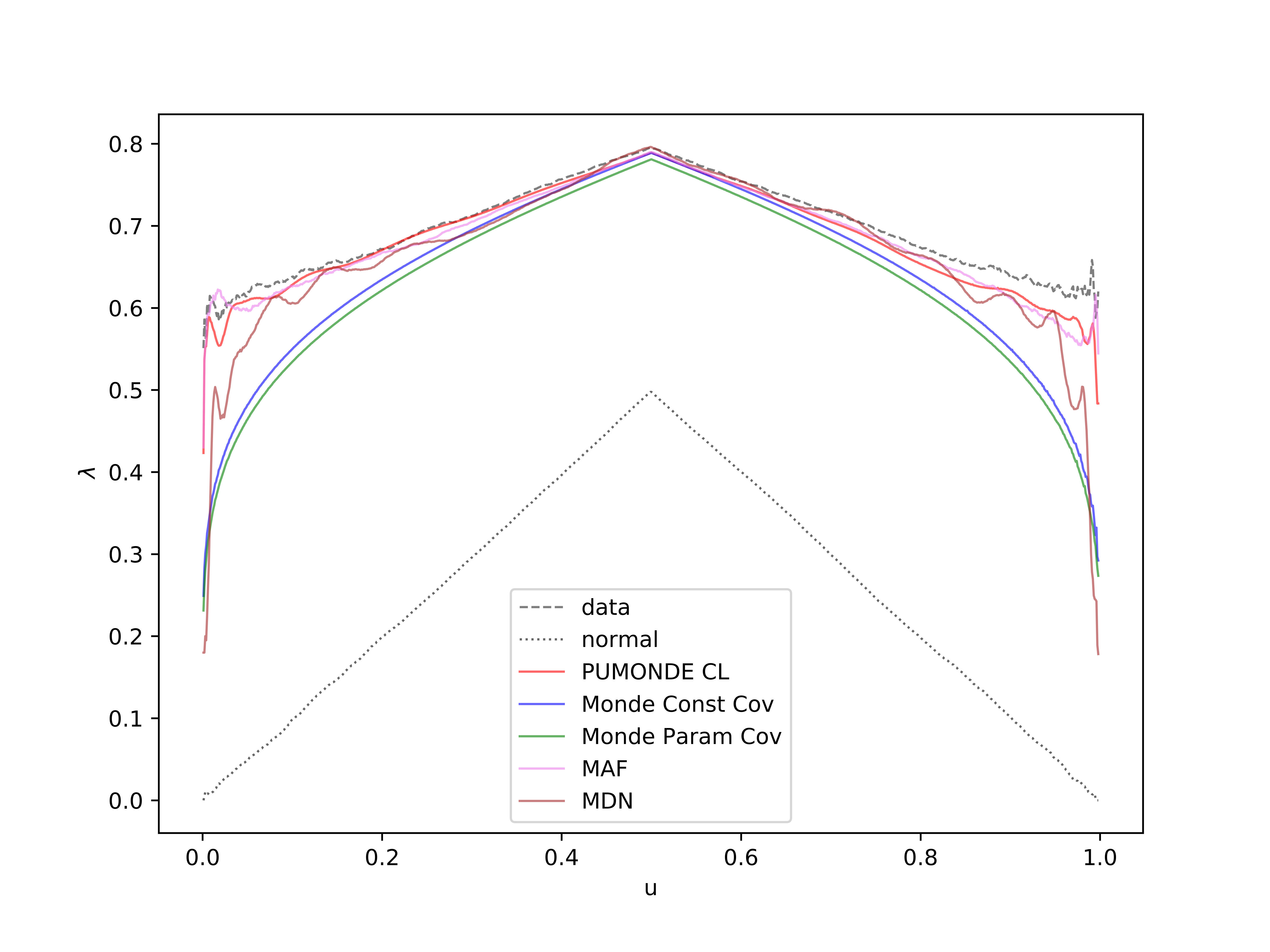}}
\end{center}
\caption{Tail Dependence Concentration Plots (better seen in
  colour). The triangle shaped curve in each plot is the tail
  dependence concentration plot for isotropic Gaussian (shown as
  comparison for curves depicting dependence and larger kurtosis). The
  first plot shows that all models correctly capture the lack of tail
  dependence in Gaussian distribution. The second plot shows that only
  PUMONDE and MAF concentration plots are close to the data
  concentration plot in the tails (when u tends to 0 or 1)}
\label{figure:tail_dependence}
\end{figure}
To illustrate concentration in the tails of the distribution, we plot
$\hat{\lambda}_{L}(u)$ for $u \in (0,0.5)$ and
$\hat{\lambda}_{R}(u)$ for $u \in (0.5,1)$ in Figure
\ref{figure:tail_dependence}. This includes models presented in this
paper and also two baseline models (MAF and MDN). We describe the
procedure used to compute these estimators in Section
\ref{sup:sec:tail_dependence}. We present two concentration plots.
The first one is for the response variable generated from the
multivariate Gaussians i.e. the first component, which does not
exhibit tail dependence. This can be observed in the curves which tend
to 0 when $u$ gets closer to 0 and 1. The second one which depicts
concentration plots for mixture component generated from the
t-distributed sample with 2 degrees of freedom clearly present tail
dependence which can be noticed by their limit converging to 0.6. We
can see that Copula and MDN models fail to capture tail dependence
(the first as expected, but the latter somehow has issues
approximating non-Gaussian tails with a mixture of Gaussians). The
PUMONDE model trained with composite likelihood (as described in
Section \ref{sec:pumonde_cl}) and MAF model are able to capture
``fatter'' tails in the data.

\subsection{TASK IV: MUTUAL INFORMATION}
\label{sec:experiment_mi}

\begin{table}[ht]
\tiny
\caption{Mutual Information.}
\begin{center}
\begingroup
\setlength{\tabcolsep}{1pt} 
\renewcommand{\arraystretch}{1} 
\begin{tabular}{lrrrrrr}
\makecell{} & \multicolumn{3}{c}{Gaussian Component} & \multicolumn{3}{c}{T Component} \\
\makecell{Model} & \makecell{$I(Y_0,Y_1)$} & \makecell{$I(Y_0,Y_2)$} & \makecell{$I(Y_1,Y_2)$} & \makecell{$I(Y_0,Y_1)$} & \makecell{$I(Y_0,Y_2)$} & \makecell{$I(Y_1,Y_2)$} \\
\hline \\
\makecell[l]{Data}		& \makecell[l]{$0.5108$}  & \makecell[l]{$0.0057$} & \makecell[l]{$0.1454$} & \makecell[l]{$0.5108$} & \makecell[l]{$0.0057$} & \makecell[l]{$0.1454$} \\
\makecell[l]{MAF} 		& \makecell[l]{$\mathbf{0.5107}$\\$\mathbf{(0.0001)}$}  & \makecell[l]{$0.0018$ \\ $(-0.004)$} & \makecell[l]{$0.1827$ \\ $(0.0373)$}   & \makecell[l]{$0.5786$\\$(0.0677)$} & \makecell[l]{$0.0831$\\$(0.0774)$} & \makecell[l]{$0.199$\\$(0.0536)$} \\
\makecell[l]{MDN} 		& \makecell[l]{$0.5172$\\$(0.0064)$}  & \makecell[l]{$0.0359$\\$(0.0301)$} & \makecell[l]{$0.1718$\\$(0.0264)$} & \makecell[l]{$0.6112$\\$(0.1004)$} & \makecell[l]{$0.1143$\\$(0.1085)$} & \makecell[l]{$0.2356$\\$(0.0901)$} \\
\makecell[l]{MONDE\\Const}& \makecell[l]{$0.4826$\\$(-0.0283)$}  & \makecell[l]{$0.0078$\\$(0.0021)$} & \makecell[l]{$\mathbf{0.1304}$\\$\mathbf{(-0.015)}$}  & \makecell[l]{$0.5363$\\$(0.0255)$} & \makecell[l]{$\mathbf{0.0414}$\\$\mathbf{(0.0357)}$} & \makecell[l]{$\mathbf{0.1431}$\\$\mathbf{(-0.0024)}$} \\
\makecell[l]{MONDE\\Param}& \makecell[l]{$0.5078$\\$(-0.003)$}  & \makecell[l]{$0.0796$\\$(0.0738)$} & \makecell[l]{$0.1303$\\$(-0.0151)$} & \makecell[l]{$0.438$\\$(-0.0728)$} & \makecell[l]{$0.0849$\\$(0.0792)$} & \makecell[l]{$0.1268$\\$(-0.0186)$} \\
\makecell[l]{PUMONDE} 	& \makecell[l]{$0.4682$\\$(-0.0425)$}  & \makecell[l]{$\mathbf{0.004}$\\$\mathbf{(-0.0017)}$} & \makecell[l]{$0.1105$\\$(-0.0349)$}  & \makecell[l]{$\mathbf{0.5307}$\\$\mathbf{(0.0199)}$} & \makecell[l]{$0.0573$\\$(0.0515)$} & \makecell[l]{$0.1621$\\$(0.0167)$}
\end{tabular}
\endgroup
\end{center}

\label{table:mutual_information}
\end{table}

Pairwise mutual information measures how much information is shared
between two random variables. It captures not only linear dependency,
but also more complex relations. It is defined as KL-divergence
between the joint bivariate marginal distribution and the product of
the corresponding univariate marginals. We now show results concerning
estimation of pairwise mutual information. The data generating process
is the same as in Section \ref{sec:tail_dependence}. We compute mutual
information by marginalizing distributions provided by the models
using numerical quadrature. We can apply this method because of the
low dimensionality of the problem.

Mutual information was computed for each pair of variables for the data generating process, two baseline models (MDN and MAF) and two of our models (Gaussian copula MONDE, PUMONDE). We compute mutual information for each mixture component separately by conditioning each model on the covariate equal to the mean vector for the given
covariate mixture component. The results are presented in Table \ref{table:mutual_information}. For each combination of model/pair of response variables/mixture component, we obtain two values: mutual information score and the absolute difference between mutual information value for the model and value for the data generating process (the difference is shown in brackets). The smallest absolute
value of the difference in a given column is highlighted in bold, indicating which model represents the closest mutual information to the one computed from the data generating process.  MONDE models are better in five out of six cases. We can conclude that models presented in this paper are competitive in encoding bivariate dependency with the current state of the art methods. Having this evidence leads us to our final Task, where we exploit estimating simultaneously multiple marginals of a common joint. CDF parameterizations are particularly attractive, as marginalization takes the same time as evaluating the joint model \citep{JoeHarry1997Mmad}, unlike some of the methods
discussed in this section.

\subsection{TASK V: BIVARIATE LIKELIHOOD}
\label{sec:bivariate_likelihood}

\begin{table}[ht]
\tiny
\caption{Bivariate Likelihood Model Comparison.}
\begin{center}
\begingroup
\setlength{\tabcolsep}{6pt} 
\renewcommand{\arraystretch}{1} 
\begin{tabular}{l|rrrr}

\makecell{} & \makecell{MDN} & \makecell{MONDE Const} & \makecell{MONDE Param} & \makecell{PUMONDE}  \\
\hline \\
\makecell[l]{MDN}		 & \makecell[l]{$NA$}  & \makecell[l]{$0$} & \makecell[l]{$0$} & \makecell[l]{$0$}  \\
\makecell[l]{MONDE Const}& \makecell[l]{$210$}  & \makecell[l]{$NA$} & \makecell[l]{$27$} & \makecell[l]{$0$}  \\
\makecell[l]{MONDE Param} & \makecell[l]{$210$}  & \makecell[l]{$183$} & \makecell[l]{$NA$} & \makecell[l]{$0$}  \\
\makecell[l]{PUMONDE} 	& \makecell[l]{$210$}  & \makecell[l]{$210$} & \makecell[l]{$210$} & \makecell[l]{$NA$}  \\
\end{tabular}
\endgroup
\end{center}

\label{table:bivariate_likelihood_summary}
\end{table}

In many practical problems, we are interested in estimating only
particular marginals. Parameters for higher order interactions are
considered to be nuisance parameters. Allowing for partial likelihood
specification is one of the primary motivations behind composite
likelihood \citep{Varin:2011}\footnote{This is not to be confused with
  another motivation, which is to provide a tractable replacement for
  the likelihood function. In this case, a full likelihood is still
  specified and of interest.  While computational tractability is a
  more common motivation in machine learning, partial specification is
  one of the main reasons for the development of composite likelihood
  in the statistics literature.}.

The problem with partial specification is that in general there are no
guarantees that the corresponding marginals come from any possible
joint distribution. On the other hand, a fully specified likelihood
has nuisance parameters. Ideally, we would like a flexible, overparameterized
joint model so that parameters are not obviously
responsible for any marginals a priori, with the objective function
regularizing them towards the marginals of interest. PUMONDE provides
such an alternative. Although high dimensional likelihoods
are intractable to compute in PUMONDE, low dimensional marginals are not.

In this section, we test the ability of our models to
encode coherent bivariate dependence in the data for problems of
larger dimensionality. For example, in finance this can be useful to
model second order dependence of returns in the portfolio of
instruments as used in computation of the Value at Risk metric
\citep{holton2003value}. We use foreign exchange financial data as
described in section \ref{sec:financial_data}.
This data contains a 21 dimensional response variable representing 1 minute losses for financial instruments at time $t_i$ and a 21 dimensional covariate variable representing 1 minute losses for all the instruments at time $t_{i-1}$. We train the estimators on such constructed data maximizing the likelihood objective. For PUMONDE, we optimize the composite likelihood objective comprised of the sum of all combinations of bivariate likelihoods. The only neural density estimator we use is MDN, fit to the 21 dimensional distribution. Marginalization in MDN is easy as it encodes a mixture of Gaussians, while the other baseline models cannot be easily marginalized.

To assess model performance, we compute the average log-likelihood for each bivariate combination of response variables on the test partition, giving 210 unique pairs. Each cell of Table \ref{table:bivariate_likelihood_summary} contains the number of times
the average log-likelihood computed for each bivariate combination of response variable was larger for model shown in the row compared to the model which is shown in the column.  We can see that the best performing model on this test is the PUMONDE which obtained larger test log likelihoods in all 210 cases when compared to each other
model. MONDE with parametrized covariance achieved better results than MDN and MONDE with constant covariance. The worst results were obtained by MDN.

\section{DISCUSSION}

We proposed a new family of methods for representing probability
distributions based on deep networks. Our method stands out from other
neural probability estimators by encoding directly the CDF. This
complements other methods for problems where the CDF representation is
particularly helpful, such as computing tail area probabilities and
computing small dimensional marginals. As future work, we will exploit
its relationship to graphical models for CDFs
\citep{Huang:2008,Silva:2011}, using PUMONDE to parameterize small
dimensional factors. Variations on soft-recursive partitioning, such
as hierarchical mixture of experts \citep{Jordan:1994}, can also be
implemented using tail events to define the partitioning
criteria. Another interesting and less straightforward venue of future
research is to exploit approximations to the likelihood based on the
link between differentiation, latent variable models and message
passing, as exploited in the context of graphical CDF models
\citep{Silva:2015,Huang:2010} and automated differentiation
\citep{Minka:2019}.

\begin{small}
\bibliography{uai2020}
\end{small}

\clearpage

\setcounter{equation}{0}
\setcounter{section}{0}
\setcounter{figure}{0}
\setcounter{table}{0}
\setcounter{page}{1}
\makeatletter
\renewcommand{\theequation}{S\arabic{equation}}
\renewcommand{\thesection}{S\arabic{section}}
\renewcommand{\thefigure}{S\arabic{figure}}
\renewcommand{\bibnumfmt}[1]{[S#1]}
\renewcommand{\citenumfont}[1]{S#1}
\makeatother

  \twocolumn[{%
  \begin{@twocolumnfalse}

\begin{center}
\textbf{\large Supplementary Material: Neural Likelihoods via Cumulative Distribution Functions}
\end{center}
\vspace{10mm}
  \end{@twocolumnfalse}
}]


\section{RELATED WORK}
\label{sec:related_detailed}

In this section, we give an overview of the ideas that inspired our work
and  constitute the fundamental building blocks of our method.

\subsection{MONOTONIC NEURAL NETWORKS FOR FUNCTION APPROXIMATION}

The first approach to model monotonic functions with neural networks
added a penalty term to the learning objective function
\citep{DBLP:conf/nips/SillA96}. No hard constraints were imposed,
which for our purposes would mean negative density functions possibly
appearing during training and testing. \cite{DBLP:conf/nips/Sill97}
proposed a model that encodes a hard monotonicity constraint, which
was deemed necessary to make learning efficient for monotone
regression functions. The inputs were first transformed linearly into
disjoint groups of hidden units, by using constrained weights which
were positive for increasing monotonicity and negative for decreasing
monotonicity (weight constraints were enforced by exponentiating each
free parameter). The groups were processed by a ``max'' operator and a
``min'' operator. The max operator modelled the convex part of the
monotone function and the min operator modelled the concave part. The
whole model could be learned by gradient descent. Authors proved that
their model could approximate any continuous monotone function with
finite first partial derivative to a desired accuracy. This model had
several hyper-parameters, including the number of groups or number of
hyper-planes within the group.

\cite{DBLP:conf/icann/Lang05} proved that if the output and
hidden-to-hidden weights were positive for a single-layer network,
then they could constrain input-to-hidden layers weights selectively
to be positive/negative depending on monotonicity constraints on
selected input variables. In this way, they could choose for which
input variables they want to preserve monotonicity of the output. They
also observed that min/max networks \citep{DBLP:conf/nips/Sill97}
tended to be more expensive to train than this simple neural network
architecture with constrained weights.

The review by \cite{DBLP:journals/nn/MininVLD10} tested several
approaches on a variety of datasets which exhibited monotonicity on
some of the inputs. Authors concluded that there was no definite
winner and the evaluated approaches excel in different areas and
applications.

\subsection{NEURAL DENSITY ESTIMATION}

One of the first methods to build conditional density estimators using
neural networks was the Mixture Density Networks (MDNs) of
\cite{Bishop94mixturedensity}.  The main motivation behind this work
was the inability of standard regression models to summarize
multimodal outputs with conditional means. MDNs parameterize
conditional mixtures of Gaussians where the neural network outputs
mixing probabilities and Gaussian parameters for each mixture.

Another approach, presented by \cite{DBLP:journals/nca/Wang94},
trained a density estimator by fitting a monotonic neural network to
match a smoothed empirical CDF. Unlike our work, this was solely a
smoother that reconstructed the empirical CDF using a function
approximator. There was no likelihood function or supervised signal.

The method presented by \cite{LIKAS2001167} tackles the problem of
normalizing the output of a neural network to directly approximate
density functions. It uses numerical integration over the domain of
the function.

Two methods were presented by
\cite{DBLP:journals/tnn/Magdon-IsmailA02} for unconditional density
estimation using neural networks and CDFs. They both rely on the
empirical CDF as targets to be approximated, with no explicit
likelihood function being used during training. The
reliance on the empirical CDF to provide training signal also implies
that there is no straightforward way of adapting it to the conditional
density estimation problem.

%
An approach based on the autoregressive factorized representation of
the density function was presented by
\cite{DBLP:journals/jmlr/LarochelleM11} and
\cite{DBLP:conf/nips/UriaML13}. In the continuous case, the proposed
model parameterizes each conditional marginal using what is
essentially a MDN \citep{Bishop94mixturedensity}. All neural networks
that parameterize the mixtures partially share parameters, in a way
that also speeds up computation as the number of parameters will grow
only linearly with dimensionality. The model was extended in
\cite{Uria:2014:DTD:3044805.3044859} to use deep neural networks to
parameterize an ensemble of variable orderings.

\cite{DBLP:conf/annpr/Trentin16} observed that using a neural network
to approximate the CDF function, and then differentiating it to obtain
a density function estimate, can give poor results. This does not
affect MONDE, as our objective function maximizes the likelihood
function instead of directly approximating a measure of distance to
the empirical CDF as done by, e.g.,
\cite{DBLP:journals/tnn/Magdon-IsmailA02}. His proposal included a
model of the density function that must be normalized numerically.
\cite{DBLP:journals/corr/abs-1804-05316} improved on the work of
\cite{DBLP:journals/tnn/Magdon-IsmailA02}, which comes from using
hard monotonicity constraints instead of the penalization approach
used by the other authors.

\cite{DinhLaurent2014NNIC} uses a transformation of density
formula. This method applies an invertible transformation, with an
easily computable determinant of the Jacobian, to map data with
complex dependencies into simple factorized parametric
distributions. The MONDE models have the advantage of providing a
directly computable CDF, with the disadvantage of not providing a
straightforward way to sample from the learned distribution. This
estimator was extended by
\cite{DinhLaurent2016DeuR,PapamakariosGeorge2017MAFf,HuangChin-Wei2018NAF,DeCaoNicola2019BNAF}
and others, mainly by using more complex transformations with
tractable Jacobians.

A model that uses efficient parameter sharing to encode autoregressive
dependency structure was applied in
\cite{GermainMathieu2015MMAf}. Authors modified the autoencoder using
autoregressive transformations so the output can represent a valid
density function. The model outputs a binary density by using logistic
outputs or mixtures of parametric models for real valued data. The
ideas presented in this work were used later in
\cite{PapamakariosGeorge2017MAFf}. We use similar parametrization in
the autoregressive MONDE model described in section
\ref{sec:autoregressive} with an additional constraint on a subset of
parameters that have to be non-negative, so the output of the
estimator represents valid conditional CDFs.

\section{THE MONOTONIC NEURAL DENSITY ESTIMATOR}

\subsection{AUTOREGRESSIVE MONDE}
\label{sec:autoregressive_sup}

\subsubsection{MADE Implementation Details}
\label{sec_sup:autoregressive}

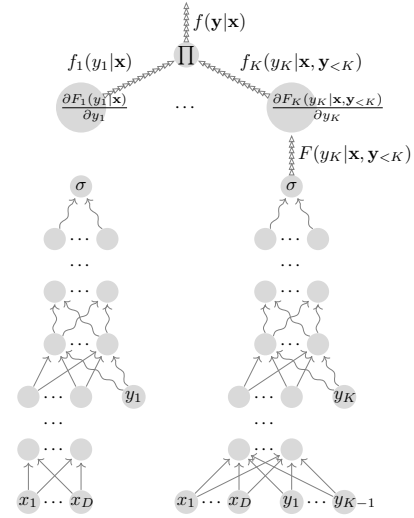
\begin{figure}
\centering
\scalebox{.7}{\def\layersep{1}

\begin{tikzpicture}[draw=black!50, node distance=\layersep]
            
    \tikzstyle{every pin edge}=[<-,shorten <=1pt]
    \tikzstyle{positive}=[snake=coil,segment aspect=0,segment amplitude=1pt];
    \tikzstyle{neuron}=[circle,fill=gray!30,minimum size=12pt,inner sep=0pt]
    \tikzstyle{input x neuron}=[neuron, fill=gray!30,text width=4mm,align=center];
    \tikzstyle{input y neuron}=[neuron, fill=gray!30,text width=4mm,align=center];
    \tikzstyle{cdf neuron}=[neuron, fill=gray!30];
    \tikzstyle{pdf neuron}=[neuron, fill=gray!30,text width=8mm,align=center];
    \tikzstyle{hidden neuron}=[neuron, fill=gray!30,text width=4mm,align=center];
    \tikzstyle{annot} = [text width=4em, text centered]
    \tikzstyle{arrow}=[shorten >=1pt,->]

    \node[input x neuron] (x1) at (0,0) {$x_1$};
    \node[input x neuron] (xd) at (\layersep,0) {$x_D$};    
    \node at ($(x1)!.5!(xd)$) {\ldots};
    
    \node[hidden neuron] (xh1) at (0,\layersep) {};
    \node[hidden neuron] (xhd) at (\layersep,\layersep) {};
    \node (xh_dots) at ($(xh1)!.5!(xhd)$) {\ldots};
    
    \path[arrow] (x1) edge (xh1);
    \path[arrow] (x1) edge (xhd);
    \path[arrow] (xd) edge (xh1);
    \path[arrow] (xd) edge (xhd);
    
    \node[hidden neuron] (xh_out1) at (0,2*\layersep) {};
    \node[hidden neuron] (xh_outd) at (\layersep,2*\layersep) {};
    \node (xh_out_dots) at ($(xh_out1)!.5!(xh_outd)$) {\ldots};
    
    \node at ($(xh_dots)!.5!(xh_out_dots)$) {\ldots};
    
    \node[input y neuron] (y_1) at (2*\layersep,2*\layersep) {$y_1$};
    
    \node[hidden neuron] (xy_1) at (0.5,3*\layersep) {};
    	\node[hidden neuron] (xy_d) at (1.5,3*\layersep) {};
    	\node (xy_dots) at ($(xy_1)!.5!(xy_d)$) {\ldots};
    	
    \path[arrow] (xh_out1) edge (xy_1);
    \path[arrow] (xh_out1) edge (xy_d);
    \path[arrow] (xh_outd) edge (xy_1);
    \path[arrow] (xh_outd) edge (xy_d);
    \draw[arrow,positive]  (y_1) -- (xy_1);
    \draw[arrow,positive] (y_1) -- (xy_d);
    
    \node[hidden neuron] (xy_h1_1) at (0.5,4*\layersep) {};
    	\node[hidden neuron] (xy_h1_d) at (1.5,4*\layersep) {};
    	\node (xy_h1_dots) at ($(xy_h1_1)!.5!(xy_h1_d)$) {\ldots};
    	
    \draw[positive,arrow] (xy_1) -- (xy_h1_1);
    \draw[positive,arrow] (xy_1) -- (xy_h1_d);
    \draw[positive,arrow] (xy_d) -- (xy_h1_1);
    \draw[positive,arrow] (xy_d) -- (xy_h1_d);
    	
    \node[hidden neuron] (xy_h2_1) at (0.5,5*\layersep) {};
    	\node[hidden neuron] (xy_h2_d) at (1.5,5*\layersep) {};
    	\node (xy_h2_dots) at ($(xy_h2_1)!.5!(xy_h2_d)$) {\ldots};
     
    \node at ($(xy_h1_dots)!.5!(xy_h2_dots)$) {\ldots};
    
    \node[cdf neuron] (cdf_1) at (1,6*\layersep) {$\sigma$};
    
    \draw[positive,arrow] (xy_h2_1) -- (cdf_1);
    \draw[positive,arrow] (xy_h2_d) -- (cdf_1);
    
    \node[pdf neuron] (pdf_1) at (1,7.5*\layersep) {$\frac{\partial{F_1(y_1|\mathbf{x})}}{\partial{y_1}}$};
    
    	 
    \node[input x neuron] (x1_K) at (3*\layersep,0) {$x_1$};
    \node[input x neuron] (xd_K) at (4*\layersep,0) {$x_D$};    
    \node at ($(x1_K)!.5!(xd_K)$) {\ldots};
	\node[input x neuron] (y1_K) at (5*\layersep,0) {$y_1$};
    \node[input x neuron] (yK_K) at (6*\layersep,0) {$y_{K-1}$};    
    \node at ($(y1_K)!.5!(yK_K)$) {\ldots};   
    
    \node[hidden neuron] (xh1_K) at (4*\layersep,\layersep) {};
    \node[hidden neuron] (xhd_K) at (5*\layersep,\layersep) {};
    \node (xh_dots_K) at ($(xh1_K)!.5!(xhd_K)$) {\ldots};
    
    \path[arrow] (x1_K) edge (xh1_K);
    \path[arrow] (x1_K) edge (xhd_K);
    \path[arrow] (xd_K) edge (xh1_K);
    \path[arrow] (xd_K) edge (xhd_K); 
    \path[arrow] (y1_K) edge (xh1_K); 
    \path[arrow] (y1_K) edge (xhd_K); 
    \path[arrow] (yK_K) edge (xh1_K); 
    \path[arrow] (yK_K) edge (xhd_K); 
    
    \node[hidden neuron] (xh_out1_K) at (4*\layersep,2*\layersep) {};
    \node[hidden neuron] (xh_outd_K) at (5*\layersep,2*\layersep) {};
    \node (xh_out_dots_K) at ($(xh_out1_K)!.5!(xh_outd_K)$) {\ldots};
    
    \node at ($(xh_dots_K)!.5!(xh_out_dots_K)$) {\ldots};
    
    \node[input y neuron] (y_K) at (6*\layersep,2*\layersep) {$y_K$}; 
    
    \node[hidden neuron] (xy_1_K) at (4.5*\layersep,3*\layersep) {};
    	\node[hidden neuron] (xy_d_K) at (5.5*\layersep,3*\layersep) {};
    	\node (xy_dots_K) at ($(xy_1_K)!.5!(xy_d_K)$) {\ldots};
    	
    \path[arrow] (xh_out1_K) edge (xy_1_K);
    \path[arrow] (xh_out1_K) edge (xy_d_K);
    \path[arrow] (xh_outd_K) edge (xy_1_K);
    \path[arrow] (xh_outd_K) edge (xy_d_K);
    \draw[arrow,positive]  (y_K) -- (xy_1_K);
    \draw[arrow,positive] (y_K) -- (xy_d_K);
	    	 
    \node[hidden neuron] (xy_h1_1_K) at (4.5*\layersep,4*\layersep) {};
    	\node[hidden neuron] (xy_h1_d_K) at (5.5*\layersep,4*\layersep) {};
    	\node (xy_h1_dots_K) at ($(xy_h1_1_K)!.5!(xy_h1_d_K)$) {\ldots};
    	
    \draw[positive,arrow] (xy_1_K) -- (xy_h1_1_K);
    \draw[positive,arrow] (xy_1_K) -- (xy_h1_d_K);
    \draw[positive,arrow] (xy_d_K) -- (xy_h1_1_K);
    \draw[positive,arrow] (xy_d_K) -- (xy_h1_d_K);
    	
    \node[hidden neuron] (xy_h2_1_K) at (4.5*\layersep,5*\layersep) {};
    \node[hidden neuron] (xy_h2_d_K) at (5.5*\layersep,5*\layersep) {};
    \node (xy_h2_dots_K) at ($(xy_h2_1_K)!.5!(xy_h2_d_K)$) {\ldots};
     
    \node at ($(xy_h1_dots_K)!.5!(xy_h2_dots_K)$) {\ldots};
    
    \node[cdf neuron] (cdf_K) at (5*\layersep,6*\layersep) {$\sigma$};
    
    \draw[positive,arrow] (xy_h2_1_K) -- (cdf_K);
    \draw[positive,arrow] (xy_h2_d_K) -- (cdf_K);
    
    \node[pdf neuron] (pdf_K) at (5*\layersep,7.5*\layersep) {$\frac{\partial{F_K(y_K|\mathbf{x},\mathbf{y}_{<K})}}{\partial{y_K}}$};
    
    \draw[snake=triangles,segment object length=3pt, segment length=3pt] (cdf_K) -- node[right]{$F(y_K|\mathbf{x},\mathbf{y}_{<K})$} (pdf_K);
    
    	 \node at ($(pdf_1)!.5!(pdf_K)$) {\ldots};
    	
    	\node[cdf neuron] (prod) at (3*\layersep,8.5*\layersep) {$\prod$};
    	
    	 \draw[snake=triangles,segment object length=3pt, segment length=3pt] (pdf_K) -- node[above right]{$f_K(y_K|\mathbf{x},\mathbf{y}_{<K})$} (prod); 
    	 \draw[snake=triangles,segment object length=3pt, segment length=3pt] (pdf_1) -- node[above left]{$f_1(y_1|\mathbf{x})$} (prod);
    	 \draw[snake=triangles,segment object length=3pt, segment length=3pt] (prod) -- node[right]{$f(\mathbf{y}|\mathbf{x})$} (3*\layersep, 9.5*\layersep);
\end{tikzpicture}}
\caption{Autoregressive MONDE, The Multivariate Monotonic Neural Density Estimator architecture.}
\label{figure:monde_ar}
\end{figure}

We now present implementation details about the model described in the
previous section.  Originally, the autoregressive model has been
implemented as shown in Figure \ref{figure:monde_ar}. But this
approach has not scaled well to a larger number of dimensions because
each component had its own independent computational graph. To speed
up the training and decrease the number of parameters, we applied the
ideas of \cite{GermainMathieu2015MMAf}, resulting in the model
presented in Figure \ref{figure:monde_ar_made}. Each layer $l$ has its
own parameter matrix that is constrained in a way such the computational
graph can represent valid conditional CDFs for each dimension, using
autoregressive representation. For example, covariates $\mathbf{x}$
and response variables $\mathbf{y}$ are transformed into the first hidden
layer according to the formula
\begin{align*}
\sigma(
\begin{bmatrix}
x_1 & \ldots & x_D & y_1^0 & \ldots & y_K^0
\end{bmatrix}
\times \mathbf{W}^1) = \\
\begin{bmatrix}
y_{1,1}^1 & y_{1,2}^1 & \ldots & y_{1,K}^1 & y_{2,1}^1 & \ldots & y_{M,K}^1
\end{bmatrix},
\end{align*}

where $\sigma(\cdot)$ is the sigmoid function (with range of $(0,1)$)
applied elementwise on the vector input; $\mathbf{W}^1$ is a
constrained matrix of size $ D+K \times KM$, where $D$ is the number
of dimensions of the covariate vector, $K$ is the dimension of the
response vector and $M$ is the number of $K$-element vectors in the
hidden layer; and matrix $\mathbf{W}^1$ has following structure,

\begin{align*}
\begin{bmatrix}
w^1_{1,1} & w^1_{1,2} & \ldots & w^1_{1,K} & \ldots & w^1_{1,KM}\\
w^1_{2,1} & w^1_{2,2} & \ldots & w^1_{2,K} & \ldots & w^1_{2,KM}\\
\ldots \\
w^1_{D,1} & w^1_{D,2} & \ldots & w^1_{D,K} & \ldots & w^1_{D,KM}\\
w^1_{D+1,1} & w^1_{D+1,2} & \ldots & w^1_{D+1,K} & \ldots & w^1_{D+1,KM}\\
\ldots \\
w^1_{D+K,1} & w^1_{D+K,2} & \ldots & w^1_{D+K,K} & \ldots & w^1_{D+K,KM}
\end{bmatrix},
\end{align*}

where

\begin{itemize}
\item $w^1_{i,.} \in \mathbb{R}$ for $i \in [1,D]$,
\item $w^1_{D+k,k+mK} \in \mathbb{R}^+ \cup \{0\}$ for $k \in [1,K]$, $m \in [0,M-1]$,
\item $w^1_{D+k_1,k_2+mK} \in \mathbb{R}$ for $k_1<k_2$, $k_{1,2} \in [1,K]$, $m \in [0,M-1]$,
\item $w^1_{D+k_1,k_2+mK}=0$ for $k_1>k_2$, $k_{1,2} \in [1,K]$, $m \in [0,M-1]$.
\end{itemize}

Non negative parameters are obtained by squaring the respective free
parameters of the transformation matrix. We constrain selected
parameters to be zero by multiplying the parameter matrix elementwise
by a mask matrix containing $0$ at locations which should be zeroed and
$1$ otherwise, as in the original MADE implementation. Each
$l-1$-th hidden layer is then transformed into the next $l$-th hidden layer
as follows:
\begin{align*}
\sigma(
\begin{bmatrix}
y_{1,1}^{l-1} & y_{1,2}^{l-1} & \ldots & y_{1,K}^{l-1} & y_{2,1}^{l-1} & \ldots & y_{M,K}^{l-1}
\end{bmatrix}
\times \mathbf{W}^l) = \\
\begin{bmatrix}
y_{1,1}^{l} & y_{1,2}^{l} & \ldots & y_{1,K}^{l} & y_{2,1}^{l} & \ldots & y_{M,K}^{l}
\end{bmatrix},
\end{align*}
where $\mathbf{W^l}$ is a constrained matrix of size $ KM \times KM$,
\begin{align*}
\begin{bmatrix}
w^l_{1,1} & w^l_{1,2} & \ldots & w^l_{1,K} & \ldots & w^l_{1,KM}\\
w^l_{2,1} & w^l_{2,2} & \ldots & w^l_{2,K} & \ldots & w^l_{2,KM}\\
\ldots \\
w^l_{KM,1} & w^l_{KM,2} & \ldots & w^l_{KM,K} & \ldots & w^l_{KM,KM}\\
\end{bmatrix},
\end{align*}
such that:
\begin{itemize}
\item $w^l_{k+m_1 K,k+m_2 K} \in \mathbb{R}^+ \cup \{0\}$ for $k \in [1,K]$, $m_{1,2} \in [0,M-1]$,
\item $w^l_{k_1+m_1 K,k_2+m_2 K} \in \mathbb{R}$ for  $k_1<k_2$, $k_{1,2} \in [1,K]$, $m_{1,2} \in [0,M-1]
$,
\item $w^l_{k_1+m_1 K,k_2+m_2 K}=0$ for  $k_1>k_2$, $k_{1,2} \in [1,K]$, $m_{1,2} \in [0,M-1]$.
\end{itemize}

The $L$-th layer outputs a $K$-dimensional vector in which each component represents the conditional CDF:
\begin{align*}
\sigma(
\begin{bmatrix}
y_{1,1}^{L-1} & y_{1,2}^{L-1} & \ldots & y_{1,K}^{L-1} & y_{2,1}^{L-1} & \ldots & y_{M,K}^{L-1}
\end{bmatrix}
\times \\ \mathbf{W}^L) =
\begin{bmatrix}
y_{1,1}^{L} & y_{1,2}^{L} & \ldots & y_{1,K}^{L}
\end{bmatrix},
\end{align*}
where $\mathbf{W}^L$ is constructed in similar way as hidden to hidden
parameter matrices. The presented composition of layers constrains the
$\mathbf{y}^L$ output to fulfil requirements that has to be met by
valid autoregressive representation of the joint CDFs i.e.
\begin{itemize}
\item $y^L_1 = t_w(y_1^+) = F_1(y_1)$,
\item $y^L_2 = t_w(y_1,y_2^+) = F_2(y_2|y_1)$,
\item $\ldots$
\item $y^L_K = t_w(\mathbf{y}_{<k},y_K^+) = F_1(y_k|\mathbf{y}_{<k})$.
\end{itemize}
Here, $t_w$ represents the final output of the computational graph of
the model as parameterized by $w={\mathbf{W}^1,...,\mathbf{W}^L}$. By
construction, $t_w(\mathbf{y}_{<k}, y_k^+)$ is nondecreasing monotonic
on input $y_k$, and unconstrained on inputs $y_1 \ldots y_{k-1}$.
Having obtained a parameterization that computes valid conditional
CDFs, we can construct the density estimator by computing the product
of the derivatives of conditional CDFs with respect to their
respective target variables,
\begin{align*}
f(\mathbf{y}|\mathbf{x})=f(y_1|\mathbf{x})f(y_2|\mathbf{x},y_1)\ldots f(y_k|\mathbf{x},\mathbf{y}_{<k})=\\
\frac{\partial{F_1(y_1|\mathbf{x})}}{\partial{y_1}} \frac{\partial{F_2(y_2|\mathbf{x},y_1)}}{\partial{y_2}} \ldots  \frac{\partial{F_K(y_K|\mathbf{x},\mathbf{y}_{<K})}}{\partial{y_K}}.
\end{align*}

We optimize parameters $w$ by maximizing expected log-likelihood using
a Monte Carlo estimate i.e. we maximize the average log-likelihood
over the batch of data points sampled randomly from the training
dataset.

\subsubsection{Finer Remarks}
\label{sec_sup:autoregressive:implementation_remarks}

In this section, we give further implementation details for the autoregressive model.

\begin{itemize}
\item We modified the MONDE approach to constrain the parameters
  matrices of \cite{GermainMathieu2015MMAf} by introducing non
  negative weights. This requirement is identical to the one used in
  univariate MONDE described in Section \ref{sec:univariate}.
\item Compared to MADE, we have the size of each hidden layer being
  equal to a multiple of the response vector size. In this way, each
  layer propagates the same number of autoregressive blocks of vectors
  that are used at the top layer to construct CDFs. This would be very
  inefficient for very high dimensional data. We have also implemented
  a version of the estimator which at each hidden layer we sample the
  nodes like it is done in MADE.
\item To stabilize the learning in the final stages of the training,
  we increase the batch size by a factor of 2 whenever we experience
  numerical problems during training (i.e. we use the last good
  parameters before gradient computation or network evaluation
  resulted in numerical problems, and restart training with doubled
  batch size). This method to increase batch size is inspired by
  \cite{SmithSamuelL.2017DDtL}. We found this to be a very important
  procedure used during training so we could achieve results
  comparable to the ones shown by \cite{HuangChin-Wei2018NAF}. We also
  increase the batch size after the performance of the estimator on
  the validation set does not improve on 10 consecutive training
  epochs. It remains to be checked whether introducing techniques like
  batch or activation normalization would render this approach
  unnecessary.
\item We found that using a scaled and translated $\tanh$ to the range
  of $(0,1)$ in all layers (hidden and final) helps to stabilize the
  learning process. We found empirically that using it causes fewer
  numerical problems than using directly the sigmoid function. There is at
  least one possible explanation of this phenomenon: $\tanh$ has
  larger gradients than a regular sigmoid but we have not verified it
  theoretically why it helped with optimization. It was already tested
  empirically that using $\tanh$ instead of sigmoid can be beneficial
  for the final result \citep[for example,][]{LiewShanSung2016Baff}. We
  tried using different non linearities in the hidden layer
  transformations such as softsign, softplus, ReLU and sigmoid but
  modified $\tanh$ led to the best results. We are aware that the
  initialization of the parameters can also affect the performance of
  the model but we have not performed any extensive study to choose the
  best approach to parameter initialization and we leave it to future
  work.
\item We tried to modify batch normalization \citep{IoffeSergey2015BNAD} which seems to be used in most recent neural density estimators. After trying it with many datasets we concluded that it sped up the rate of convergence but we obtained worse results in all cases compared with MONDE variations not using it. We think it requires further research and theoretical insight how to modify the batch normalization to help with training the autoregressive MONDE estimator.
\item We also trained a version of the model that used mixture of CDFs at the output of each autoregressive component. It is just an analogue to the Mixture Density Networks, where the base distributions are MONDE models and convex weights are given by a NN. This was achieved by the last layer being constructed from the softmax scaled parameter per component so we could compute mixture of multiple CDF outputs. We found that by using this extension we did not obtain statistically significantly better result so we decided not to include them in the experiments section.
\end{itemize}

\subsection{MULTIVARIATE COPULA MODELS}

\subsubsection{MONDE Parameterized Covariance}

\begin{figure}
\centering
\scalebox{.7}{\input{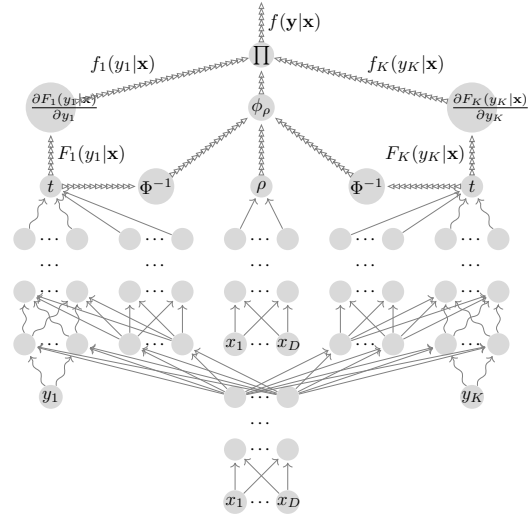}}
\caption{Multivariate Monotonic Neural Density Estimator with Gaussian
  Copula Dependency and Parameterized Covariance.}
\label{figure:monde_mv_param_copula}
\end{figure}

Figure \ref{figure:monde_mv_param_copula} contains a diagram of the
MONDE Copula model with parameterized covariance. It differs from the
MONDE Copula Constant Covariance only by an additional neural network
which encodes the covariance of the output vector. The neural network maps
covariate vector $\mathbf{x}$ into vectors $\mathbf{u}(\mathbf{x})$
and $\mathbf{d}(\mathbf{x})$ which are then used to construct a correlation
matrix as shown in Equation
\ref{eq:monde_copul:cor_cov_parametrization}.

\subsubsection{Details}

\begin{figure}
\centering
\scalebox{.7}{\def\layersep{1}

\begin{tikzpicture}[draw=black!50, node distance=\layersep]
            
    \tikzstyle{every pin edge}=[<-,shorten <=1pt]
    \tikzstyle{positive}=[snake=coil,segment aspect=0,segment amplitude=1pt];
    \tikzstyle{neuron}=[circle,fill=gray!30,minimum size=12pt,inner sep=0pt]
    \tikzstyle{input x neuron}=[neuron, fill=gray!30,text width=4mm,align=center];
    \tikzstyle{input y neuron}=[neuron, fill=gray!30,text width=4mm,align=center];
    \tikzstyle{cdf neuron}=[neuron, fill=gray!30];
    \tikzstyle{pdf neuron}=[neuron, fill=gray!30,text width=8mm,align=center];
    \tikzstyle{hidden neuron}=[neuron, fill=gray!30,text width=4mm,align=center];
    \tikzstyle{annot} = [text width=4em, text centered]
    \tikzstyle{arrow}=[shorten >=1pt,->]

    \node[input x neuron] (x1) at (0,0) {$x_1$};
    \node[input x neuron] (xd) at (\layersep,0) {$x_D$};    
    \node at ($(x1)!.5!(xd)$) {\ldots};
    
    \node[hidden neuron] (xh1) at (0,\layersep) {};
    \node[hidden neuron] (xhd) at (\layersep,\layersep) {};
    \node (xh_dots) at ($(xh1)!.5!(xhd)$) {\ldots};
    
    \path[arrow] (x1) edge (xh1);
    \path[arrow] (x1) edge (xhd);
    \path[arrow] (xd) edge (xh1);
    \path[arrow] (xd) edge (xhd);
    
    \node[hidden neuron] (xh_out1) at (0,2*\layersep) {};
    \node[hidden neuron] (xh_outd) at (\layersep,2*\layersep) {};
    \node (xh_out_dots) at ($(xh_out1)!.5!(xh_outd)$) {\ldots};
    
    \node at ($(xh_dots)!.5!(xh_out_dots)$) {\ldots};
    
	
    \node[input y neuron] (y_1) at (-\layersep,2*\layersep) {$y_1$};    
    
    \node[hidden neuron] (xy_1_1) at (-0.5,3*\layersep) {};
    	\node[hidden neuron] (xy_1_d) at (-1.5,3*\layersep) {};
    	\node (xy_1_dots) at ($(xy_1_1)!.5!(xy_1_d)$) {\ldots};
    	
    \path[arrow] (xh_out1) edge (xy_1_1);
    \path[arrow] (xh_out1) edge (xy_1_d);
    \path[arrow] (xh_outd) edge (xy_1_1);
    \path[arrow] (xh_outd) edge (xy_1_d);
    \draw[arrow,positive]  (y_1) -- (xy_1_1);
    \draw[arrow,positive] (y_1) -- (xy_1_d);
    
    \node[hidden neuron] (xy_1_h1_1) at (-0.5,4*\layersep) {};
    	\node[hidden neuron] (xy_1_h1_d) at (-1.5,4*\layersep) {};
    	\node (xy_1_h1_dots) at ($(xy_1_h1_1)!.5!(xy_1_h1_d)$) {\ldots};
    	
    \draw[positive,arrow] (xy_1_1) -- (xy_1_h1_1);
    \draw[positive,arrow] (xy_1_1) -- (xy_1_h1_d);
    \draw[positive,arrow] (xy_1_d) -- (xy_1_h1_1);
    \draw[positive,arrow] (xy_1_d) -- (xy_1_h1_d);
    	
    \node[hidden neuron] (xy_1_h2_1) at (-0.5,5*\layersep) {};
    	\node[hidden neuron] (xy_1_h2_d) at (-1.5,5*\layersep) {};
    	\node (xy_1_h2_dots) at ($(xy_1_h2_1)!.5!(xy_1_h2_d)$) {\ldots};
     
    \node at ($(xy_1_h1_dots)!.5!(xy_1_h2_dots)$) {\ldots};
    
    \node[cdf neuron] (cdf_1) at (-2,6*\layersep) {$t$};
    
    \draw[positive,arrow] (xy_1_h2_1) -- (cdf_1);
    \draw[positive,arrow] (xy_1_h2_d) -- (cdf_1);
    
    \node[pdf neuron] (pdf_1) at (-2,7.5*\layersep) {$\frac{\partial{F_1(y_1|\mathbf{x})}}{\partial{y_1}}$};
    
    \draw[snake=triangles,segment object length=3pt, segment length=3pt] (cdf_1) -- node[right]{$F_1(y_1|\mathbf{x})$} (pdf_1);
    
    	 \node[cdf neuron] (quantile_1) at (-0.5,6*\layersep) {$\Phi^{-1}$};
    	 
    	 \draw[snake=triangles,segment object length=3pt, segment length=3pt] (cdf_1) -- (quantile_1);

    \node[input y neuron] (y_k) at (2*\layersep,2*\layersep) {$y_K$};
    
    \node[hidden neuron] (xy_k_1) at (1.5,3*\layersep) {};
    	\node[hidden neuron] (xy_k_d) at (2.5,3*\layersep) {};
    	\node (xy_k_dots) at ($(xy_k_1)!.5!(xy_k_d)$) {\ldots};
    	
    \path[arrow] (xh_out1) edge (xy_k_1);
    \path[arrow] (xh_out1) edge (xy_k_d);
    \path[arrow] (xh_outd) edge (xy_k_1);
    \path[arrow] (xh_outd) edge (xy_k_d);
    \draw[arrow,positive]  (y_k) -- (xy_k_1);
    \draw[arrow,positive] (y_k) -- (xy_k_d);
    
    \node[hidden neuron] (xy_k_h1_1) at (1.5,4*\layersep) {};
    	\node[hidden neuron] (xy_k_h1_d) at (2.5,4*\layersep) {};
    	\node (xy_k_h1_dots) at ($(xy_k_h1_1)!.5!(xy_k_h1_d)$) {\ldots};
    	
    \draw[positive,arrow] (xy_k_1) -- (xy_k_h1_1);
    \draw[positive,arrow] (xy_k_1) -- (xy_k_h1_d);
    \draw[positive,arrow] (xy_k_d) -- (xy_k_h1_1);
    \draw[positive,arrow] (xy_k_d) -- (xy_k_h1_d);
    	
    \node[hidden neuron] (xy_k_h2_1) at (1.5,5*\layersep) {};
    	\node[hidden neuron] (xy_k_h2_d) at (2.5,5*\layersep) {};
    	\node (xy_k_h2_dots) at ($(xy_k_h2_1)!.5!(xy_k_h2_d)$) {\ldots};
     
    \node at ($(xy_k_h1_dots)!.5!(xy_k_h2_dots)$) {\ldots};
    
    \node[cdf neuron] (cdf_k) at (3,6*\layersep) {$t$};
    
    \draw[positive,arrow] (xy_k_h2_1) -- (cdf_k);
    \draw[positive,arrow] (xy_k_h2_d) -- (cdf_k);
    
    \node[pdf neuron] (pdf_k) at (3,7.5*\layersep) {$\frac{\partial{F_K(y_K|\mathbf{x})}}{\partial{y_K}}$};
    
    \draw[snake=triangles,segment object length=3pt, segment length=3pt] (cdf_k) -- node[left]{$F_K(y_K|\mathbf{x})$} (pdf_k);
     
    	 \node[cdf neuron] (quantile_k) at (1.5,6*\layersep) {$\Phi^{-1}$};
    	 
    	 \draw[snake=triangles,segment object length=3pt, segment length=3pt] (cdf_k) -- (quantile_k);

	
	\node[cdf neuron] (correlation) at (0.5,6*\layersep) {$\mathbf{\rho}$};
	
    	 \draw[snake=triangles,segment object length=3pt, segment length=3pt] (quantile_1) -- (correlation);
    	 \draw[snake=triangles,segment object length=3pt, segment length=3pt] (quantile_k) -- (correlation);
    	 
    	 \node[cdf neuron] (copula) at (0.5,7.5*\layersep) {$\phi_{\mathbf{\rho}}$};
    	 
    	 \draw[snake=triangles,segment object length=3pt, segment length=3pt] (quantile_1) -- (copula);
    	 \draw[snake=triangles,segment object length=3pt, segment length=3pt] (quantile_k) -- (copula);
    	 \draw[snake=triangles,segment object length=3pt, segment length=3pt] (correlation) -- (copula);
    	 
    	 \node[cdf neuron] (prod) at (0.5,8.5*\layersep) {$\prod$};
    	 
    	 \draw[snake=triangles,segment object length=3pt, segment length=3pt] (copula) -- (prod);
    	 \draw[snake=triangles,segment object length=3pt, segment length=3pt] (pdf_1) -- node[above left]{$f_1(y_1|\mathbf{x})$} (prod);
    	 \draw[snake=triangles,segment object length=3pt, segment length=3pt] (pdf_k) -- node[above right]{$f_K(y_K|\mathbf{x})$} (prod);
    	 
    	  \draw[snake=triangles,segment object length=3pt, segment length=3pt] (prod) -- node[right]{$f(\mathbf{y}|\mathbf{x})$} (0.5, 9.5*\layersep);

\end{tikzpicture}}
\caption{Multivariate Monotonic Neural Density Estimator with Gaussian Copula Dependency and Constant Covariance, model without two partitions for each marginal distribution.}
\label{figure:monde_mv_constant_copula_org}
\end{figure}

\begin{figure}
\centering
\scalebox{.7}{\def\layersep{1}

\begin{tikzpicture}[draw=black!50, node distance=\layersep]
            
    \tikzstyle{every pin edge}=[<-,shorten <=1pt]
    \tikzstyle{positive}=[snake=coil,segment aspect=0,segment amplitude=1pt];
    \tikzstyle{neuron}=[circle,fill=gray!30,minimum size=12pt,inner sep=0pt]
    \tikzstyle{input x neuron}=[neuron, fill=gray!30,text width=4mm,align=center];
    \tikzstyle{input y neuron}=[neuron, fill=gray!30,text width=4mm,align=center];
    \tikzstyle{cdf neuron}=[neuron, fill=gray!30];
    \tikzstyle{pdf neuron}=[neuron, fill=gray!30,text width=8mm,align=center];
    \tikzstyle{hidden neuron}=[neuron, fill=gray!30,text width=4mm,align=center];
    \tikzstyle{annot} = [text width=4em, text centered]
    \tikzstyle{arrow}=[shorten >=1pt,->]

    \node[input x neuron] (x1) at (0,0) {$x_1$};
    \node[input x neuron] (xd) at (\layersep,0) {$x_D$};    
    \node at ($(x1)!.5!(xd)$) {\ldots};
    
    \node[hidden neuron] (xh1) at (0,\layersep) {};
    \node[hidden neuron] (xhd) at (\layersep,\layersep) {};
    \node (xh_dots) at ($(xh1)!.5!(xhd)$) {\ldots};
    
    \path[arrow] (x1) edge (xh1);
    \path[arrow] (x1) edge (xhd);
    \path[arrow] (xd) edge (xh1);
    \path[arrow] (xd) edge (xhd);
    
    \node[hidden neuron] (xh_out1) at (0,2*\layersep) {};
    \node[hidden neuron] (xh_outd) at (\layersep,2*\layersep) {};
    \node (xh_out_dots) at ($(xh_out1)!.5!(xh_outd)$) {\ldots};
    
    \node at ($(xh_dots)!.5!(xh_out_dots)$) {\ldots};
    
	
    \node[input y neuron] (y_1) at (-2*\layersep,2*\layersep) {$y_1$};    
    
    \node[hidden neuron] (xy_1_1) at (-1.5,3*\layersep) {};
    	\node[hidden neuron] (xy_1_d) at (-2.5,3*\layersep) {};
    	\node (xy_1_dots) at ($(xy_1_1)!.5!(xy_1_d)$) {\ldots};
    	
    \path[arrow] (xh_out1) edge (xy_1_1);
    \path[arrow] (xh_out1) edge (xy_1_d);
    \path[arrow] (xh_outd) edge (xy_1_1);
    \path[arrow] (xh_outd) edge (xy_1_d);
    \draw[arrow,positive]  (y_1) -- (xy_1_1);
    \draw[arrow,positive] (y_1) -- (xy_1_d);
    
    \node[hidden neuron] (xy_1_h1_1) at (-1.5,4*\layersep) {};
    	\node[hidden neuron] (xy_1_h1_d) at (-2.5,4*\layersep) {};
    	\node (xy_1_h1_dots) at ($(xy_1_h1_1)!.5!(xy_1_h1_d)$) {\ldots};
    	
    \draw[positive,arrow] (xy_1_1) -- (xy_1_h1_1);
    \draw[positive,arrow] (xy_1_1) -- (xy_1_h1_d);
    \draw[positive,arrow] (xy_1_d) -- (xy_1_h1_1);
    \draw[positive,arrow] (xy_1_d) -- (xy_1_h1_d);
    	
    \node[hidden neuron] (xy_1_h2_1) at (-1.5,5*\layersep) {};
    	\node[hidden neuron] (xy_1_h2_d) at (-2.5,5*\layersep) {};
    	\node (xy_1_h2_dots) at ($(xy_1_h2_1)!.5!(xy_1_h2_d)$) {\ldots};
     
    \node at ($(xy_1_h1_dots)!.5!(xy_1_h2_dots)$) {\ldots};
    
    \node[cdf neuron] (cdf_1) at (-2,6*\layersep) {$t$};
    
    \draw[positive,arrow] (xy_1_h2_1) -- (cdf_1);
    \draw[positive,arrow] (xy_1_h2_d) -- (cdf_1);
    
    \node[pdf neuron] (pdf_1) at (-2,7.5*\layersep) {$\frac{\partial{F_1(y_1|\mathbf{x})}}{\partial{y_1}}$};
    
    \draw[snake=triangles,segment object length=3pt, segment length=3pt] (cdf_1) -- node[right]{$F_1(y_1|\mathbf{x})$} (pdf_1);
    
    	 \node[cdf neuron] (quantile_1) at (-0.5,6*\layersep) {$\Phi^{-1}$};
    	 
    	 \draw[snake=triangles,segment object length=3pt, segment length=3pt] (cdf_1) -- (quantile_1);

    \node[input y neuron] (y_k) at (3*\layersep,2*\layersep) {$y_K$};
    
    \node[hidden neuron] (xy_k_1) at (2.5,3*\layersep) {};
    	\node[hidden neuron] (xy_k_d) at (3.5,3*\layersep) {};
    	\node (xy_k_dots) at ($(xy_k_1)!.5!(xy_k_d)$) {\ldots};
    	
    \path[arrow] (xh_out1) edge (xy_k_1);
    \path[arrow] (xh_out1) edge (xy_k_d);
    \path[arrow] (xh_outd) edge (xy_k_1);
    \path[arrow] (xh_outd) edge (xy_k_d);
    \draw[arrow,positive]  (y_k) -- (xy_k_1);
    \draw[arrow,positive] (y_k) -- (xy_k_d);
    
    \node[hidden neuron] (xy_k_h1_1) at (2.5,4*\layersep) {};
    	\node[hidden neuron] (xy_k_h1_d) at (3.5,4*\layersep) {};
    	\node (xy_k_h1_dots) at ($(xy_k_h1_1)!.5!(xy_k_h1_d)$) {\ldots};
    	
    \draw[positive,arrow] (xy_k_1) -- (xy_k_h1_1);
    \draw[positive,arrow] (xy_k_1) -- (xy_k_h1_d);
    \draw[positive,arrow] (xy_k_d) -- (xy_k_h1_1);
    \draw[positive,arrow] (xy_k_d) -- (xy_k_h1_d);
    	
    \node[hidden neuron] (xy_k_h2_1) at (2.5,5*\layersep) {};
    	\node[hidden neuron] (xy_k_h2_d) at (3.5,5*\layersep) {};
    	\node (xy_k_h2_dots) at ($(xy_k_h2_1)!.5!(xy_k_h2_d)$) {\ldots};
     
    \node at ($(xy_k_h1_dots)!.5!(xy_k_h2_dots)$) {\ldots};
    
    \node[cdf neuron] (cdf_k) at (3,6*\layersep) {$t$};
    
    \draw[positive,arrow] (xy_k_h2_1) -- (cdf_k);
    \draw[positive,arrow] (xy_k_h2_d) -- (cdf_k);
    
    \node[pdf neuron] (pdf_k) at (3,7.5*\layersep) {$\frac{\partial{F_K(y_K|\mathbf{x})}}{\partial{y_K}}$};
    
    \draw[snake=triangles,segment object length=3pt, segment length=3pt] (cdf_k) -- node[left]{$F_K(y_K|\mathbf{x})$} (pdf_k);
     
    	 \node[cdf neuron] (quantile_k) at (1.5,6*\layersep) {$\Phi^{-1}$};
    	 
    	 \draw[snake=triangles,segment object length=3pt, segment length=3pt] (cdf_k) -- (quantile_k);
    	 
    \node[input x neuron] (x1_cov) at (0,3*\layersep) {$x_1$};
    \node[input x neuron] (xd_cov) at (\layersep,3*\layersep) {$x_D$};    
    \node at ($(x1_cov)!.5!(xd_cov)$) {\ldots};
    
    \node[hidden neuron] (xh1_cov) at (0,4*\layersep) {};
    \node[hidden neuron] (xhd_cov) at (\layersep,4*\layersep) {};
    \node (xh_dots_cov) at ($(xh1_cov)!.5!(xhd_cov)$) {\ldots};
    
    \path[arrow] (x1_cov) edge (xh1_cov);
    \path[arrow] (x1_cov) edge (xhd_cov);
    \path[arrow] (xd_cov) edge (xh1_cov);
    \path[arrow] (xd_cov) edge (xhd_cov);
    
    \node[hidden neuron] (xh_out1_cov) at (0,5*\layersep) {};
    \node[hidden neuron] (xh_outd_cov) at (\layersep,5*\layersep) {};
    \node (xh_out_dots_cov) at ($(xh_out1_cov)!.5!(xh_outd_cov)$) {\ldots};
    
    \node at ($(xh_dots_cov)!.5!(xh_out_dots_cov)$) {\ldots};
    
    \path[arrow] (xh_out1_cov) edge (correlation);
    \path[arrow] (xh_outd_cov) edge (correlation);
    	 
	
	\node[cdf neuron] (correlation) at (0.5,6*\layersep) {$\mathbf{\rho}$};
	    	 
    	 \node[cdf neuron] (copula) at (0.5,7.5*\layersep) {$\phi_{\mathbf{\rho}}$};
    	 
    	 \draw[snake=triangles,segment object length=3pt, segment length=3pt] (quantile_1) -- (copula);
    	 \draw[snake=triangles,segment object length=3pt, segment length=3pt] (quantile_k) -- (copula);
    	 \draw[snake=triangles,segment object length=3pt, segment length=3pt] (correlation) -- (copula);
    	 
    	 \node[cdf neuron] (prod) at (0.5,8.5*\layersep) {$\prod$};
    	 
    	 \draw[snake=triangles,segment object length=3pt, segment length=3pt] (copula) -- (prod);
    	 \draw[snake=triangles,segment object length=3pt, segment length=3pt] (pdf_1) -- node[above left]{$f_1(y_1|\mathbf{x})$} (prod);
    	 \draw[snake=triangles,segment object length=3pt, segment length=3pt] (pdf_k) -- node[above right]{$f_K(y_K|\mathbf{x})$} (prod);
    	 
    	  \draw[snake=triangles,segment object length=3pt, segment length=3pt] (prod) -- node[right]{$f(\mathbf{y}|\mathbf{x})$} (0.5, 9.5*\layersep);

\end{tikzpicture}}
\caption{Multivariate Monotonic Neural Density Estimator with Gaussian Copula Dependency and Parameterized Covariance, model without two partitions for each marginal distribution.}
\label{figure:monde_mv_param_copula_org}
\end{figure}
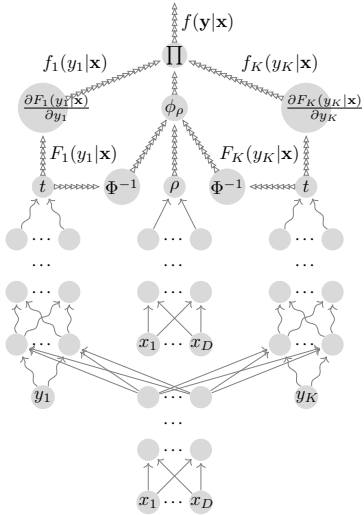

When we were experimenting with MONDE Copula models, we tried
different approaches which were not used in the final implementation
of our algorithms:

\begin{itemize}
\item We tried to parameterize the correlation matrix as proposed by
  \cite{RapisardaFrancesco2007Pcag}. However, we obtained inferior
  results compared to the method presented in Equation
  \ref{eq:monde_copul:cor_cov_parametrization}.
\item We tried to pre-train the models with first fitting each
  univariate marginal of the model to the unconditional empirical
  distributions as given by the data using the mean squared error
  objective, followed by maximizing likelihood objective given by the
  entire copula model. This method also did not bring any improvements
  in convergence speed nor better final results. To keep things simple, we drop it
  from the final implementation.
\item Initially, our multivariate copula models did not have two
  vertical partitions for each marginal distribution (the part of the
  computational graph that computes $t_i(y_i,\mathbf{x})$). The
  initial models without two partitions are shown in Figure
  \ref{figure:monde_mv_constant_copula_org} and
  \ref{figure:monde_mv_param_copula_org} for constant and parametrized
  covariance respectively. The final models are presented in Figure
  \ref{figure:monde_mv_constant_copula} and Figure
  \ref{figure:monde_mv_param_copula}. This extension is constructed in
  a way so that the monotonicity of $t_i$ with respect to $y_i$ is not
  destroyed i.e. we allow connections with unconstrained weights from
  the covariate partition into the partition processing the response
  variable but not the other way around. These additional connections
  in each of the $t_i$ transformations allowed us to obtain better
  results.
\end{itemize}

\subsection{PUMONDE}

\subsubsection{Architecture Justification}

In this section, we show why we chose the proposed computational graph
for PUMONDE model. For simplicity of exposition, we focus on the
bivariate case, but the explanation applies to the general
multivariate case.

To obtain a valid distribution function $F(y_1,y_2|\mathbf{x})$ and
the corresponding density function $f(y_1,y_2|\mathbf{x})$, we need to
constrain our neural network to meet following conditions:
\begin{enumerate}
    \item\label{pumonde:limit1} $\lim_{y_1 \rightarrow -\infty,y_2 \rightarrow -\infty} F(y_1,y_2|x) =0$
    \item\label{pumonde:limit2} $\lim_{y_1 \rightarrow +\infty,y_2 \rightarrow +\infty} F(y_1,y_2|x) = 1$
    \item\label{pumonde:mix_deriv_positive} $\frac{\partial^2F}{\partial y_1 \partial y_2 } \in \mathbb{R}^+$
    \item\label{pumonde:mono_deriv_positive} $\frac{\partial^2F}{\partial y_k^2} \in \mathbb{R}$ for $k=1,2$
\end{enumerate}

To meet constraints \ref{pumonde:limit1} and \ref{pumonde:limit2}, we
parameterize the distribution function as the ratio of two non-negative
monotonic functions,
\[
F_w(y_1,y_2~|~\mathbf{x})=\frac{t(m(h_{xy1}(y_1,h_x(x)), h_{xy2}(y_2,h_x(x))))}{t(1,1)}.
\]

We chose all the transforms of $t(\cdot)$ to be the softplus function,
which we will denote with the symbol $s_+$ for the rest of this
section. There are numerous reasons for this choice. First, $s_+$ meets
the requirement for the output to be non-negative and monotonically
increasing with respect to its input.  It is required by the
monotonicity property of the computational graph to use monotonic
transformations for all computations transforming response variables.
We also used $s_+$ and not, for example, $\tanh$ because the second
order derivative of $s_+$ with respect to its input is positive (the
softplus function is convex throughout all of its domain). The second
order derivative of $\tanh$ with respect to its input can be negative,
which breaks constraint \ref{pumonde:mix_deriv_positive}.  The
multiplication layer (here symbolically written as $m(\cdot)$), which
can be seen in the middle of the computation graph, receives
positive valued transformations of $y_1$ and $y_2$ ($h_{xyi}(\cdot)$
uses only sigmoid non-linearities). This composition allows us to
fulfil constraint \ref{pumonde:mix_deriv_positive}. This comes from
the fact that
 \begin{align*}
  &\displaystyle \frac{\partial^2 s_{+}( \sigma(g_1(y_1)) \times \sigma(g_2(y_2)))}{\partial y_1 \partial y_2}=\\
  &\displaystyle \frac{\partial s_{+}^{'}(\cdot) \times \sigma(g_1(y_1)) \times \sigma^{'}(g_2(y_2)) \times g_2^{'}(y_2)}{\partial y_1} =\\
  &\displaystyle [ s_{+}^{''}(\cdot) \times \sigma(g_2(y_2)) \times  \sigma(g_1(y_1)) +  s_{+}^{'}(\cdot) ] \times \\
  &\displaystyle \sigma^{'}(g_1(y_1)) \times g_1^{'}(y_1)  \times \sigma^{'}(g_2(y_2)) \times g_2^{'}(y_2),
\end{align*}
where $\sigma(g_i(\cdot))=h_{xyi}(\cdot)$ and $g_i$ is the monotonic
transformation with respect of $y_i$ using only $\sigmoid$s which
contains all but the last transformation of $h_{xyi}(\cdot)$. Symbol
$s_{+}^{'}(\cdot)$ denotes the derivative of the function with respect
to its input from the previous expression in the equation.

Therefore, that last expression always evaluates to a non-negative
number. We see that, if we used $\sigmoid$ or $\tanh$ in the layers
following the multiplication layer ($m(\cdot)$), we would not be
guaranteed to obtain a non-negative result. Using the sigmoid
activation function before the multiplication layer still allows to
meet constraint \ref{pumonde:mono_deriv_positive} because the second
derivative of $\sigma$ with respect to its input can be positive and
negative (because this function is convex and concave depending on the
input).  It is also imperative that the inputs into the multiplication
layer (outputs of the $h_{xyi}(\cdot)$s transformations) of PUMONDE is
non-negative, because otherwise it would break the monotonicity
property of the model with respect to response variables.

\subsubsection{Implementation Remarks}

We found that using the vanilla softplus activation can lead to numerical
problems in models which have many layers (we found that the minimum
number of layers at which the computation resulted in numerical
problems also is dependent on dataset itself). After running several
experiments, we concluded that the most numerically stable
approximation of the softplus is $\log{(1+\exp{(-|x|)})}+\max(x,0)$.

\subsection{COMPUTATIONAL COMPLEXITY}

Applying auto differentiation twice on the univariate response MONDE
(with respect to inputs and then with respect to parameters) is more
costly by a constant factor than applying it once in the model
parameterizing the pdf directly. The training of the multivariate
estimators like PUMONDE mirrors the problem with the exact inference
in dense Markov Networks i.e. it scales exponentially in the
dimensionality of the problem. We mitigate this issue by exploiting
structure in the Gaussian copula and by the use of composite
likelihood. In follow-up work, we want to adapt our method to
structured CDFs which were introduced in \cite{Huang:2008}.  PUMONDE
can also motivate future message-passing views of autodiff
\cite{Minka:2019}.

\section{BASELINE MODELS IMPLEMENTATION}
We found that the implementation of models using mixture components
like RNADE and MDN requires a few tricks to make them obtain good
results. First, we needed to tweak the minimum value allowed for the
scale parameters of the Gaussian components. If we allow it to be
arbitrarily small, the models frequently put a lot of weight on one
mixture component which has very high precision, particularly on
finite-resolution data in which repeated values occur to some
extent. This artificially inflated the average log-likelihood,
sometimes by a large amount. This is a well known problem with mixture
models. We checked that to achieve the best performance of such
models, the minimum allowed value of the mixture weight would have to
be adjusted for each dataset separately. In practice, we set this
threshold to be the smallest one not causing the optimization process
to misbehave, taking into consideration the already large dimension of
the hyperparameter space we optimize over.

\section{EXPERIMENTS}
\label{sup:sec:experiments}

\subsection{EXPERIMENTS SETUP}

We split each dataset into train, validation and test
partitions. Models are only trained on the train partition.
Hyperparameters are chosen by selecting the best model with respect to
the log-likelihood computed on the validation set. The search is done
via exhaustive search over a predefined grid of hyperparameters. Table
\ref{table:hyper_param_grid} shows the search space that was used for
different models and experiments/datasets.
We use the early stopping technique on the
validation set to prevent overfitting with patience of 30 epochs
i.e. we stop training when the likelihood on validation dataset does not
improve for 30 consecutive epochs. The best model on the validation
set is chosen and log-likelihoods of test points are computed. We
compare the models performance by running pairwise t-tests on these
values or use other performance metric adequate for a given
experiment.  We use the ADAM \citep{KingmaDiederik2017AAMf} version of
stochastic gradient descent to optimize the models. We tried different
parameters of the optimizer but we settled with default ones that are
used in the Tensorflow i.e. beta\_1=$0.9$, beta\_2=$0.999$,
epsilon=$1e-07$. The learning rate and the batch size were set as
given in Table \ref{table:hyper_param_grid}.  In some of the
experiments, we used a modified learning process where we
increase the batch size with a pre-defined schedule. We give the rationale
behind using it in Section
\ref{sec_sup:autoregressive:implementation_remarks}.

\subsection{EXPERIMENTS}

All dimensions of the datasets used in our experiments are provided in Table
\ref{table:data_sets_meta}. Experiments described in Sections
\ref{sec:synthetic_data}, \ref{sec:low_dimensional_data} and
\ref{sec:financial_data} have datasets arranged in a way to test the
model performance on the following learning tasks:
1) unconditional/unsupervised: $dim(\mathbf{x})=0$
and $dim(\mathbf{y})>0$, 2) univariate: $dim(\mathbf{x})>0$ and
$dim(\mathbf{y})=1$, 3) small dimensional multivariate:
$dim(\mathbf{x})>0$ and $dim(\mathbf{y})>1$ probability density
functions. In other experiments the dimensionality of the response and
covariate vectors depends on the task performed.

\begin{table}[t]
\tiny
\caption{Datasets dimensions.}
\begin{center}
\begin{tabular}{llll}
\multicolumn{1}{c}{\bf DATA SET} & \multicolumn{1}{c}{\bf OBSERVATIONS} & \multicolumn{1}{c}{\bf X DIM} & \multicolumn{1}{c}{\bf Y DIM} \\
\hline \\
Sin Normal                  &        10000 &           1 &           1 \\
Sin T                       &        10000 &           1 &           1 \\
Inv Sin Normal              &        10000 &           1 &           1 \\
Inv Sin T                   &        10000 &           1 &           1 \\
UCI Redwine 2D              &         1599 &           7 &           2 \\
UCI Redwine Unsupervised    &         1599 &           0 &           9 \\
UCI Whitewine 2D            &         4898 &           1 &           2 \\
UCI Parkinsons 2D           &         5875 &          13 &           2 \\
MV Nonlinear                &        10000 &           1 &           2 \\
UCI Whitewine Unsupervised  &         4898 &           0 &           3 \\
UCI Parkinsons Unsupervised &         5875 &           0 &          15 \\
ETF 1D                      &         1073      &           1 &           1 \\
ETF 2D                      &         1073      &           2 &           2 \\
FX All Assets Predicted     &        28781      &          16 &           8 \\
FX EUR And GBP Predicted    &        28773      &          32 &           2 \\
FX EUR Predicted            &        28749      &          80 &           1 \\
Classification (FX)         &        91910      &          21 &           3 \\
UCI Power                       &      2049280      &           0 &           6 \\
UCI Gas                         &      1052065      &           0 &           8 \\
UCI Hepmass                     &       525123      &           0 &          21 \\
UCI Miniboone                   &        36488      &           0 &          43 \\
UCI Bsds300                     &      1300000      &           0 &          63 \\
Mixture Process             &       100000      &           2 &           3 \\
FX (Bivariate Likelihood)   &        91549      &          21 &          21 \\

\end{tabular}
\end{center}

\label{table:data_sets_meta}
\end{table}

\subsection{SYNTHETIC DATA}
\label{sec:synthetic_data}

\begin{table*}[t]
\tiny
\caption{Synthetic Data - Mean Loglikelihood.}
\begin{center}
\begin{tabular}{lrrrrrrrrrrr}
{} & \makecell{RNADE\\Laplace} & \makecell{RNADE\\Normal} & \makecell{RNADE\\Deep\\Normal} & \makecell{RNADE\\Deep\\Laplace} & \makecell{MONDE\\Const\\Cov} & \makecell{MONDE\\Param\\Cov} & \makecell{MONDE\\AR} & \makecell{PUMONDE} & \makecell{MDN}  & \makecell{True\\LL}\\
\hline \\
\makecell[l]{Sin Normal}& 0.115 & 0.118 & \textbf{0.155 }& 0.130 & 0.136 &  &  &  & 0.134 &  0.176 \\
\makecell[l]{Sin T}& -0.200 & -0.193 & \textbf{-0.194 }& -0.205 & \textbf{-0.178 }&  &  &  & -0.317 &  -0.163 \\
\makecell[l]{Inv Sin Normal}& 0.186 & 0.212 & \textbf{0.253 }& 0.226 & 0.174 &  &  &  & 0.227  &  \\
\makecell[l]{Inv Sin T}& -0.083 & -0.109 & -0.132 & \textbf{-0.063 }& -0.089 &  &  &  & -0.199  &  \\
\makecell[l]{MV Nonlinear}& -6.196 & -6.067 & -5.695 & -5.281 & -5.095 & -5.074 & -5.135 & \textbf{-5.033 }& -5.247 & -4.973 \\
\end{tabular}
\end{center}

\label{table:generated_ll}
\end{table*}

\begin{figure}[t]
\begin{center}
\subfloat[sin normal]{\includegraphics[scale=0.1]{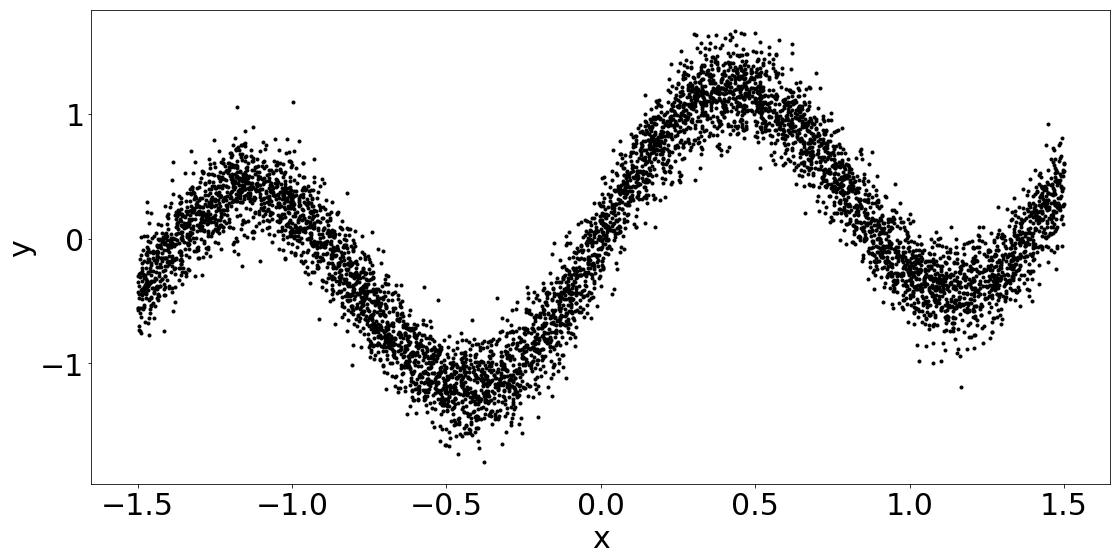}}
\subfloat[sin t]{\includegraphics[scale=0.1]{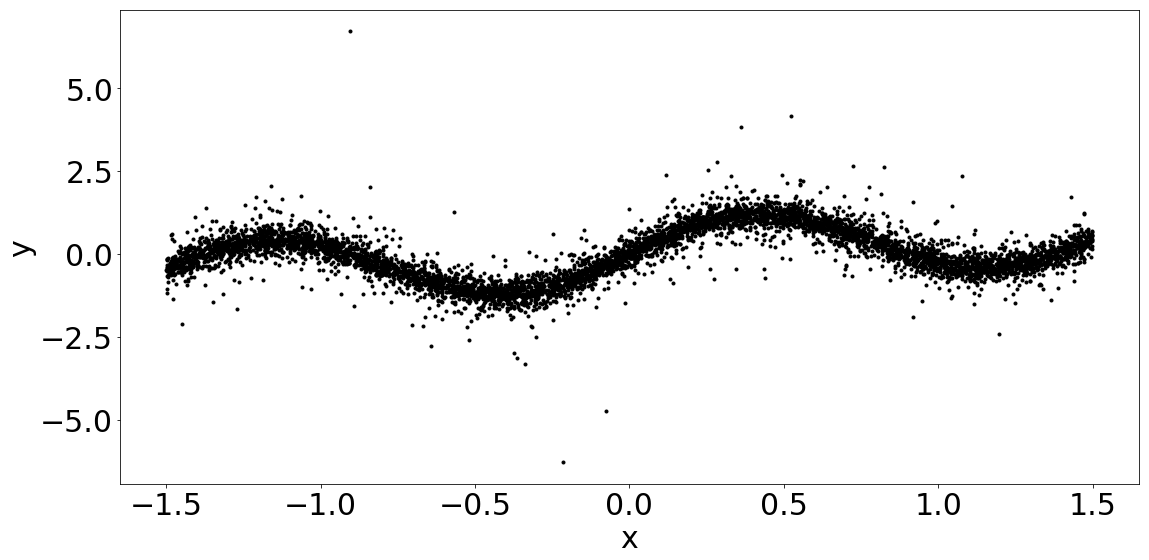}}\\
\subfloat[inv sin normal]{\includegraphics[scale=0.1]{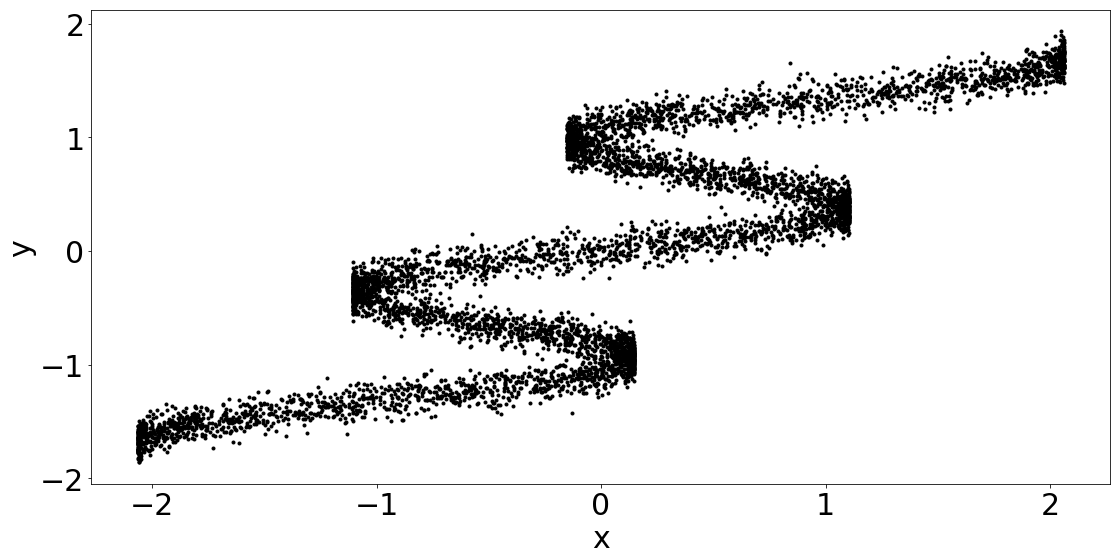}}
\subfloat[inv sin t]{\includegraphics[scale=0.1]{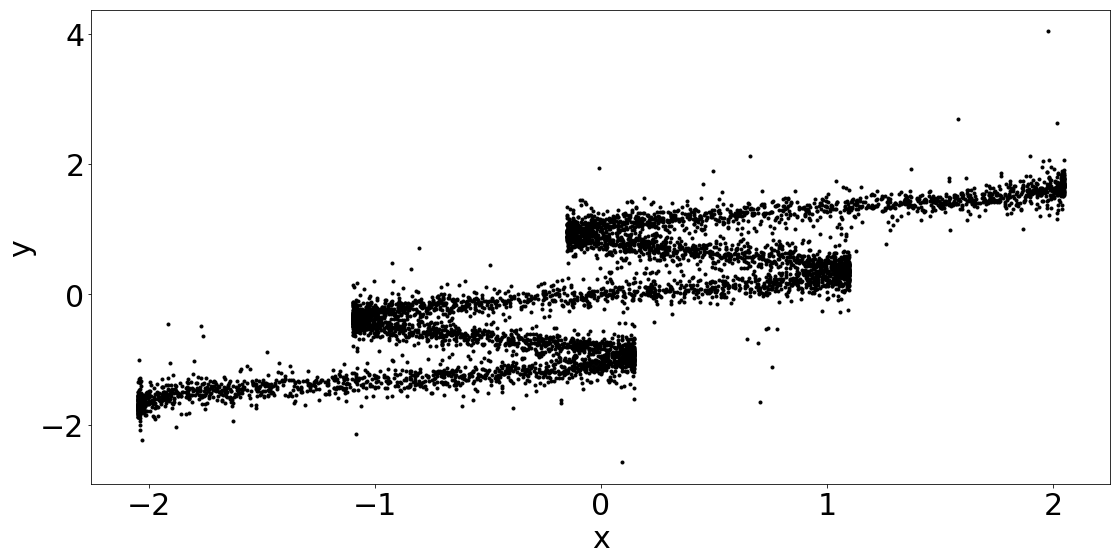}}
\end{center}
\caption{Generated Univariate Response Data.}
\label{figure:generated_data}
\end{figure}

Results from experiments with synthetic data are shown in Table
\ref{table:generated_ll}. The table displays the average
log-likelihood computed on the test partition of the dataset. In each
row, we highlight the best result for the model which achieved
significantly better average log-likelihood than the rest of the
models. We use three datasets in this experiment. The univariate
covariate is generated unconditionally from the uniform $[0, 1]$
distribution for all three datasets. Then, depending on the dataset,
the response variables are sampled from the Gaussian or Student-t
distributions with means parameterized by a sinusoid function with an
input that is a linear transformation of the covariate:
\begin{align*}
X &\sim Uniform(-1.5,1.5)\\
Y &\sim N(\sin(4X)+0.5X, 0.2) \text{ or}\\
Y &\sim t(3, \sin(4X)+0.5X, 0.2).
\end{align*}
We also test the models by swapping the response variable with the
covariate variable, so that we can check whether they can encode the
multimodality in $Y$. Those inverted datasets have "Inv" added to
their names. We show scatter plots of these datasets in Figure
\ref{figure:generated_data}.

\begin{figure}[t]
\begin{center}
\includegraphics[scale=0.2]{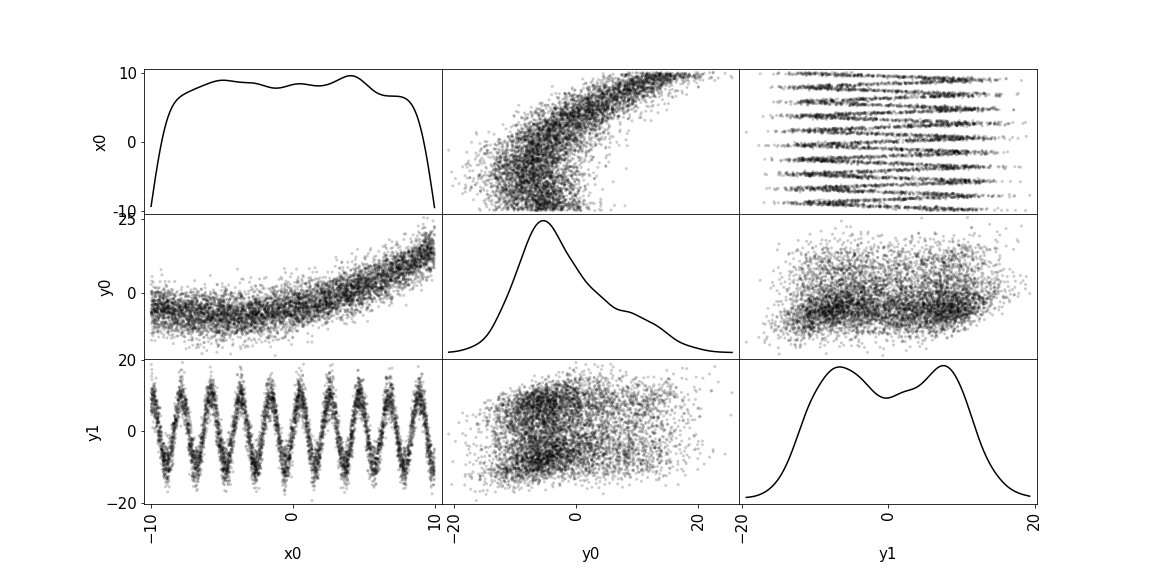}
\end{center}
\caption{Generated Multivariate Response Data.}
\label{figure:generated_mv_nonlinear}
\end{figure}

The next dataset has a bivariate response variable
distributed according to the Gaussian distribution. The mean of each
dimension is parameterized by a different nonlinear function with respect
to the covariate. This bivariate response variable has correlated
dimensions, each of them having a different variance:
\begin{align*}
X &\sim Uniform(-10,10)\\
\Sigma &=
\begin{bmatrix}
4 & 0 \\
0 & 3
\end{bmatrix}
\begin{bmatrix}
1 & 0.7 \\
0.7 & 1
\end{bmatrix}
\begin{bmatrix}
4 & 0 \\
0 & 3
\end{bmatrix}
\\
Y &\sim N([0.1*\sqrt{X}+X-5, 10sin(3X)], \Sigma).
\end{align*}
We present data generated from this random process in Figure
\ref{figure:generated_mv_nonlinear} in a grid, with pairwise
scatter plots off the diagonal and marginal densities on the
diagonal cells.

\begin{figure}[t]
\begin{center}
\subfloat[RNADE N]{\includegraphics[width=2cm,height=2cm,keepaspectratio]{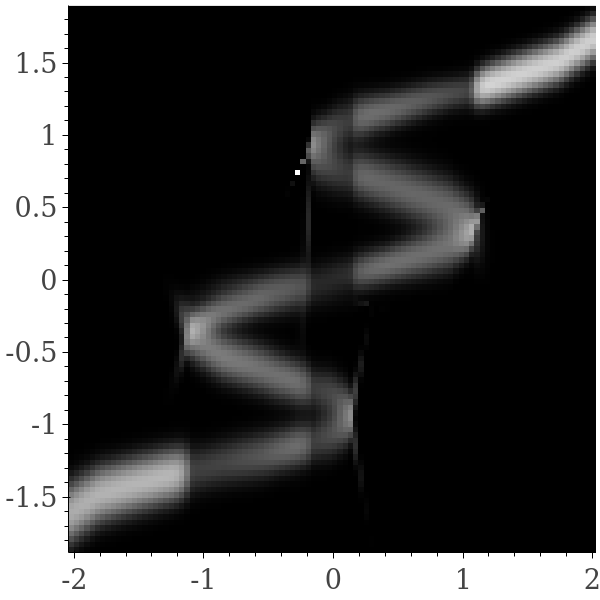}}
\hfill
\subfloat[RNADE L]{\includegraphics[width=2cm,height=2cm,keepaspectratio]{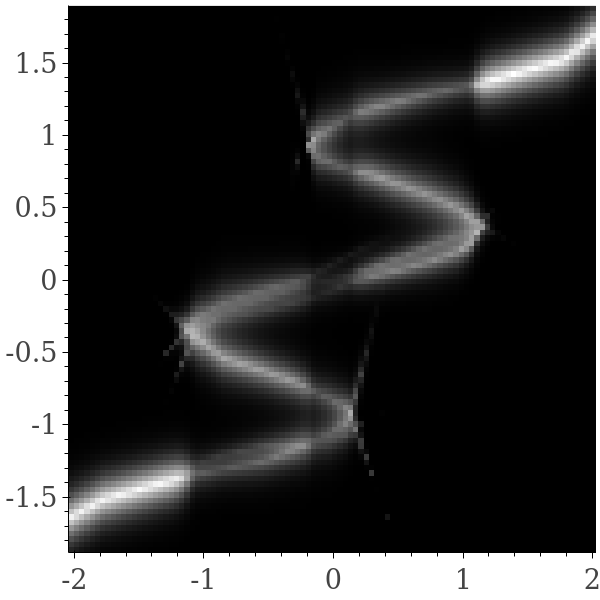}}
\hfill
\subfloat[RNADE DEEP N]{\includegraphics[width=2cm,height=2cm,keepaspectratio]{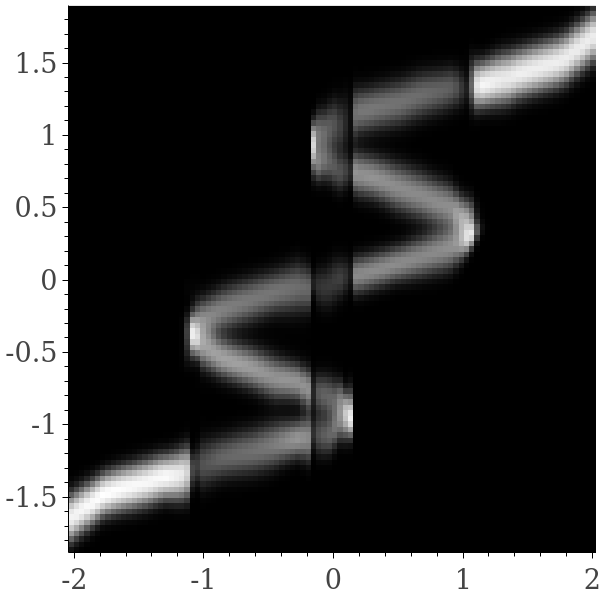}}
\\
\subfloat[RNADE DEEP L]{\includegraphics[width=2cm,height=2cm,keepaspectratio]{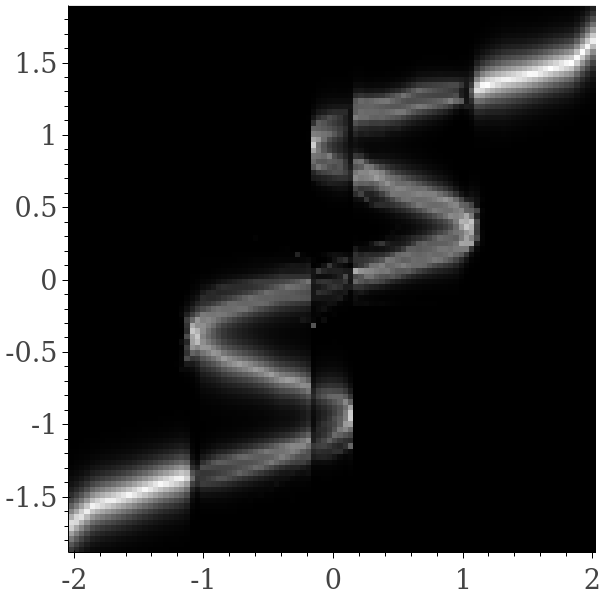}}
\hfill
\subfloat[MDN]{\includegraphics[width=2cm,height=2cm,keepaspectratio]{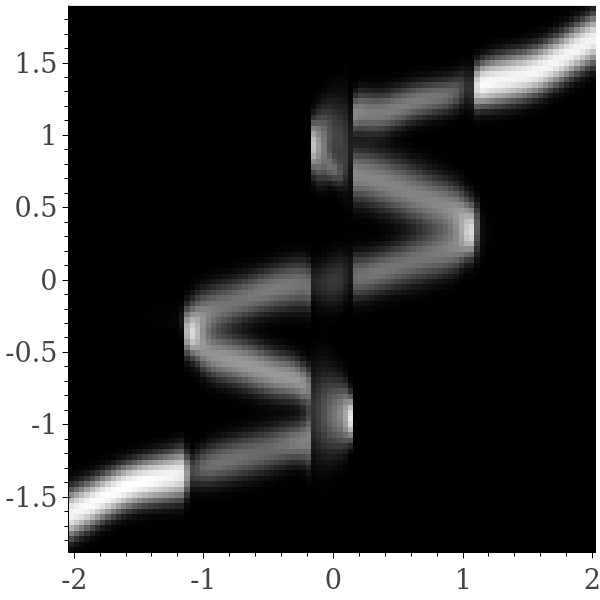}}
\hfill
\subfloat[MONDE]{\includegraphics[width=2cm,height=2cm,keepaspectratio]{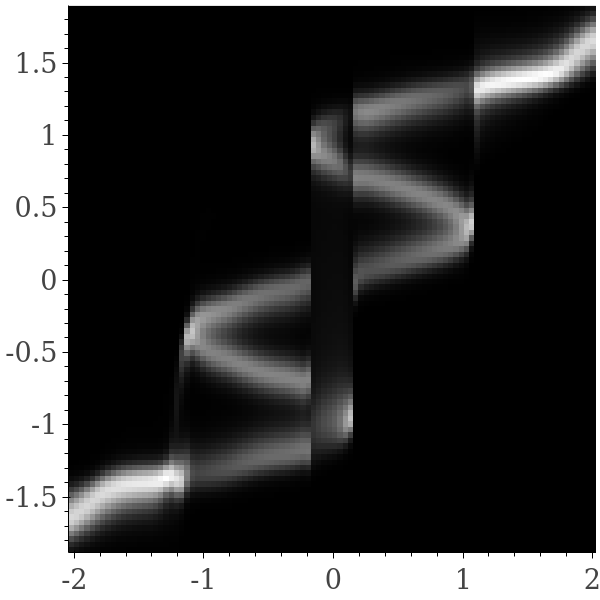}}
\end{center}
\caption{Density Heatmap - Inv Sin Normal.}
\label{figure:density_heatmap_inv_sin}
\end{figure}

Fitting the data by conditional mean models, such as a regular neural
network with mean squared error, is not suitable in the case when we
deal with multimodal output. In this case, we should be using models
that can encode the probabilistic structure of the data. We show in
Figure \ref{figure:density_heatmap_inv_sin} that proposed and baseline
models can capture the multimodality correctly.

\begin{figure}[h]
\subfloat[Gen Distribution]{\includegraphics[width=2.5cm,height=2.5cm,keepaspectratio]{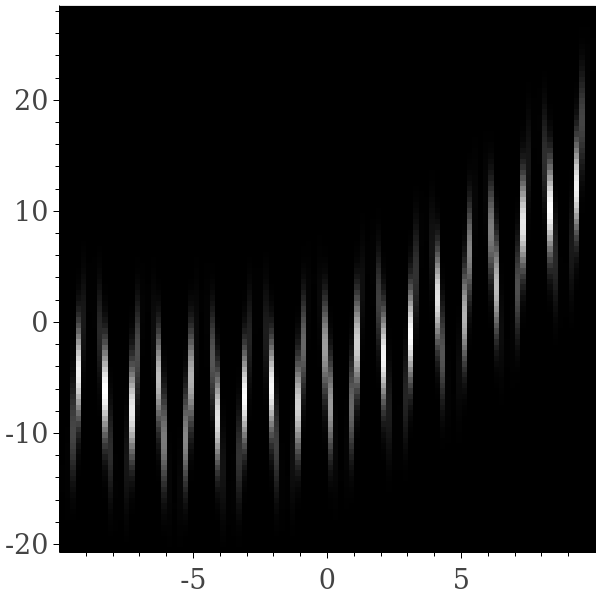}}
\hspace{.01\textwidth}%
\subfloat[RNADE N]{\includegraphics[width=2.5cm,height=2.5cm,keepaspectratio]{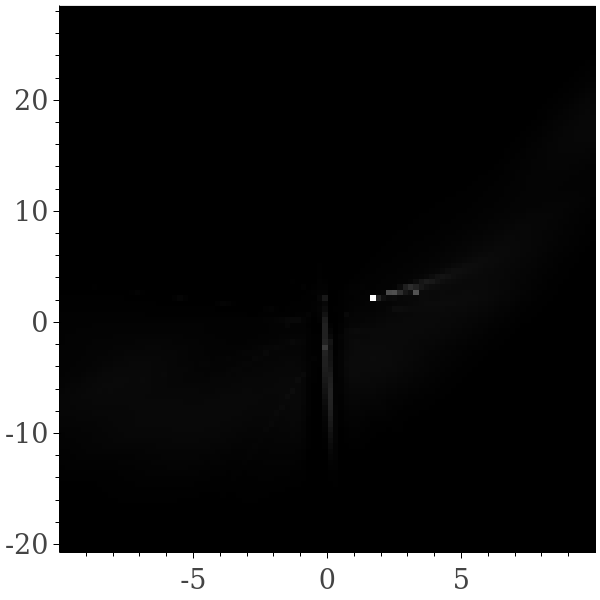}}
\hspace{.01\textwidth}%
\subfloat[RNADE L]{\includegraphics[width=2.5cm,height=2.5cm,keepaspectratio]{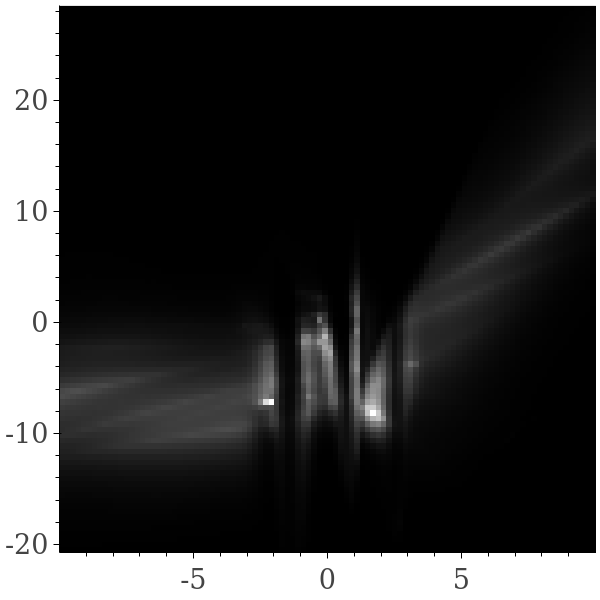}}
\\
\subfloat[RNADE DEEP N]{\includegraphics[width=2.5cm,height=2.5cm,keepaspectratio]{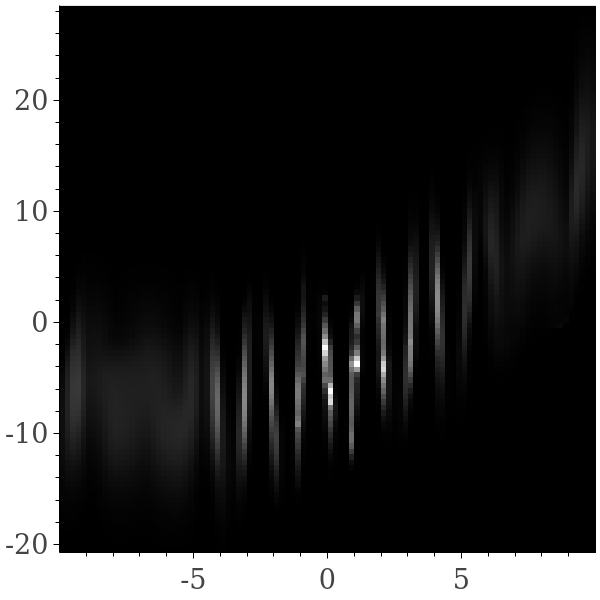}}
\hspace{.01\textwidth}%
\subfloat[RNADE DEEP L]{\includegraphics[width=2.5cm,height=2.5cm,keepaspectratio]{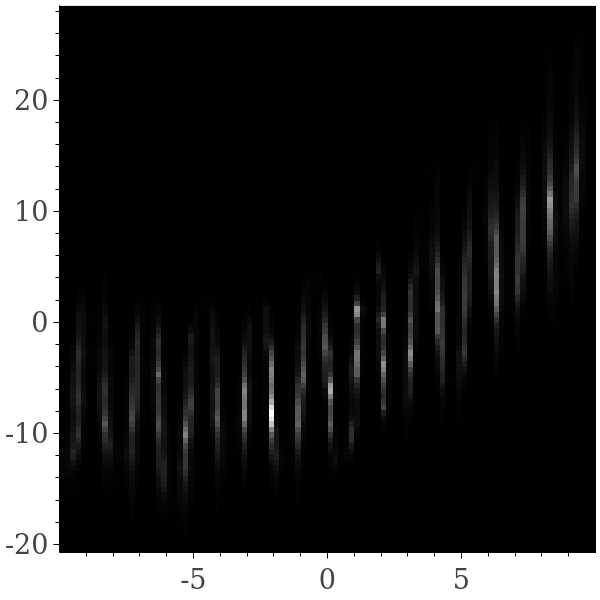}}
\hspace{.01\textwidth}
\subfloat[MDN Const Cov]{\includegraphics[width=2.5cm,height=2.5cm,keepaspectratio]{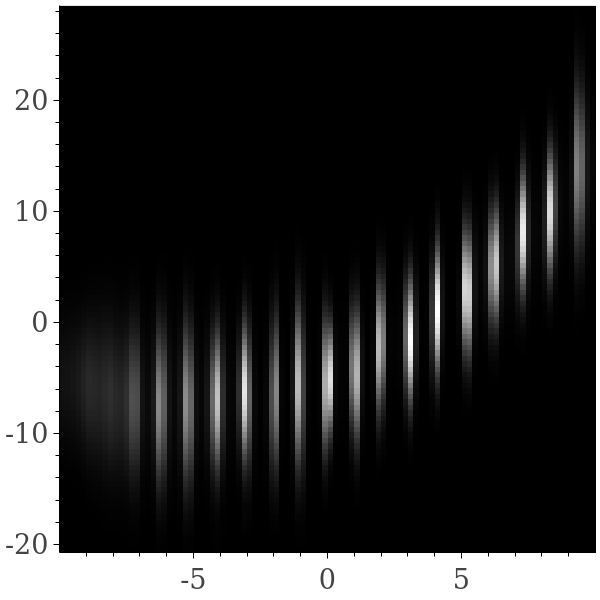}}
\\
\subfloat[MDN Param Cov]{\includegraphics[width=2.5cm,height=2.5cm,keepaspectratio]{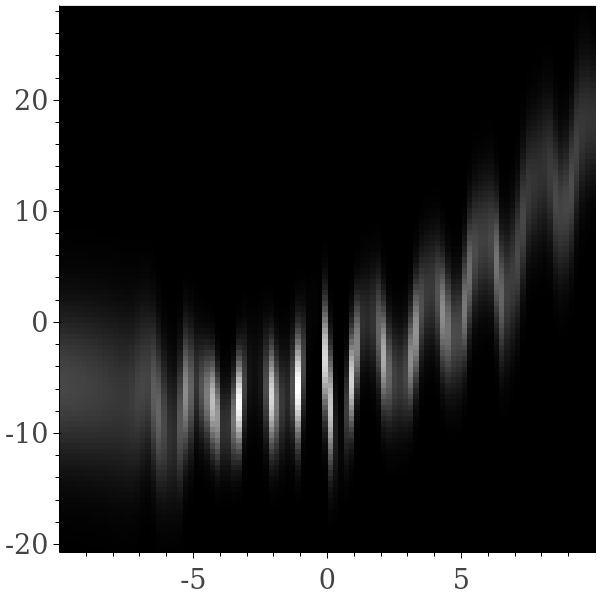}}
\hspace{.01\textwidth}%
\subfloat[MONDE Const Cov]{\includegraphics[width=2.5cm,height=2.5cm,keepaspectratio]{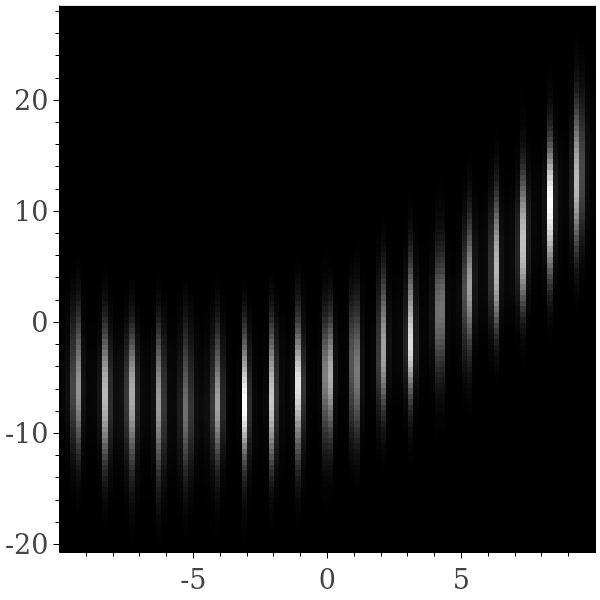}}
\hspace{.01\textwidth}%
\subfloat[MONDE Param Cov]{\includegraphics[width=2.5cm,height=2.5cm,keepaspectratio]{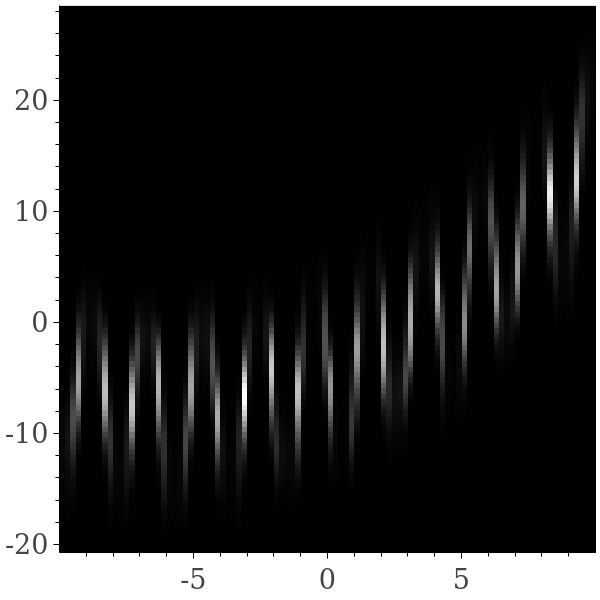}}
\\
\subfloat[MONDE AR]{\includegraphics[width=2.5cm,height=2.5cm,keepaspectratio]{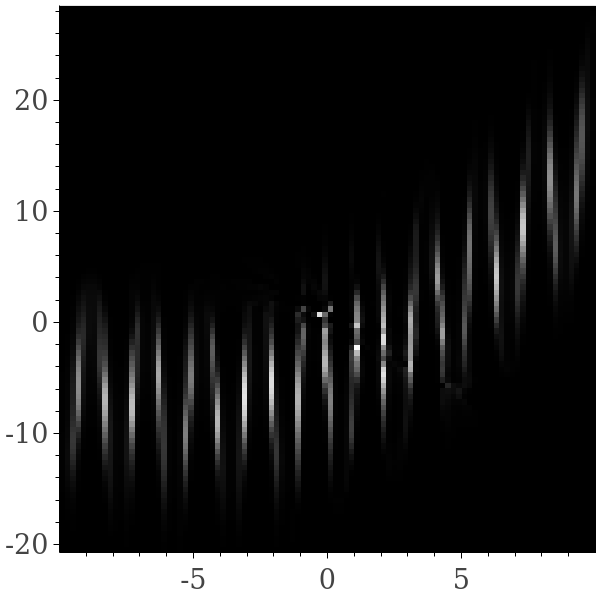}}
\hspace{.01\textwidth}%
\subfloat[PUMONDE]{\includegraphics[width=2.5cm,height=2.5cm,keepaspectratio]{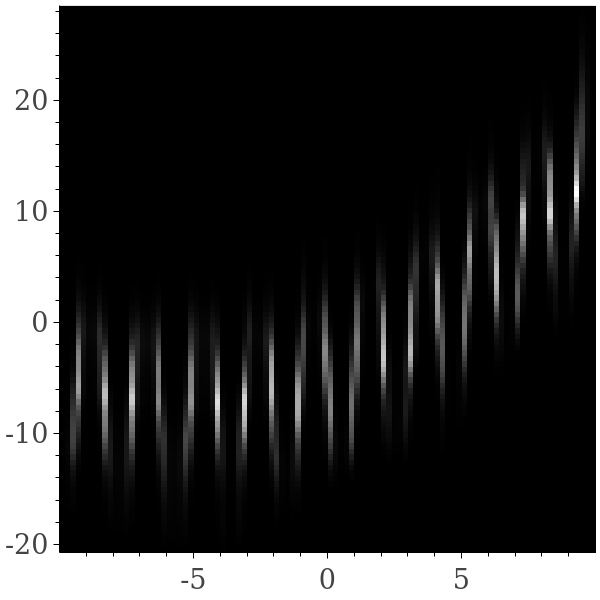}}
\caption{Density Heatmap ($f(y_0, y_1 = 0~|~x)$) - MV Nonlinear.}
\label{figure:density_heatmap_mv}
\end{figure}

In Figure \ref{figure:density_heatmap_mv}, we show density heatmaps
computed from different density estimators and data generating
processes. In these plots, we show density $f(y_0, y_1 = 0~|~x)$ of
baseline and proposed models for a grid of points spanned over $y_0$ and
$x$. We see that RNADE and MDN models have difficulty in encoding the
probabilistic structure, but our models can capture it well, which is
also confirmed by the test log-likelihood evaluations shown in Table
\ref{table:generated_ll}.

\subsection{SMALL UCI DATASETS}
\label{sec:low_dimensional_data}

\begin{table*}[t]
\tiny
\caption{Mean Loglikelihoods - Small UCI Datasets.}
\begin{center}
\begin{tabular}{lrrrrrrrrrr}
{} & \makecell{RNADE\\Laplace} & \makecell{RNADE\\Normal} & \makecell{RNADE\\Deep\\Normal} & \makecell{RNADE\\Deep\\Laplace} & \makecell{MONDE\\Const\\Cov} & \makecell{MONDE\\Param\\Cov} & \makecell{MONDE\\AR} & \makecell{PUMONDE} & \makecell{MDN} \\
\hline
\makecell[l]{UCI Redwine 2D}& -2.543 & -2.506 & -3.310 & -2.343 & -2.672 & -2.462 & \textbf{-1.795 }& \textbf{-1.997 }& -2.250 \\
\makecell[l]{UCI Redwine\\Unsupervised}& -6.574 & -6.496 & -8.244 & -8.297 & -2.992 & \textbf{-1.879 }& -8.077 &  & -8.676  \\
\makecell[l]{UCI Whitewine 2D}& -1.958 & -1.956 & -1.957 & -1.968 & -1.910 & \textbf{-1.891 }& \textbf{-1.915 }& \textbf{-1.901 } & -1.940 \\
\makecell[l]{UCI Parkinsons 2D}& -1.406 & -1.323 & -1.424 & -2.910 & -4.032 & -4.766 & \textbf{-1.134 }& -1.254 & -1.368  \\
\makecell[l]{UCI Parkinsons\\Unsupervised}& 0.999 & 0.304 & -3.265 & -3.214 & 1.328 & -3.654 & \textbf{3.055 }&  & -0.870 \\
\end{tabular}
\end{center}

\label{table:uci_ll}
\end{table*}

The results from the experiments using smaller UCI datasets
\citep{Dua:2019} are shown in Table \ref{table:uci_ll}. The UCI
datasets are preprocessed in the same way as done by
\cite{DBLP:conf/nips/UriaML13} i.e., we removed categorical variables
and variables which have absolute value of Pearson coefficient
correlation with other variable larger than 0.98. We use each UCI
dataset in two separate experiments. Firstly, by assuming the two last
columns as response variables (ordering as defined by the
documentation of the data), while the remaining columns constitute the
covariates. These are the datasets with suffix ``2D''. Secondly, we
perform experiments by using all columns as response variables, hence
assuming the covariate vector is empty. These datasets are given the
suffix ``unsupervised''.

MONDE models achieve the best results among all model trained in these datasets.

\begin{figure}[t!]
\begin{center}
\subfloat[RNADE]{\includegraphics[scale=0.4]{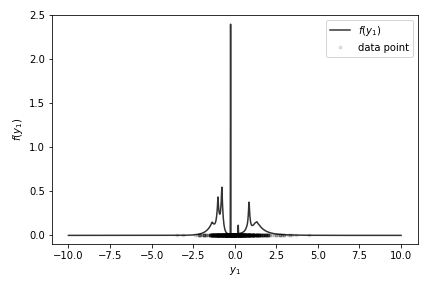}}\\
\subfloat[MONDE]{\includegraphics[scale=0.4]{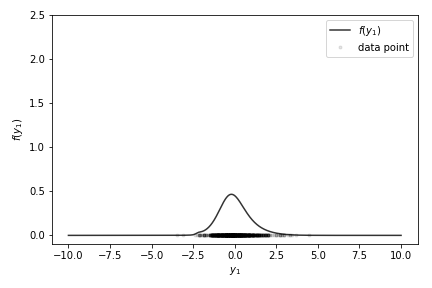}}
\end{center}
\caption{Densities estimated by the RNADE and MONDE AR models for the
  first variable in AR ordering when the corresponding variable had a
  small number of values. The data points are also shown on the $x$
  axis as black dots.}
\label{figure:categorical_problem}
\end{figure}

We report that when we trained the models on datasets composed solely
of continuous variables where some columns have a considerably small
number of unique values (but still specified as real type variables by
the dataset documentation), the RNADE models tend to ``overfit'' by
placing very narrow Gaussians at particular points (the test
likelihood is high, but this is an artefact of treating essentially
discrete data as if it had a density). This was especially a problem
for RNADE model, when one of such problematic variables was the first
variable in the autoregressive expansion for the joint probability. In
this case, because train, validation and test data contained a lot of
points with the same values, RNADE could create overfitted first
unconditional densities. This is not a realistic training procedure,
since the data here is discrete for all practical purposes, resulting
on unbounded test ``densities'' being easily achieved depending on the
minimal scale allowed for a Gaussian mixture component in the training
procedure. In Figure \ref{figure:categorical_problem}, we show an
example of the $f(y_1)$ density estimated by the RNADE model using a
mixture of Laplace distributions, and the density of the same variable
estimated by the autoregressive MONDE model. The data is the UCI
whitewine dataset (used as an unsupervised case i.e. all variables
treated as response variables with covariate set being empty). We
checked that MONDE models were not impacted by this spurious
``continuous'' variables and fitted smooth distributions. In our final
experiment runs, we imposed the rule to remove a column if it has below
5\% of unique values, compared to the number of samples. Only when we
removed such columns we were able to train the RNADE models to the
satisfactory level of generalization.

\subsection{FINANCIAL DATASETS}
\label{sec:financial_data}

\begin{table*}[hbt!]
\tiny
\caption{Mean Loglikelihoods - Financial Datasets.}
\begin{center}
\begin{tabular}{lrrrrrrrrrr}
{} & \makecell{RNADE\\Laplace} & \makecell{RNADE\\Normal} & \makecell{RNADE\\Deep\\Normal} & \makecell{RNADE\\Deep\\Laplace} & \makecell{MONDE\\Const\\Cov} & \makecell{MONDE\\Param\\Cov} & \makecell{MONDE\\AR} & \makecell{PUMONDE} & \makecell{MDN} \\
\hline \\
\makecell[l]{ETF 1D}& \textbf{-1.416 }& -1.495 & \textbf{-1.422 }& \textbf{-1.408 }& \textbf{-1.383 }&  &   & \textbf{-1.398 }&  \\
\makecell[l]{ETF 2D}& -1.501 & \textbf{-1.472 }& -1.857 & -1.588 & \textbf{-1.426 }& \textbf{-1.466 }& \textbf{-1.401 }& -1.599 & \textbf{-1.441 } \\
\makecell[l]{FX EUR Predicted}& \textbf{-1.054 }& -1.074 & -1.093 & -1.272 & -1.081 &  &   & -1.185 &  \\
\makecell[l]{FX EUR GBP Predicted}& -2.070 & -2.072 & -2.268 & \textbf{-2.024 }& -2.292 & -2.162 & -2.074 & -2.048 & -2.130  \\
\makecell[l]{FX ALL Predicted}& -2.940 & -2.985 & -3.479 & -3.741 & -4.853 & -8.107 & \textbf{-2.838 }&  & -5.604 \\
\end{tabular}
\end{center}

\label{table:fin_ll}
\end{table*}

The results from experiments with financial datasets are shown in
Table \ref{table:fin_ll}.  We use two different sources of the
financial data.

The first source is the the Yahoo
service\footnote{https://github.com/ranaroussi/yfinance} from which we
downloaded exchange traded funds dataset. We obtain two time series of
daily close prices between dates: 03.01.2011 and 14.04.2015. The ``ETF
1D'' are daily returns of the SPY financial instrument. The response
and covariate variables in this dataset are consecutive returns in the
time series accordingly. The ``ETF 2D'' are daily returns of the SPY
and DIA symbols. The returns are aggregated into the final dataset
using the same transformation as in the fist univariate case, but this
time both instrument returns are combined together so response and
covariate vectors have two components.

The second source is the FXCM
repository\footnote{{https://github.com/fxcm/MarketData}} from which
we downloaded the foreign exchange tick data. We downloaded prices for the
following currency pairs: EURUSD, GBPUSD, USDJPY, USDCHF, USDCAD,
NZDUSD, NZDJPY, GBPJPY for the period between 05.01.2015 and
30.01.2015. Each currency pair dataset contains top of the book tick
level bid and offer prices. We computed the returns of the mid point
prices for these time series. Then we subsumpled each time series using
a 1 minute interval and aligned all of them into one data frame. This
table of aligned currency pairs' returns was used to create three
datasets. ``FX EUR predicted'' contains the EURUSD returns as the
response variable and ten previous historical returns from all
currency pairs as covariates i.e. if we have EURUSD return at time $t$,
we collected for this response covariates as returns for all instruments
at times $t-1, t-2, \ldots , t-10$. The ``FX EUR and GBP predicted''
dataset was constructed as the previous one but with two response
variables i.e. EURUSD and GBPUSD returns and the covariates
constructed from all four previous historical returns plus hour of day
variable. The ``FX all assets predicted'' dataset contains all
contemporaneous currency pairs returns as response and two
autoregressive returns of all currency pairs as covariates.  We see
from the results in Table \ref{table:fin_ll} that there is an
improvement in using our approach in an autoregressive representation
and other versions of our model are also comparable with recently
proposed solutions.

\subsubsection{Classification}

In Section \ref{sec:classification} we used the following foreign
exchange instruments to construct the experiment dataset: AUDCAD, AUDJPY,
AUDNZD, EURCHF, NZDCAD, NZDJPY, NZDUSD, USDCHF, USDJPY, EURUSD, GBPUSD
and USDCAD.

\subsubsection{Bivariate likelihood}

In Section \ref{sec:bivariate_likelihood} we used the following foreign
exchange instruments to construct the experiment dataset: AUDCAD, AUDJPY,
AUDNZD, AUDCHF, EURAUD, EURCHF, NZDCAD, CADCHF, EURJPY, NZDJPY,
GBPNZD, GBPJPY, NZDUSD, USDCHF, GBPCHF, USDJPY, EURUSD, GBPUSD,
EURGBP, USDCAD and NZDCHF.

\subsection{TAIL DEPENDENCE}
\label{sup:sec:tail_dependence}

To compute tail dependence, we use empirical functions
suggested by \citep{CopTileDependence1} and \citep{Venter2001}:
\begin{align}
\label{eq:upper_tail_dep_est}
\hat{\lambda}_{L}(u) &= \frac{\sum_{k=1}^n \mathbf{1}(Y_{i,k} \leq \hat{F}_{i}^{-1}(u), Y_{j,k} \leq \hat{F}_{j}^{-1}(u))}{\sum_{k=1}^n \mathbf{1}(Y_{i,k} \leq \hat{F}_{i}^{-1}(u))} \\
\label{eq:lower_tail_dep_est}
\hat{\lambda}_{R}(u) &= \frac{\sum_{k=1}^n \mathbf{1}(Y_{i,k} > \hat{F}_{i}^{-1}(u), Y_{j,k} > \hat{F}_{j}^{-1}(u))}{\sum_{k=1}^n \mathbf{1}(Y_{i,k} > \hat{F}_{i}^{-1}(u))},
\end{align}
for the models where it is possible to directly sample new data (like
MAF and MDN), and for datasets used for training generated from the
mixture model. We plug-in the distribution transformation
$\hat{F}_{i}(y_{i,k})$ computed directly from sampled data using the
rank function.

To compute tail dependence indices for models which we cannot sample new
data easily but we can compute marginal distribution functions (like
PUMONDE, MONDE Copula) we apply the following process: 1) Estimate the marginal
quantile functions numerically conditioning it on the mean of one of
the mixtures: $\hat{F}_i^{-1}(\cdot|mean(\mathbf{x}_c))$; 2) Generate
vector $\mathbf{u}$ as a grid of points equidistantly spaced between
$(0,1)$ (we use the same grid $\mathbf{u}$ used for computing tail
indices for models that are easy to sample from); 3) Compute
$\hat{F}_k^{-1}(\mathbf{u}|mean(\mathbf{x}_c))$ for $k=i,j$ 4) Compute
$\hat{F}_{ij}(\hat{F}_i^{-1}(\mathbf{u}|mean(\mathbf{x}_c),
\hat{F}_j^{-1}(\mathbf{u}|mean(\mathbf{x}_c))| mean(\mathbf{x}_c))$
directly from the model and substitute it into the $\lambda_L(u)$ and
$\lambda_R(u)$ equations from Section \ref{sec:tail_dependence}.

\section{SOURCE CODE}
Source code is provided in \url{https://github.com/pawelc/NeuralLikelihoods}.

\onecolumn
\begin{tiny}
\begin{longtable}{lllll}
\caption{Hyper parameters search space}\\
\multicolumn{1}{c}{\bf EXPERIMENT}  & \multicolumn{1}{c}{\bf MODEL} & \multicolumn{1}{c}{\bf SEARCH SPACE} & \multicolumn{1}{c}{\bf DATA SET/BEST PARAMETERS} \\
\hline \\
\endhead

\multicolumn{5}{r}{{(\textit{table continues on the next page})}} \\
\endfoot
\endlastfoot
\label{table:hyper_param_grid}
\makecell[l]{Synthetic Data,\\Small UCI Data,\\Financial Data} & \makecell[l]{RNADE Normal} &  \makecell[l]{hidden units $\in (20,200)$\\ number of components in mixture $\in (1,100)$ \\ learning rate $\in (10^{-4}, 10^{-2})$ \\ batch size $=200$}  & \makecell[l]{Sin Normal (30;93;0.0044)\\Sin T (90;7;0.0017)\\Inv Sin Normal (69;7;0.0022)\\Inv Sin T (69;7;0.0022)\\MV Nonlinear (194;100;0.0089)\\UCI Redwine 2D (170;50;0.001)\\UCI Redwine Unsupervised (164;10;0.0011)\\UCI Whitewine 2D (100;24;0.0012)\\UCI Whitewine Unsupervised (56;30;0.0002)\\UCI Parkinsons 2D (104;11;0.0064)\\UCI Parkinson Unsupervised (200;24;0.0006)\\ETF 1D (69;7;0.0022)\\ETF 2D (164;1;0.0041)\\FX EUR Predicted (176;83;0.0046)\\FX EUR GBP Predicted (189;78;0.0027)\\FX All predicted (69;7;0.0022)} \\
\hline \\
\makecell[l]{Synthetic Data,\\Small UCI Data,\\Financial Data} & \makecell[l]{RNADE Laplace} &  \makecell[l]{hidden units $\in (20,200)$\\ number of components in mixture $\in (1,100)$ \\ learning rate $\in (10^{-4}, 10^{-2})$ \\ batch size $=200$}  & \makecell[l]{Sin Normal (100;24;0.0012)\\Sin T (56;30;0.0002)\\Inv Sin Normal (172;32;0.0011)\\Inv Sin T (200;67;0.0046)\\MV Nonlinear (90;67;0.0074)\\UCI Redwine 2D (20;100;0.0065)\\UCI Redwine Unsupervised (69;7;0.0022)\\UCI Whitewine 2D (69;7;0.0022)\\UCI Whitewine Unsupervised (100;24;0.0012)\\UCI Parkinsons 2D (90;67;0.0074)\\UCI Parkinson Unsupervised (138;25;0.0021)\\ETF 1D (200;1;0.01)\\ETF 2D (200;1;0.01)\\FX EUR Predicted (47;5;0.0017)\\FX EUR GBP Predicted (146;7;0.0011)\\FX All predicted (69;7;0.0022)} \\
\hline \\
\makecell[l]{Synthetic Data,\\Small UCI Data,\\Financial Data} & \makecell[l]{MONDE Const Cov} &  \makecell[l]{number of layers for x transformation $\in (0,3)$\\ number of hidden units per layer for x transformation $\in (10,200)$ \\number of layers y transformation $\in (1,5)$\\number of hidden units per layer for y transformation $\in (10,200)$\\learning rate $\in (10^{-4},10^{-2}$\\batch size $=200$}  & \makecell[l]{Sin Normal (1;131;4;10;0.0016)\\Sin T (1;112;2;110;0.0008)\\Inv Sin Normal (0;0;2;126;0.0004)\\Inv Sin T (1;112;2;110;0.0008)\\MV Nonlinear (3;113;1;89;0.0002)\\UCI Redwine 2D (0;0;4;10;0.0001)\\UCI Redwine Unsupervised (0;0;2;65;0.0002)\\UCI Whitewine 2D (0;0;3;28;0.0064)\\UCI Whitewine Unsupervised (0;0;3;200;0.0001)\\UCI Parkinsons 2D (1;112;2;110;0.0008)\\UCI Parkinson Unsupervised (0;0;1;10;0.0005)\\ETF 1D (3;200;1;200;0.0001)\\ETF 2D (0;0;1;200;0.0003)\\FX EUR Predicted (0;0;1;10;0.0001)\\FX EUR GBP Predicted (2;10;4;10;0.01)\\FX All predicted (0;0;5;200;0.0039)} \\
\hline \\
\makecell[l]{Synthetic Data,\\Small UCI Data,\\Financial Data} & \makecell[l]{MONDE Param Cov} &  \makecell[l]{number of layers for x transformation $\in (0,3)$\\ number of hidden units per layer for x transformation $\in (10,200)$ \\number of layers y transformation $\in (1,5)$\\number of hidden units per layer for y transformation $\in (10,200)$\\ number of layers for x covariance transformation $\in (1,3)$\\ number of hidden units per layer for x covariance transformation $\in (10,200)$ \\learning rate $\in (10^{-4},10^{-2}$\\batch size $=200$}  & \makecell[l]{MV Nonlinear (3;200;4;116;0;0;0.0003)\\UCI Redwine 2D (0;0;2;200;0;0;0.01)\\UCI Redwine Unsupervised (0;0;2;65;0;0;0.0002)\\UCI Whitewine 2D (2;17;2;35;2;79;0.0068)\\UCI Whitewine Unsupervised (0;0;4;69;0;0;0.0001)\\UCI Parkinsons 2D (1;121;1;184;0;0;0.0066)\\UCI Parkinson Unsupervised (0;0;4;70;0;0;0.0011)\\ETF 2D (1;54;3;127;?;?;0.0081)\\FX EUR GBP Predicted (0;0;5;10;?;?;0.0001)\\FX All predicted (1;68;3;191;?;?;0.01)} \\
\hline \\
\makecell[l]{Synthetic Data,\\Small UCI Data,\\Financial Data} & \makecell[l]{MDN} &  \makecell[l]{number of hidden layers $\in (1,6)$\\number of hidden units per layer $\in (20,200)$\\number of components in mixture $\in (1,100)$\\learning rate $\in (10^{-4},10^{-2}$\\batch size $=200$}  & \makecell[l]{Sin Normal (6;20;1;0.0001)\\Sin T (4;74;45;0.0003)\\Inv Sin Normal (2;82;95;0.0034)\\Inv Sin T (1;141;60;0.0022)\\MV Nonlinear (3;61;54;0.0067)\\UCI Redwine 2D (6;20;100;0.0002)\\UCI Redwine Unsupervised (6;188;14;0.0099)\\UCI Whitewine 2D (4;74;45;0.0003)\\UCI Whitewine Unsupervised (6;20;100;0.01)\\UCI Parkinsons 2D (1;104;11;0.0064)\\UCI Parkinson Unsupervised (6;94;98;0.0026)\\ETF 1D (2;114;9;0.0068)\\ETF 2D (2;114;9;0.0068)\\FX EUR Predicted (1;24;99;0.0001)\\FX EUR GBP Predicted (6;172;32;0.0011)\\FX All predicted (6;181;59;0.0011)} \\
\hline \\
\makecell[l]{Synthetic Data,\\Small UCI Data,\\Financial Data} & \makecell[l]{MONDE AR} &  \makecell[l]{number of hidden layers for x transformation $\in (0,3)$\\number of hidden units per x transformation layer $\in (10,200)$\\ number of hidden layers for y transformation $\in (1,5)$\\number of hidden units per y transformation layer $\in (10,200)$ \\learning rate $\in (10^{-4},10^{-2}$\\batch size $=200$}  & \makecell[l]{MV Nonlinear (1;22;1;180;0.0001)\\UCI Redwine 2D (0;0;3;28;0.0064)\\UCI Redwine Unsupervised (0;0;1;37;0.0038)\\UCI Whitewine 2D (0;0;2;159;0.0007)\\UCI Whitewine Unsupervised (0;0;2;10;0.0036)\\UCI Parkinsons 2D (1;121;1;184;0.0066)\\UCI Parkinson Unsupervised (1;130;1;199;0.0064)\\ETF 1D\\ETF 2D (0;0;1;176;0.01)\\FX EUR GBP Predicted (0;0;5;10;0.01)\\FX All predicted (0;0;3;28;0.0064)} \\
\hline \\
\makecell[l]{Synthetic Data,\\Small UCI Data,\\Financial Data} & \makecell[l]{PUMONDE} &  \makecell[l]{number of hidden layers for x transformation $\in (1,3)$\\number of hidden units per x transformation layer $\in (10,200)$\\ number of hidden layers for y transformation $\in (1,3)$\\number of hidden units per y transformation layer $\in (10,200)$\\ number of hidden layers for xy transformation $\in (1,3)$\\number of hidden units per xy transformation layer $\in (10,200)$ \\learning rate $\in (10^{-4},10^{-3}$\\batch size $=200$}  & \makecell[l]{MV Nonlinear (0;0;3;67;2;118;0.0002)\\UCI Redwine 2D (1;37;3;88;1;129;0.0007)\\UCI Redwine Unsupervised\\UCI Whitewine 2D (2;17;2;35;2;79;0.0068)\\UCI Parkinsons 2D (2;117;2;35;2;79;0.0068)\\ETF 2D (0;0;3;67;2;118;0.0002)\\FX EUR GBP Predicted (0;0;3;67;2;118;0.0002)} \\
\hline \\
\makecell[l]{Density Estimation} & \makecell[l]{MONDE MADE} &  \makecell[l]{number of hidden layers $\in \{8,10\}$\\number of blocks $\in \{40,60\}$ \\ start batch size $=128$ \\ batch size increments $=3$ \\ learning rate $=0.001$}   & \makecell[l]{Power (10;40)} \\
\hline \\
\makecell[l]{Density Estimation} & \makecell[l]{MONDE MADE} &  \makecell[l]{number of hidden layers $\in \{8,10\}$\\number of blocks $\in \{60,80\}$ \\ start batch size $=128$ \\ batch size increments $=5$ \\ learning rate $=0.001$}   & \makecell[l]{Gas (10;80) \\ Hepmass (8;60)} \\
\hline \\
\makecell[l]{Density Estimation} & \makecell[l]{MONDE MADE} &  \makecell[l]{number of hidden layers $\in \{3,5,7,8\}$\\number of blocks $\in \{10,50,80\}$ \\ start batch size $=128$ \\ batch size increments $=5$ \\ batch size increase patience threshold = $=20$ \\ learning rate $=0.001$}   & \makecell[l]{Miniboone (5;10)} \\
\hline \\
\makecell[l]{Density Estimation} & \makecell[l]{MONDE MADE} &  \makecell[l]{number of hidden layers $\in \{3,5,7\}$\\number of blocks $\in \{10,30,40\}$ \\ start batch size $=128$ \\ batch size increments $=3$ \\ batch size increase patience threshold = $=20$ \\ learning rate $=0.001$}   & \makecell[l]{Bsds300 (3;10)} \\
\hline \\
\makecell[l]{Classification} & \makecell[l]{MONDE Const Cov} & \makecell[l]{number of layers for x transformation $\in \{2,4\}$\\ number of hidden units per layer for x transformation $\in \{50,100\}$ \\number of layers y transformation $\in \{2,4\}$\\number of hidden units per layer for y transformation $\in \{50,100\}$\\number of hidden units per layer in y transformation used for x,y transformation $=30$\\ covariance learning rate $=0.05$ \\ start batch size $=128$ \\ batch size increments $=3$ \\ batch size increase patience threshold = $=20$ \\ learning rate $=0.001$}  & \makecell[l]{Classification (FX)  (2;100;4;100)} \\
\hline \\
\makecell[l]{Classification} & \makecell[l]{MONDE Param Cov} & \makecell[l]{number of layers for x transformation $\in \{2,4\}$\\ number of hidden units per layer for x transformation $\in \{50,100\}$ \\number of layers y transformation $\in \{2,4\}$\\number of hidden units per layer for y transformation $\in \{50,100\}$\\ number of layers used for covariance parametrization $\in \{2,4\}$ \\ number of hidden units per layer in covariance parametrization $\in \{50,100\}$ \\ number of hidden units per layer in y transformation used for x,y transformation $=30$ \\ start batch size $=128$ \\ batch size increments $=3$ \\ batch size increase patience threshold = $=20$ \\ learning rate $=0.001$}  & \makecell[l]{Classification (FX)  (2;100;4;50;2;100)} \\
\hline \\
\makecell[l]{Classification} & \makecell[l]{PUMONDE} & \makecell[l]{number of layers for x transformation $\in \{3,4\}$ \\ number of hidden units per layer for x transformation $\in \{50,100\}$ \\ number of layers for y transformation $\in \{3,4\}$ \\ number of hidden units per layer for y transformation $\in \{50,100\}$ \\ number of layers for x,y transformation $\in \{3,4\}$ \\ number of hidden units per layer for x,y transformation $\in \{50,100\}$ \\ number of hidden units per layer in y transformation used for x,y transformation $=30$   \\ start batch size $=128$ \\ batch size increments $=3$ \\ batch size increase patience threshold = $=20$ \\ learning rate $=0.001$}  & \makecell[l]{Classification (FX)  (3;50;4;50;4;50)} \\
\hline \\
\makecell[l]{Classification} & \makecell[l]{NN Classifier} & \makecell[l]{number of layers $\in \{2, 3,5\}$ \\ number of hidden units per layer$\in \{50,100\}$  \\ start batch size $=128$ \\ batch size increments $=3$ \\ batch size increase patience threshold = $=20$ \\ learning rate $=0.001$}  & \makecell[l]{Classification (FX)  (3;100)} \\
\hline \\
\makecell[l]{Classification} & \makecell[l]{Random Forest} & \makecell[l]{number of estimators $\in \{10,50,100\}$ \\ maximum tree depth $\in \{5,10,20\}$}  & \makecell[l]{Classification (FX)  (100;20)} \\
\hline \\
\makecell[l]{Classification} & \makecell[l]{Xgb Classifier} & \makecell[l]{subsample $\in \{0.1, 0.2, 0.3, 0.4, 0.5, 0.6, 0.7, 0.8, 1.0\}$}  & \makecell[l]{Classification (FX)  (0.3)} \\
\hline \\
\makecell[l]{Tail Dependence\\Mutual Information} & \makecell[l]{MONDE Const Cov} & \makecell[l]{number of layers for x transformation $\in \{3,4\}$\\ number of hidden units per layer for x transformation $\in \{50,100\}$ \\number of layers y transformation $\in \{3,4\}$\\number of hidden units per layer for y transformation $\in \{50,100\}$\\number of hidden units per layer in y transformation used for x,y transformation $=30$\\ covariance learning rate $=0.05$ \\ start batch size $=128$ \\ batch size increments $=3$ \\ batch size increase patience threshold = $=20$ \\ learning rate $=0.001$}  & \makecell[l]{Mixture Process (3;100;3;50)} \\
\hline \\
\makecell[l]{Tail Dependence\\Mutual Information} & \makecell[l]{MONDE Param Cov} & \makecell[l]{number of layers for x transformation $\in \{3,4\}$\\ number of hidden units per layer for x transformation $\in \{50,100\}$ \\number of layers y transformation $\in \{3,4\}$\\number of hidden units per layer for y transformation $\in \{50,100\}$\\ number of layers used for covariance parametrization $\in \{3,4\}$ \\ number of hidden units per layer in covariance parametrization $\in \{50,100\}$ \\ number of hidden units per layer in y transformation used for x,y transformation $=30$ \\ start batch size $=128$ \\ batch size increments $=3$ \\ batch size increase patience threshold = $=20$ \\ learning rate $=0.001$}  & \makecell[l]{Mixture Process (3;50;3;50;4;50)} \\
\hline \\
\makecell[l]{Tail Dependence\\Mutual Information} & \makecell[l]{PUMONDE} & \makecell[l]{number of layers for x transformation $\in \{3,4\}$ \\ number of hidden units per layer for x transformation $\in \{50,100\}$ \\ number of layers for y transformation $\in \{3,4\}$ \\ number of hidden units per layer for y transformation $\in \{50,100\}$ \\ number of layers for x,y transformation $\in \{3,4\}$ \\ number of hidden units per layer for x,y transformation $\in \{50,100\}$ \\ number of hidden units per layer in y transformation used for x,y transformation $=30$  \\ start batch size $=128$ \\ batch size increments $=3$ \\ batch size increase patience threshold = $=20$ \\ learning rate $=0.001$}  & \makecell[l]{Mixture Process (3;100;3;100;4;100)} \\
\hline \\
\makecell[l]{Tail Dependence\\Mutual Information} & \makecell[l]{MAF} & \makecell[l]{number of bijectors $\in \{2,4,5,8\}$ \\ number of layers in each bijector $\in \{1,2,4\}$ \\ number of hidden units per layer in bijector $\in \{64,128,512\}$ \\ number of hidden units for covariate transformation $\in \{16,32,64\}$ \\ batch normalization $\in \{True,False\}$ \\ learning rate $\in \{1e^{-3},1e^{-4}\}$  \\ batch size $=128$}  & \makecell[l]{Mixture Process (5;1;512;16;True;0.001)} \\
\hline \\
\makecell[l]{Tail Dependence\\Mutual Information} & \makecell[l]{MDN} & \makecell[l]{number of hidden layers $\in (2,3,5)$\\number of hidden units per layer $\in (128,512)$\\number of components in mixture $\in (2,10,50,100)$ \\ learning rate $\in \{1e^{-3},1e^{-4}\}$ \\ batch size $=128$ }  & \makecell[l]{Mixture Process (5;128;100;0.001)} \\
\hline \\
\makecell[l]{Bivariate Likelihood} & \makecell[l]{MONDE Const Cov} & \makecell[l]{number of layers for x transformation $\in \{2,4\}$\\ number of hidden units per layer for x transformation $\in \{50,100\}$ \\number of layers y transformation $\in \{2,4\}$\\number of hidden units per layer for y transformation $\in \{50,100\}$\\number of hidden units per layer in y transformation used for x,y transformation $=30$\\ covariance learning rate $=0.05$ \\ start batch size $=128$ \\ batch size increments $=3$ \\ batch size increase patience threshold = $=20$ \\ learning rate $=0.001$}  & \makecell[l]{FX (2;50;2;50)} \\
\hline \\
\makecell[l]{Bivariate Likelihood} & \makecell[l]{MONDE Param Cov} & \makecell[l]{number of layers for x transformation $\in \{2,4\}$\\ number of hidden units per layer for x transformation $\in \{50,100\}$ \\number of layers y transformation $\in \{2,4\}$\\number of hidden units per layer for y transformation $\in \{50,100\}$\\ number of layers used for covariance parametrization $\in \{2,4\}$ \\ number of hidden units per layer in covariance parametrization $\in \{50,100\}$ \\ number of hidden units per layer in y transformation used for x,y transformation $=30$ \\ start batch size $=128$ \\ batch size increments $=3$ \\ batch size increase patience threshold = $=20$ \\ learning rate $=0.001$}  & \makecell[l]{FX (2;50;4;100;2;100)} \\
\hline \\
\makecell[l]{Bivariate Likelihood} & \makecell[l]{PUMONDE} & \makecell[l]{number of layers for x transformation $\in \{2,3\}$ \\ number of hidden units per layer for x transformation $\in \{128,256\}$ \\ number of layers for y transformation $\in \{2,3\}$ \\ number of hidden units per layer for y transformation $\in \{64,128\}$ \\ number of layers for x,y transformation $\in \{2,3\}$ \\ number of hidden units per layer for x,y transformation $\in \{64,128\}$ \\ number of hidden units per layer in y transformation used for x,y transformation $=16$  \\ batch size $=128$ \\ learning rate $=0.001$}  & \makecell[l]{FX (2;128;3;128;3;64)} \\
\hline \\
\makecell[l]{Bivariate Likelihood} & \makecell[l]{MDN} & \makecell[l]{number of hidden layers $\in (2,3,5)$\\number of hidden units per layer $\in (128,512)$\\number of components in mixture $\in (2,10,50,100)$ \\ learning rate $\in \{1e^{-3},1e^{-4}\}$ \\ batch size $=128$ }  & \makecell[l]{FX (2;512;2;0.0001)} \\
\hline \\

\end{longtable}
\end{tiny}
\twocolumn

\end{document}